\documentclass[11pt,a4paper]{apriel}

\usepackage{amsfonts}
\usepackage{amsmath}
\usepackage{amssymb}
\usepackage{graphicx}
\usepackage{xcolor}
\usepackage{booktabs}
\usepackage{bm}
\usepackage{multirow}
\usepackage{mathtools}
\DeclarePairedDelimiter{\abs}{\lvert}{\rvert}
\usepackage{subcaption}
\usepackage{algorithm}
\usepackage{algorithmic}
\usepackage{xspace}

\providecommand{\Super}{Super\xspace}
\def\bx{\boldsymbol{x}}
\providecommand{\beginappendix}{\appendix}

\usepackage{bbold}

\graphicspath{{./}}

\title{\Super~Apriel: One Checkpoint, Many Speeds}

\newcommand{\snrAff}{\raisebox{.28em}{\scalebox{0.7}{\textbf{1}}}}
\newcommand{\snAff}{\raisebox{.28em}{\scalebox{0.7}{\textbf{2}}}}
\newcommand{\utorontoAff}{\raisebox{.28em}{\scalebox{0.7}{\textbf{3}}}}
\newcommand{\vectorAff}{\raisebox{.28em}{\scalebox{0.7}{\textbf{4}}}}
\newcommand{\udemAff}{\raisebox{.28em}{\scalebox{0.7}{\textbf{5}}}}
\newcommand{\milaAff}{\raisebox{.28em}{\scalebox{0.7}{\textbf{6}}}}
\newcommand{\commaAff}{\raisebox{.28em}{\scalebox{0.7}

{\textbf{,}\hspace{0.1em}}}}
\usepackage{tikz}
\DeclareRobustCommand{\dasharrow}[1][black]{
\tikz[baseline=-0.5ex]{\draw[dashed, line width=0.8pt, #1, ->] (0,0) -- (1,0);}
}
\definecolor{sftfull}{HTML}{0816E4}
\definecolor{sfttargeted}{HTML}{558B2F}
\definecolor{sftsingle}{HTML}{7B1FA2}
\usepackage{pifont}
\usepackage{tikz}
\usetikzlibrary{shapes.geometric, arrows.meta, positioning, calc}

\authorTeam{SLAM Labs}

\authorCore[\snrAff]{Oleksiy Ostapenko\coremark}
\authorCore[\snrAff]{Raymond Li\coremark}
\authorCore[\snrAff]{Torsten Scholak\coremark}

\authorOther[\snrAff\commaAff\utorontoAff\commaAff\vectorAff]{Alireza Mousavi-Hosseini}
\authorOther[\snAff]{Aman Tiwari}
\authorOther[\snrAff]{Denis Kocetkov}
\authorOther[\snrAff]{Joel Lamy Poirier}
\authorOther[\snAff]{Kelechi Ogueji}
\authorOther[\snrAff\commaAff\udemAff\commaAff\milaAff]{Nanda H Krishna}
\authorOther[\snrAff]{Rafael Pardinas}
\authorOther[\snAff]{Sathwik Tejaswi Madhusudhan}
\authorOther[\snAff]{Shruthan Radhakrishna}

\authorOrg[\snAff]{Srinivas Sunkara}
\authorOrg[\snrAff]{Val\'{e}rie B\'{e}caert}

\affiliation[\snrAff]{ServiceNow Research}
\affiliation[\snAff]{ServiceNow}
\affiliation[\utorontoAff]{University of Toronto}
\affiliation[\vectorAff]{Vector Institute}
\affiliation[\udemAff]{Universit\'{e} de Montr\'{e}al}
\affiliation[\milaAff]{Mila}
\contribution[]{
  \coremark~marks core contributors; authors sorted alphabetically.
  See \hyperref[sec:contributions]{Author Contributions}.
}
\correspondence{\href{mailto:torsten.scholak@googlemail.com}{\texttt{torsten.scholak@googlemail.com}}, \href{mailto:oleksiy.ostapenko@servicenow.com}{\texttt{oleksiy.ostapenko@servicenow.com}}}

\metadata[\huggingfacelogo Models:]{
  \href{https://huggingface.co/ServiceNow-AI/SuperApriel-15B-Base}{\texttt{SuperApriel-15B-Base}} \quad
  \href{https://huggingface.co/ServiceNow-AI/SuperApriel-15B-Instruct}{\texttt{SuperApriel-15B-Instruct}}
}
\metadata[\huggingfacelogo Dev Models:]{
  \href{https://huggingface.co/ServiceNow-AI/SuperApriel-0.5B-Base}{\texttt{SuperApriel-0.5B-Base}} \quad
}
\metadata[\githublogo Training Code:]{
  \href{https://github.com/ServiceNow/Fast-LLM}{\texttt{Fast-LLM}}~\sans{(training)} \quad
  \href{https://github.com/ServiceNow/pipeline-rl}{\texttt{pipeline-rl}}~\sans{(async RL)}
}
\metadata[\githublogo Placement Optimization:]{
  \texttt{place-layers}~\sans{(coming soon)}
}
\metadata[\wandblogo Training Logs:]{
  \href{https://wandb.ai/servicenow-team/Super_Apriel/groups/SuperApriel-15B}{\texttt{SuperApriel-15B}} \quad
  \href{https://wandb.ai/servicenow-team/Super_Apriel/groups/SuperApriel-0.5B}{\texttt{SuperApriel-0.5B}}
}

\abstract{
We release \Super~Apriel, a 15B-parameter supernet in which every decoder layer
provides four trained mixer choices---Full Attention~(FA), Sliding Window
Attention~(SWA), Kimi Delta Attention~(KDA), and Gated DeltaNet~(GDN).
A \emph{placement} selects one mixer per layer; placements can be switched
between requests at serving time without reloading weights, enabling multiple
speed presets from a single checkpoint.  The shared checkpoint also enables
speculative decoding without a separate draft model.
The \texttt{all-FA} preset matches the Apriel~1.6 teacher on all reported
benchmarks; recommended hybrid presets span $2.9\times$ to $10.7\times$ decode
throughput at 96\% to 77\% quality retention, with throughput advantages that
compound at longer context lengths.
With four mixer types across 48 layers, the configuration space is vast.
A surrogate that predicts placement quality from the per-layer mixer assignment
makes the speed--quality landscape tractable and identifies the best tradeoffs
at each speed level.
We investigate whether the best configurations at each speed level can be
identified early in training or only after convergence.  Rankings stabilize
quickly at 0.5B scale, but the most efficient configurations exhibit higher
instability at 15B, cautioning against extrapolation from smaller models.
\Super~Apriel is trained by stochastic distillation from a frozen Apriel~1.6
teacher, followed by supervised fine-tuning.  We release the supernet weights,
Fast-LLM training code, vLLM serving code, and a placement optimization toolkit.
}

\begin{document}
\maketitle

\section{Introduction}
\label{sec:intro}

As language models move toward longer contexts and higher serving throughput,
the inference cost of full attention (FA) increasingly becomes a bottleneck.
In autoregressive decoding, each FA layer reads and updates a KV cache that
grows linearly with sequence length. In practical serving systems, KV-cache
memory often limits batching and throughput~\citep{kwon2023pagedattention}.
Moreover, attention during decoding is frequently memory-bound: each step
processes a short query against a long cached context, so data movement rather
than arithmetic dominates runtime~\citep{shah2024flashattention3}.

A large body of work replaces FA in some or all layers with more
efficient alternatives---sliding-window
attention~\citep{beltagy2020longformer}, sparse attention
patterns~\citep{child2019sparse,zaheer2020bigbird}, linear-recurrence
sequence models~\citep{gu2023mamba}, and approaches that reduce
KV-cache footprint at inference time~\citep{wu2024lckv}.
In particular, \emph{hybrid} architectures that retain FA in a subset of layers
and use efficient mixers elsewhere can deliver large generation-speed
improvements while remaining competitive in quality~\citep{blakeman2025nemotronh,zuo2025falcon,chen2025minimaxm1,kimilinear2025,kimi2025linear}.

The central question in hybrids is not whether to mix mechanisms, but
\emph{where}: which layers should keep FA and which should use an efficient
alternative. The field has largely treated this optimization problem as
something that has to be settled \emph{once}. Some models fix placement at
design time (before pretraining), typically by ratio/spacing rules that are
tuned empirically by ablations~\citep{blakeman2025nemotronh,zuo2025falcon,chen2025minimaxm1,kimi2025linear}.
Other models determine placement during conversion, starting from a pretrained
attention teacher and deciding which layers to replace post hoc, again using
ablations and/or search to validate the choice~\citep{ostapenko2025aprielh1,gu2025jetnemotron}.
Despite these different workflows, the release artifact is typically the same:
a single checkpoint with \emph{one fixed placement}, exposing one operating
point.

A single placement can serve only one point on the speed--quality curve,
forcing a compromise prior to deployment.  This limitation surfaces in
several practical scenarios:
\begin{itemize}
  \item \textbf{Workload heterogeneity:}  Long-context decode-heavy
    traffic and short-prompt high-batch serving need different
    throughput--quality operating points. A fixed placement cannot
    satisfy both.
  \item \textbf{Load-adaptive serving:}  Deploying an FA-heavy placement
    during low-traffic hours for maximum quality and switching to an
    efficient placement during peak hours requires runtime flexibility
    that a single frozen architecture cannot provide.
  \item \textbf{Task-sensitive quality tradeoffs:}  Not all capabilities
    degrade equally as FA layers are replaced: long-range retrieval is
    far more sensitive than local reasoning or generation, so the right
    operating point depends on the application's reliance on faithful
    recall over long contexts.
\end{itemize}
Serving these diverse operating points from fixed-placement models
requires a separate training phase, separate deployment, and repeated
validation per checkpoint.

\Super~Apriel targets a different deployment goal: \emph{flexible placement at
serving time} from a single release.
Derived from Apriel~1.6~\citep{apriel16blog}, a 48-layer multimodal transformer,
\Super~Apriel is to our knowledge the first release of
a \emph{token-mixer supernet} in which \emph{every decoder layer} exposes four
\emph{trained} token mixer options that can be selected at runtime: full attention
(FA), Sliding Window Attention (SWA), Kimi Delta Attention (KDA)~\citep{kimi2025linear},
and Gated DeltaNet (GDN)~\citep{yang2025gateddeltanet}. A \emph{placement}
selects one mixer per layer, yielding a member of a family of architectures
that share most parameters---FFNs, embeddings, and normalization---and
differ only in the mixer blocks. Placements can be switched at serving
time from a single checkpoint, making the throughput--quality tradeoff a
runtime control surface rather than a release-time commitment.

This flexibility is useful only if placement can be chosen well, and that
is genuinely hard.  Layers are not interchangeable: in a trained
transformer, attention layers specialize into position-specific
computational roles, so some can be approximated by efficient mixers with
negligible loss while others are
critical~\citep{elhage2021framework}.  Prior work addresses this by
searching for a single good placement and committing to
it~\citep{ostapenko2025aprielh1,gu2025jetnemotron}---a pragmatic choice,
since leaving placement unconstrained would simply shift the hard
problem to serving time.  \Super~Apriel resolves this
differently.  Because all mixer options are already trained at every
layer, and because a fitted surrogate model makes
the speed--quality landscape tractable (Section~\ref{sec:placement}), we
can sweep the Pareto frontier \emph{before release} and identify a
good placement preset at each speed target. Serving a particular
workload then reduces to selecting the appropriate preset from a curated
menu, rather than navigating the $4^{48}$ combinatorial space directly.
Table~\ref{tab:presets} summarizes the recommended presets, where the \texttt{all-FA} preset matches the Apriel~1.6 teacher across the reported
benchmarks.  The fastest recommended hybrid preset achieves
10.7$\times$ decode throughput at 77\% quality retention,
and all operating points are served from a single checkpoint and deployment.
We summarize the main targeted deployment strategies for \Super~Apriel in Appendix~\ref{app:motivation_serving}.

Our contributions are:
\begin{itemize}
  \item \textbf{\Super~Apriel attention supernet.} A single checkpoint in which every
    decoder layer provides four trained mixer choices (FA, SWA, KDA, GDN),
    enabling flexible runtime placement and multiple speed presets from one
    deployment.  The shared checkpoint also enables speculative decoding
    without a separate draft model (Section~\ref{sec:speculative}).
    Throughput advantages compound with context length: efficient
    placements gain 80--155\% relative speedup from 16K to 32K sequence
    length, substantially outpacing external hybrid baselines
    (5--46\% gains).
  \item \textbf{Training recipe.} Stochastic distillation from a frozen Apriel~1.6
    teacher on a curated data mixture, training all mixers simultaneously in a
    single run, followed by supervised fine-tuning.
  \item \textbf{Placement optimization.} A cluster-expansion surrogate
    that admits exact cost-constrained optimization, efficiently sweeping
    the speed--quality Pareto frontier.
  \item \textbf{Placement landscape dynamics.} Controlled ablations on a
    0.5B dev model and the 15B supernet investigate whether and how
    placement rankings evolve during training.  Rankings stabilize early
    overall, but frontier placements exhibit higher volatility---especially
    at 15B scale, where small-scale findings do not automatically transfer.
  \item \textbf{Open-source release.} Supernet weights, Fast-LLM~\citep{fastllm} training code, vLLM serving code, and a placement optimization toolkit.
\end{itemize}

Section~\ref{sec:arch} describes the mixer vocabulary and supernet architecture.
Section~\ref{sec:training} covers distillation and supervised fine-tuning.
Section~\ref{sec:placement} presents the placement optimization framework.
Section~\ref{sec:landscape_dynamics} investigates how placement rankings
evolve during training via controlled ablations on both the 0.5B development
model and the 15B model.
Section~\ref{sec:inference} covers inference and deployment.
Section~\ref{sec:eval} reports benchmark results.
Section~\ref{sec:limitations} discusses limitations.
Section~\ref{sec:outlook} discusses planned post-training stages and
deployment directions.

\begin{figure}[t!]
    \centering
    \includegraphics[width=\linewidth]{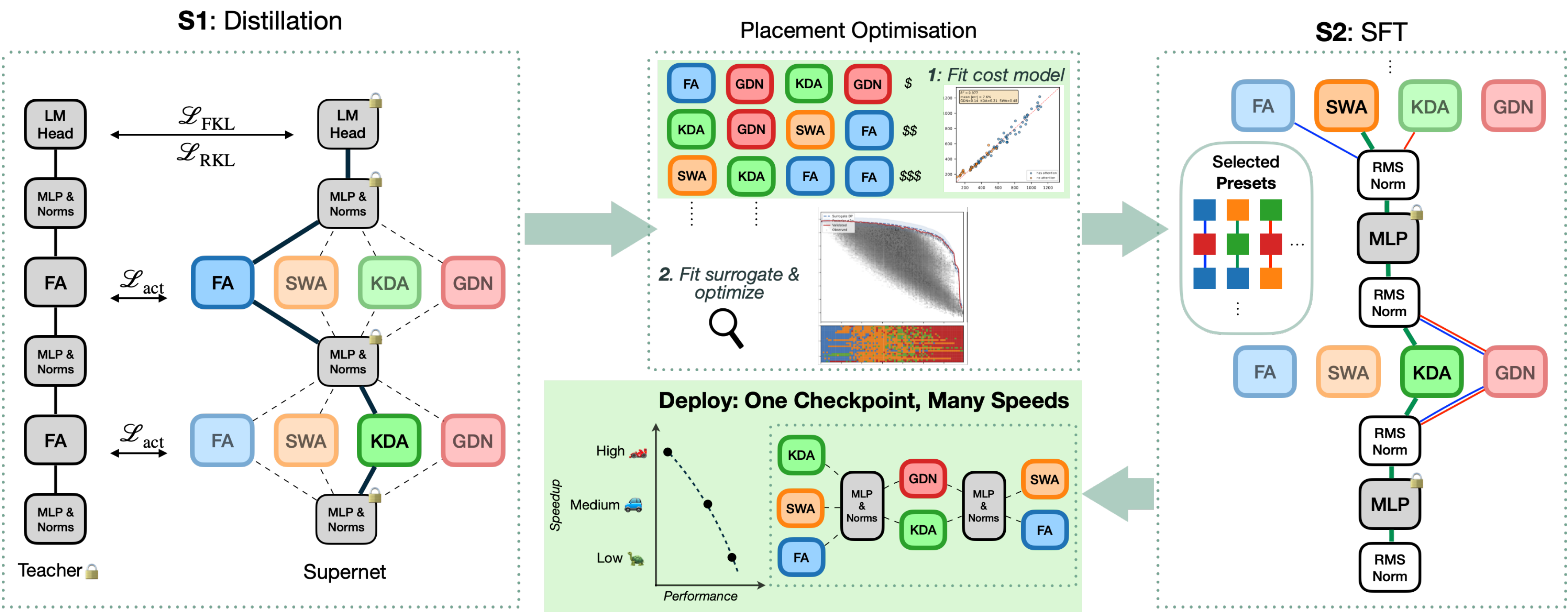}
    \caption{\Super~Apriel architecture and training recipe. We leave out residual connections for simplicity. Training starts with \texttt{S1: Distillation} phase (Section~\ref{sec:training:distillation}), where at every step and every layer, a mixer is drawn uniformly from one of four mixer types. Distillation uses activation matching, forward KL, and reverse KL losses. All parameters (MLP, LayerNorm, etc) except mixers remain frozen throughout distillation.
    After distillation, we perform placement optimization (Section~\ref{sec:placement}) and find a set of candidate placements for each target cost-tier. In the second training stage \texttt{S2: SFT:} we perform targeted training sampling exclusively from a set of candidate presets. One may also perform SFT on all placements and use global sampling of mixers in this phase (Section~\ref{sec:training:sft}). The final checkpoint supports multiple placements that are at the throughput-performance Pareto frontier according to the cost model used in the placement optimization phase.}
    \label{fig:supernet_architecture}
\end{figure}

\begin{figure}[htbp]
    \centering
    \includegraphics[width=1\linewidth]{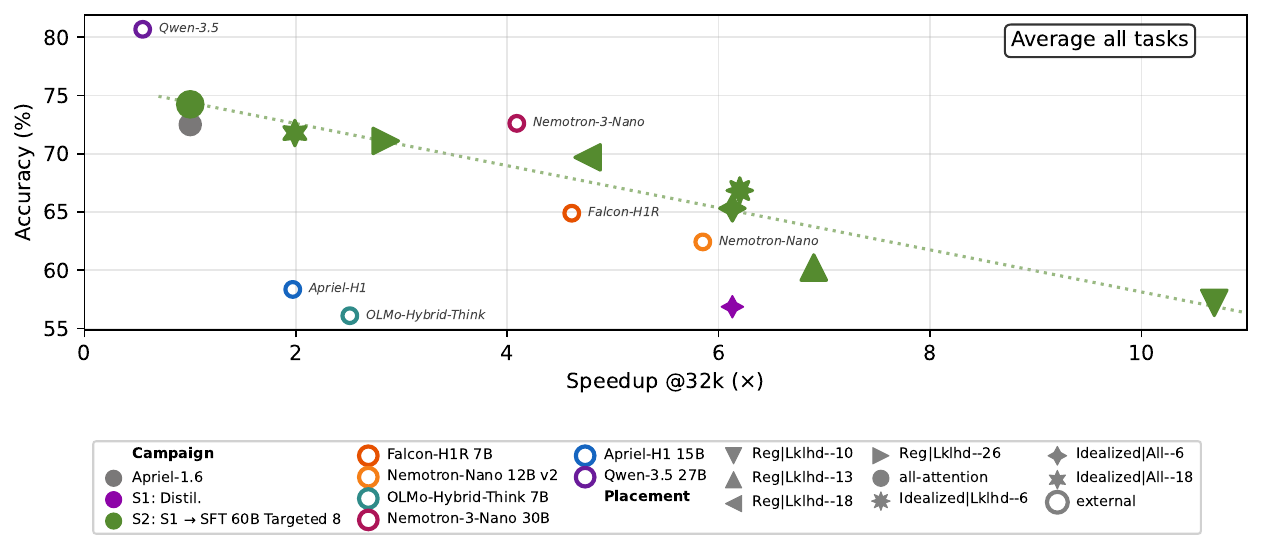}
    \caption{Performance of \Super~Apriel 15B versus decode throughput speedup @32k sequence length (over 500 samples) relative to the \texttt{all-FA} preset (Apriel-1.6). Detailed description of the stages can be found in Section~\ref{sec:training}. Each color represents a model checkpoint. Marker shapes denote the layer placement configuration. Filled markers are \Super~Apriel variants at different cost–performance operating points; open circles are external hybrid baselines. The dotted line shows a linear fit to the \texttt{S2: S1 $\rightarrow$ SFT 60B Targeted 8} checkpoint placements. Higher and further right is better. Full data is presented in Table~\ref{tab:eval_results_on_dev}, evaluation details including filters and benchmarks are detailed in Section~\ref{sec:eval:suite}, details on throughput benchmarking are in Appendix~\ref{app:throughput_eval}.}
    \label{fig:per_task15B_all}
\end{figure}

\begin{figure}[t]
  \centering
  \begin{subfigure}[t]{0.48\textwidth}
    \centering
    \includegraphics[width=\linewidth]{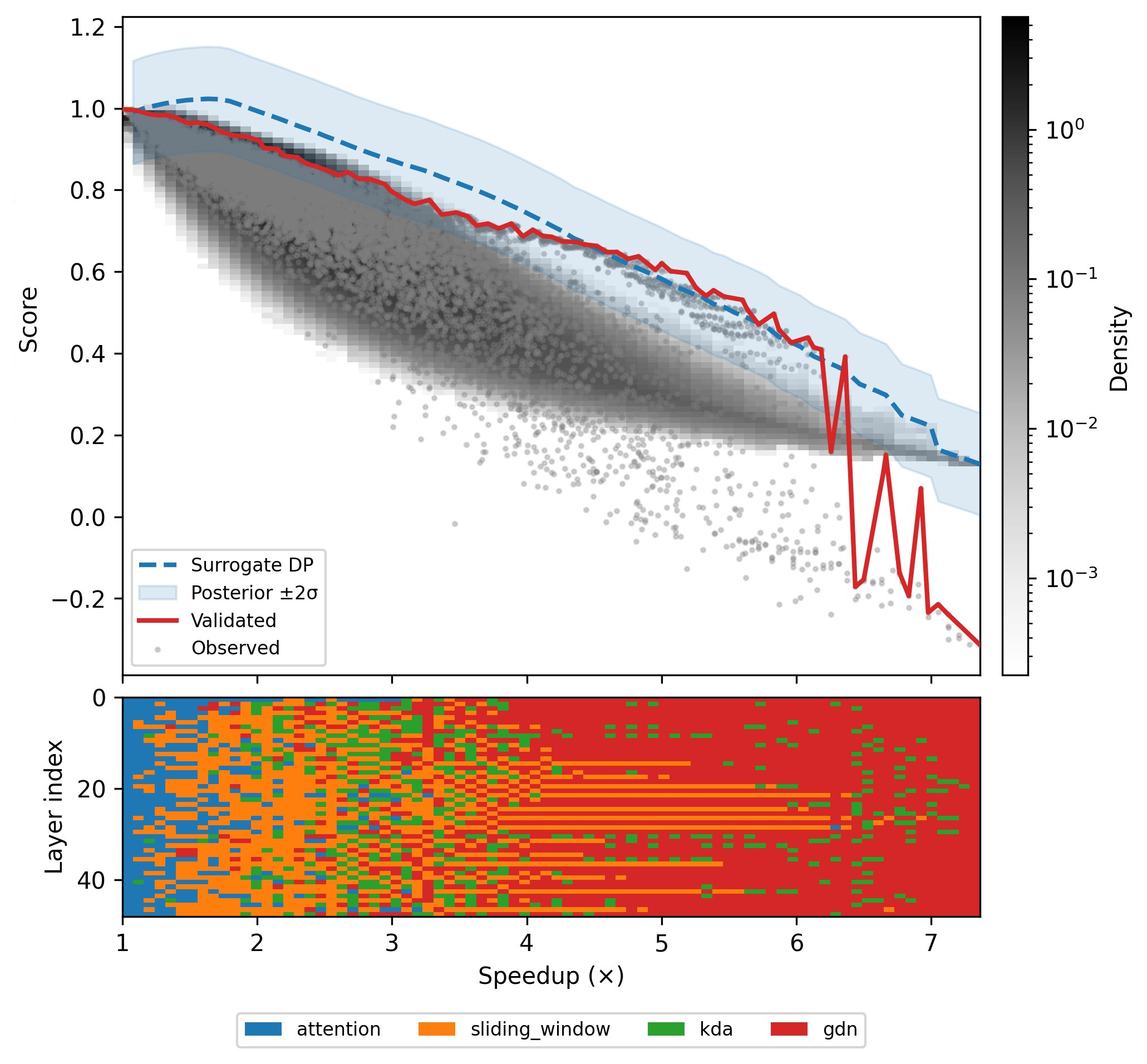}
    \caption{15B supernet Pareto frontier}
    \label{fig:pareto}
  \end{subfigure}
  \hfill
  \begin{subfigure}[t]{0.48\textwidth}
    \centering
    \includegraphics[width=\linewidth]{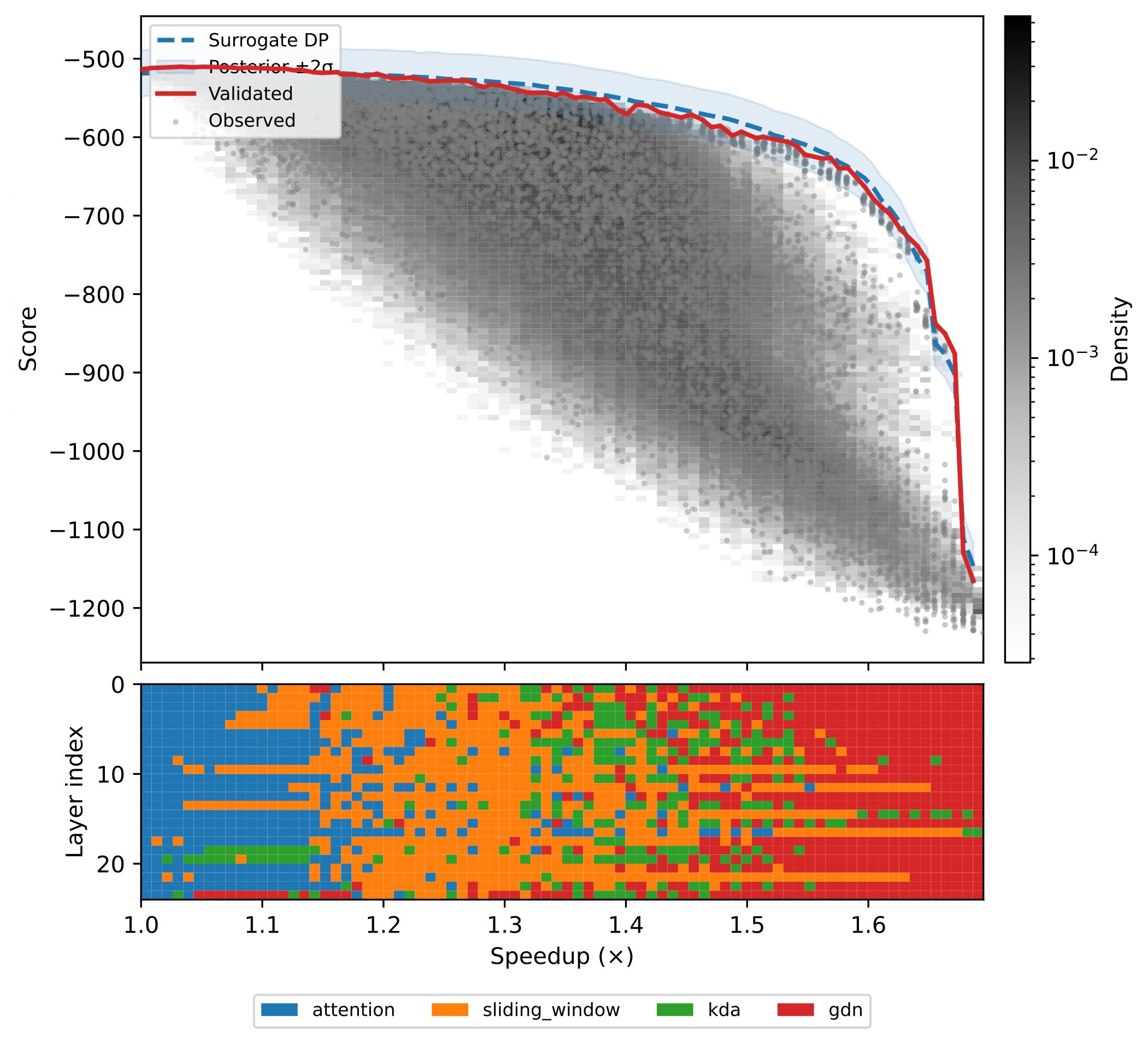}
    \caption{0.5B development supernet Pareto frontier}
    \label{fig:pareto_05b}
  \end{subfigure}
  \caption{Pareto frontiers for (a) 15B supernet and (b) 0.5B development supernet: placement quality vs.\ inference speedup (according to the \texttt{Regression}-based cost model, see Section~\ref{sec:placement}). \textbf{Top panels:}~Score for all feasible placements (grayscale density, log scale), observed supernet evaluations (grey dots), surrogate-predicted optima (blue curve), and validated optima (red solid line).
  For the 15B model, the score is the average normalized log-likelihood on MATH500, AIME24, and AIME25 traces.
  For the 0.5B model, the score is the average raw log-likelihood on MATH500 traces.
  \textbf{Bottom panels:}~Layer-wise mixer assignment of the optimal placement at each speedup level. Mid-network layers (18--28) retain full attention longest. The 0.5B model reproduces the same qualitative landscape structure as the 15B supernet, confirming findings in Section~\ref{sec:landscape_dynamics}.}
  \label{fig:pareto_combined}
\end{figure}

\section{Architecture}
\label{sec:arch}

\paragraph{Teacher.}
Apriel~1.6~\citep{apriel16blog} is a 15B-parameter multimodal
transformer derived from Pixtral-12B via depth
upscaling (increasing the number of layers)~\citep{radhakrishna2025apriel15}: 48~decoder layers with
grouped-query attention (32~query / 8~KV heads, $d_h{=}128$), hidden
dimension $d{=}5120$, SiLU-gated FFN of width 14\,336, vocabulary of
131\,072 tokens, and a context window of 262\,400 positions.  A Pixtral
vision encoder and learned projector handle image inputs.

\paragraph{Mixer vocabulary.}
Each layer's sequence mixer can be one of four types spanning a range of
computational cost and expressiveness.  Full Attention~(FA) computes
exact pairwise token interactions with $O(n^2)$ time and $O(n)$ KV cache
per layer.  Sliding Window Attention~(SWA) restricts attention to a
local window of $w = 4096$ tokens, bounding decode cost independent of
context length.  Gated DeltaNet~(GDN)~\citep{yang2025gateddeltanet} combines
scalar gating with the delta rule in a matrix-valued recurrent state,
operating in $O(n)$ time with $O(1)$ fixed-size state.  Kimi Delta
Attention~(KDA)~\citep{kimi2025linear} extends GDN with channel-wise
(per-dimension) gating, so that different state dimensions decay at
different rates.  Table~\ref{tab:mixers} summarizes the four types.

\begin{table}[t]
  \centering
  \caption{Mixer vocabulary. Relative cost of each token mixer (attention = 1.00),
  measured on H100 with 15B-parameter Apriel~1.6 hybrids,
  sequence length 16k.
\textbf{Cost} (regression) --- cost is represented by coefficients of a linear fit to $1/\text{throughput}$ (each mixer's count $\geq 3$ layers);
Section~\ref{sec:placement} and Appendix~\ref{app:cost_models} provide further details.}
  \label{tab:mixers}
  \begin{tabular}{llclcc}
    \toprule
    \textbf{Mixer} & \textbf{Mechanism} & \textbf{State kind} & \textbf{State size / layer}
      & \textbf{Cost} \\
    & & & & (regression)  \\
    \midrule
    FA  & Softmax GQA        & KV cache    & $O(n \cdot d)$             & 1.00 \\
    SWA & Windowed GQA       & Bounded KV  & $O(w \cdot d)$             & 0.48 \\
    KDA & Channel-wise delta & Recurrent   & $32 \times 128 \times 128$ & 0.21 \\
    GDN & Gated delta rule   & Recurrent   & $8 \times 128 \times 128$  & 0.14 \\
    \bottomrule
  \end{tabular}
\end{table}

Throughput depends on the \emph{allocation} (how many of each type)
rather than the \emph{placement} (which layers get which type), because
each layer's kernel time is independent of position.  Full architectural
details --- recurrence equations, projection structures, convolution
configurations --- are given in Appendix~\ref{app:mixer_details}. For the rest of this manuscript, we default to the following vocabulary: \emph{placement} --- assigns a mixer type to each layer; \emph{preset} --- refers to placements that were selected for targeted training or deployment.

Figures~\ref{fig:per_task15B_all} and~\ref{fig:pareto_combined} preview the
main result.  The \texttt{all-FA} preset matches the Apriel~1.6 teacher
(74.2 vs.\ 73.9 all-tasks average), and recommended hybrid presets span
$2.9\times$ to $10.7\times$ decode throughput at 96\% to 77\% quality
retention from a single checkpoint (Table~\ref{tab:presets}).
The remaining sections describe how these results are obtained;
Section~\ref{sec:eval} presents the full analysis.

\section{Supernet Training}
\label{sec:training}

\subsection{Model Surgery}
\label{sec:training:surgery}

Super Apriel 15B is derived from Apriel~1.6 by equipping every decoder layer with all $\abs{\mathcal{M}}{=}4$ mixer types ($\mathcal{M} = \{\text{FA}, \text{SWA}, \text{KDA}, \text{GDN}\}$); with four copies present, the checkpoint contains 25B~parameters, of which ${\sim}$15B are active
in any single placement.  FA and SWA weights come directly from the
teacher.
The GDN and KDA layers are initialized using the Delta-in-the-Llama (DIL) and Kimi-in-the-Llama (KIL) recipes (Appendix~\ref{app:init}) where GDN and KDA projections' weights are copied from the source model's own attention weights, without requiring any prior distillation checkpoint --- a procedure that generalizes to any GQA transformer.
Section~\ref{sec:landscape_dynamics} applies this to a 0.5B model
derived from an external Qwen2-0.5B~\citep{yang2024qwen2} architecture.

\subsection{Distillation}
\label{sec:training:distillation}
Throughout this report, we refer to this stage as \texttt{S1: Distill}. As explained below, for the \Super~Apriel 15B version, this stage itself is composed of two substages. First, GDN and KDA mixers were trained in an earlier distillation run against the Apriel~1.5 teacher~\citep{radhakrishna2025apriel15}; because 1.6 is a post-trained continuation of 1.5 with the same decoder architecture, the transferred weights remain compatible (Appendix~\ref{app:init}).  Approximately 80\% of active parameters per placement (FFNs, embeddings, layer norms, vision encoder) are shared; only the mixer weights differ. Next, all mixers are trained against Apriel 1.6 teacher.

\paragraph{Local sampling.}
At each forward pass, exactly one mixer is active per layer, selected
uniformly at random and independently across layers.  The selection seed
is derived deterministically from the global training seed and the
optimizer step count, ensuring reproducibility across checkpoint
restarts.  Throughout training, each mixer at each layer receives gradient in
approximately $1/\abs{\mathcal{M}}$ of all steps.

\paragraph{Objective.}
The two substages differ in teacher model and loss
composition. The supernet is trained by distilling from the frozen teacher
using a composite loss:
\[
  \mathcal{L} = \lambda_{\text{act}} \, \mathcal{L}_{\text{act}}
  + \lambda_{\text{RKL}} \, \mathcal{L}_{\text{RKL}}
  + \lambda_{\text{FKL}} \, \mathcal{L}_{\text{FKL}}
\]
Activation distillation ($\mathcal{L}_{\text{act}}$) matches
intermediate representations at layer boundaries with a mean squared-error loss, providing a dense
per-layer gradient signal.  Reverse KL ($\mathcal{L}_{\text{RKL}}$)
aligns the student's output distribution with the teacher's.  Forward
KL ($\mathcal{L}_{\text{FKL}}$), added in distillation stage~2, provides a
complementary mode-covering term.  Per-stage weights are in
Table~\ref{tab:training}.

\paragraph{Training data.}
The two distillation stages total 266B~tokens.
Stage~1 distills from the Apriel~1.5 teacher for 197B~tokens,
training the GDN and KDA mixers from their initialized state.
Stage~2 switches to the Apriel~1.6 teacher and continues for
69B~tokens on the same data mixture drawn from the Apriel pretraining
corpus and supervised fine-tuning data.  From Apriel-H1~\citep{ostapenko2025aprielh1},
we learned that distillation data composition had a larger effect on
quality preservation than the algorithm, learning rate schedule, or
initialization.  Distilling on raw
pretraining data failed: reasoning quality collapsed.  Distilling on
high-quality reasoning traces (multi-step proofs, structured
problem-solving, code with logical dependencies) succeeded.
We attribute this to the teacher's reasoning capability depending on
specific attention circuits that form during training on structured
data.  When efficient mixers replace these circuits, they must learn
alternative computational paths, which requires concentrated examples
that exercise the same behaviors.

\begin{table}[h]
\centering
\caption{Training data composition: token proportions by source category. We use the \texttt{Distillation} dataset for all the distillation runs and some ablations. We use \texttt{SFT} dataset for all post-distillation fine-tuning runs unless stated otherwise. Some earlier training runs used a starred \texttt{SFT$^*$} version, which excludes the tool use subset.}
\label{tab:data-mix}
\begin{tabular}{lr}
\toprule
\textbf{Category} & \textbf{Distillation (\%)}
\\
\midrule
Reasoning traces \& SFT & 29.3  \\
Code & 10.0 \\
Math \& STEM & 5.0 \\
Web / encyclopedic text & 6.7 \\
Image (multimodal) & 49.0 \\
\midrule
\textbf{Category} & \textbf{SFT (\%)} \\
\midrule
SFT: Code & 36.1 \\
SFT: Math \& STEM & 38.7 \\
SFT: Chat, generic reasoning \& IF & 11.3 \\
SFT: Tool use & 12.0 \\
SFT: Safety, robustness \& content moderation & 2.0 \\
\bottomrule
\end{tabular}
\end{table}

\begin{figure}[htbp]
    \centering
    \includegraphics[width=\linewidth]{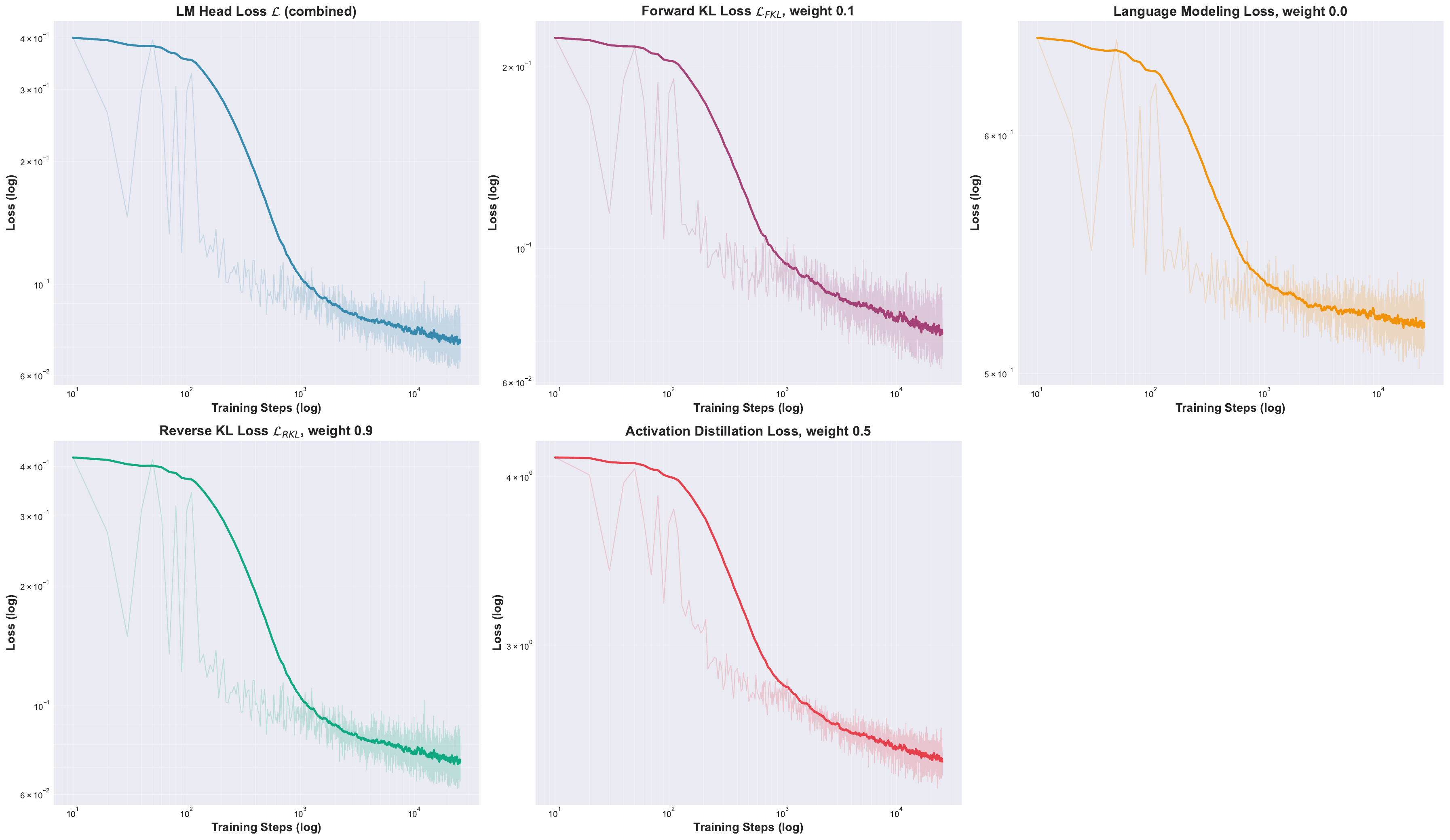}
  \caption{Training losses for stage 1 distillation performed for 25k steps with subsequence learning rate annealing phase (not shown here). We use batch size of 120 and sequence length of 16,384 in this stage. Language modeling loss is tracked but not used in training (see weight = 0.0). We use local mixer sampling at this stage (see Section~\ref{sec:gs}). The losses are plotted without scaling, the scaling coefficient weights $\alpha$ used during training are in the titles of each plots.}
  \label{fig:training_losses_distillation}
\end{figure}

\paragraph{Compute.}
Training is implemented in Fast-LLM~\citep{fastllm} with AdamW
($\beta_1 = 0.9$, $\beta_2 = 0.95$, weight decay $0.1$), a
learning rate of $10^{-5}$, and bfloat16 mixed precision.  Each step
processes 120~sequences of 16\,384~tokens (${\sim}$2M~tokens). Unless stated otherwise, we only train the mixer weights and freeze all shared parameters (see~Figure \ref{fig:mxro_vs_all} for ablation).
Stage~1 uses up to 192~H100 GPUs; Stage~2 uses 120.  Total cost is
${\sim}$70K~H100 GPU-hours across both stages.  Throughput varies
between 1000--1200 tokens/s/GPU: steps sampling more FA layers are
slower; steps with more GDN/KDA layers are faster.  Per-stage
hyperparameters are in Table~\ref{tab:training}. We plot loss curves for $\mathcal{L}_{\text{act}}$, $\mathcal{L}_{\text{RKL}}$,
$\mathcal{L}_{\text{FKL}}$ for the second distillation stage against Apriel~1.6 teacher in Figure~\ref{fig:training_losses_distillation}.

\subsection{Supervised Fine-Tuning}
\label{sec:training:sft}

Throughout this report, we refer to this stage or checkpoints coming out of it as \texttt{S2: S1 $\rightarrow$ SFT \dots}. After distillation, the shared parameters---FFNs, embeddings, layer norms,
output head---remain as they were in the teacher, having never been updated to
account for the new mixer types.  This constraint is by design: freezing shared
parameters ensures that each mixer learns against a stable reference
representation.  But it may impose a quality ceiling.  Tasks that depend on tight
coordination between mixer outputs and downstream layers---multi-step reasoning,
long-range factual recall---are most affected, because the shared
representations cannot adjust to the different computational signatures of
GDN and KDA\@.  Apriel-H1~\citep{ostapenko2025aprielh1} exhibited the same
pattern: conversion regressed reasoning benchmarks, and supervised fine-tuning
(SFT) on task-specific data recovered them.

\begin{figure}[htbp]
    \centering
    \includegraphics[width=0.9\linewidth]{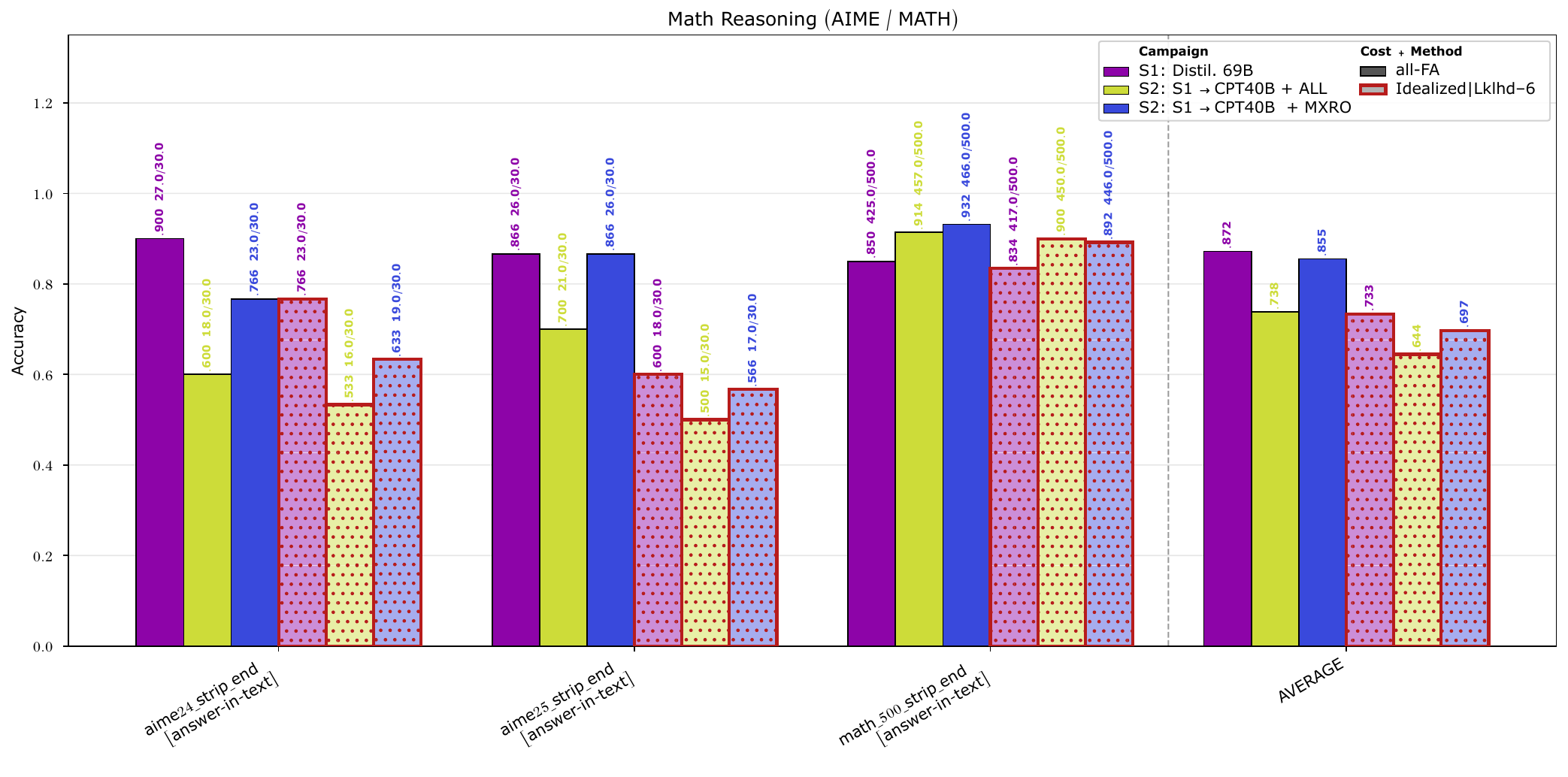}
  \caption{Comparison of mixer only training and freezing all other shared parameters (MXRO) vs. training all parameters (ALL) on the \texttt{S1: Distil.} \Super~Apriel 15B checkpoint finetuned for 40B tokens on the distillation dataset (denoted as CPT 40B here).}
  \label{fig:mxro_vs_all}
\end{figure}

For \Super~Apriel, however, we take a more conservative approach: shared
parameters remain frozen during SFT as well, and only mixer weights are
trained. This is motivated by an ablation presented in Figure~\ref{fig:mxro_vs_all} that compares full fine-tuning vs. mixer-only fine-tuning with the latter performing better after a short training run. On long SFT runs mixer-only training may sacrifice the potential quality recovery from shared-parameter adaptation but preserves the stable reference representation that all placements depend on.

In SFT, we train the mixers
on instruction-tuning data using standard next-token prediction,
with no teacher signal.
The sequence length doubles to 32\,768~tokens,
matching the deployment context window.
For a supernet, this raises a design
question that does not arise in single-architecture models: which placements
should receive gradient signal during training?

\paragraph{Placement Sampling.}
\label{sec:gs}
Local sampling, used in distillation, draws each layer's mixer independently and uniformly from the $\abs{\mathcal{M}}$ types.  The number of placements compatible with
an allocation $(n_m)_{m \in \mathcal{M}}$ is the multinomial coefficient
$L!/\prod_m n_m!$, which is maximized at the balanced allocation
$n_m = L/\abs{\mathcal{M}}$ and falls off rapidly: for $L{=}48$,
$\abs{\mathcal{M}}{=}4$, the balanced allocation $(12,12,12,12)$ is
${\sim}10^{26}$ times more likely to be sampled than all-attention, which
has exactly one compatible placement.
Allocations near the mode dominate; extreme allocations are vanishingly
rare under local sampling.

To mitigate this, in the SFT phase we follow a different strategy. Here the pool of possible placements that can be sampled during the training is either \emph{unconstrained} (includes all possible placements) -- in which case we use the \emph{global sampling} strategy that equalizes the probability of sampling extreme allocations detailed in Appendix~\ref{app:global_placement}; or \emph{constrained}. The latter \emph{targeted} placement sampling constrains the pool of possible placements to a short list of candidate placements to concentrate
gradient signal on the configurations most likely to be deployed. Targeted
training is more data-efficient per placement but introduces a coupling between
training and evaluation that is absent under the unconstrained sampling: the shared
representation adapts to the targeted placements' loss surfaces, and subsequent
evaluation of those placements cannot distinguish genuine quality from
self-reinforcing specialization (Section~\ref{sec:landscape_dynamics} formalizes this failure mode and tests it experimentally). The targeted approach also presumes that the set of
high-value placements is known before SFT begins; identifying it is the subject
of the next section.

\paragraph{Compute.}
SFT is initialized from the distillation endpoint \texttt{S1: Distil.} and trains with AdamW
($\beta_1 = 0.9$, $\beta_2 = 0.95$, weight decay $0.1$), a learning rate of
$10^{-5}$ with 1\,000 warmup steps followed by constant schedule, and bfloat16
mixed precision.  Each step processes 120~sequences of 32\,768~tokens
(${\sim}$3.9M~tokens).  Training uses 120~H100 GPUs with ZeRO-2 parallelism
and 8-way tensor parallelism, achieving ${\sim}$1\,690~tokens/s/GPU\@.

\paragraph{SFT Campaigns.} In this report we explore the following three SFT campaigns: \texttt{S2: S1 $\rightarrow$ SFT$^*$ 137B Full  supernet} --- SFT fine-tuning of the full supernet on 137B tokens using global sampling; \texttt{S2: S1 $\rightarrow$ SFT$^*$ 40B Single Placement} --- single placement trained for 40B tokens; \texttt{S2: S1 $\rightarrow$ SFT 60B Targeted 8} --- targeted SFT on 60B tokens with layout pool constrained to 7 Pareto optimal presets (listed in Table~\ref{tab:presets}) identified on the post distillation checkpoint using procedure outlines in Section~\ref{sec:placement} + the \texttt{all-FA} placement. Notably, some of these runs are marked with \texttt{SFT$^*$}, where the ``*'' marks a version of the SFT dataset from Table~\ref{tab:data-mix} focused on math and reasoning traces with tool use category excluded.

\section{Placement Optimization}
\label{sec:placement}

\paragraph{Problem.}
Given a supernet with $L = 48$ layers and $\abs{\mathcal{M}} = 4$ mixer types, a
placement $\mathbf{x} = (x_1, \ldots, x_L)$ assigns a mixer to each
layer.  The size of the search space is $4^{48} \approx 7.9 \times 10^{28}$.  The
optimization problem is:
\[
  \mathbf{x}^* = \arg\max_{\mathbf{x} \in \mathcal{M}^L} \; S(\mathbf{x})
  \quad \text{subject to} \quad
  C(\mathbf{x}) \leq B
\]
where $S(\mathbf{x})$ is a quality score from the supernet, $C(\mathbf{x})$ an inference cost, and $B$ is a
cost budget.  Each evaluation of $S$ runs a benchmark suite on the supernet, individual evaluations take seconds, but the search space makes exhaustive enumeration infeasible.

\paragraph{Cost model.}
We use a linear model to represent the inference cost of mixer allocations.
Empirical throughput measurements show that we can model per-token-latency (or the inverse of the throughput) as a linear function of the number of each mixer-type.
Exceptions to this linear fit are singleton placements, which we ignore during the refinement phase (see Appendix~\ref{app:cost_models}).
The cost of a placement $\mathbf{x}$ is expressed as:
\[
C(\mathbf{x}) = \sum_{i=1}^{L} c(x_i)
\]
In this manuscript we report results for two  variants of this cost model. In our preliminary experiments we explored the \texttt{Idealized} cost model. It measures throughput on pure placements (all 48 layers of a single mixer type) and assumes each layer contributes independently: $c_X = \text{throughput}_{\text{FA}} / \text{throughput}_X$. A more principled cost model fits the above equation to 84 mixed placements via ordinary least squares \texttt{Regression}. Regression costs are consistently higher than idealized costs (Table~\ref{tab:mixer_costs}), reflecting this mixing overhead. We report results from placements optimized under both cost models; full details are in Appendix~\ref{app:cost_models}. As per Table~\ref{tab:mixer_costs}, throughout the main text of this report we use \texttt{Idealized}($w$ = 2048)\footnote{We note that we apply this cost model to perform surrogate guided search to a supernet with SWA with $w$=4096.} and \texttt{Regression} (clean, $w$=4096) cost models for surrogate-guided search, where $w$ denotes the window size of the used SWA mixer.

\paragraph{Cluster expansion.}
We decompose the score function using the cluster expansion from
statistical physics~\citep{sanchez1984cluster,defontaine1994cluster}:
layers correspond to lattice sites, mixer types to atomic species, and
the supernet score to the energy.  The expansion expresses $S$ as a sum
of $n$-body interactions:
\[
  S(\mathbf{x}) = \sum_i V^{(1)}_i(x_i)
  + \sum_{i < j} V^{(2)}_{ij}(x_i, x_j)
  + \cdots
\]
In practice, each potential corresponds to a learned coefficient on an indicator feature: $V^{(1)}_i(m)$ is the coefficient for ``mixer $m$ at layer $i$'', and $V^{(2)}_{ij}(m, m')$ is the coefficient for ``mixer $m$ at layer $i$ and mixer $m'$ at layer $j$''.

Two hyperparameters control the expansion's expressiveness:
\begin{itemize}
  \item \textbf{Order}: the number of layers that interact in each term.  Order~1 (unary) captures per-layer effects; order~2 (pairwise) captures how pairs of mixer choices interact; order~3 captures triplet interactions.
  \item \textbf{Range}: the maximum distance between interacting layers.  Range~1 restricts pairwise features to adjacent layers $(i, i{+}1)$; range~2 adds layer pairs up to two apart $(i, i{+}2)$, and so on.
\end{itemize}
Assuming that layer interactions decay geometrically with distance through the residual
stream, justifies truncation to low order ($\leq 3$) and short range ($\leq 3$).  Feature counts at each expansion order are given in
Appendix~\ref{app:search}.

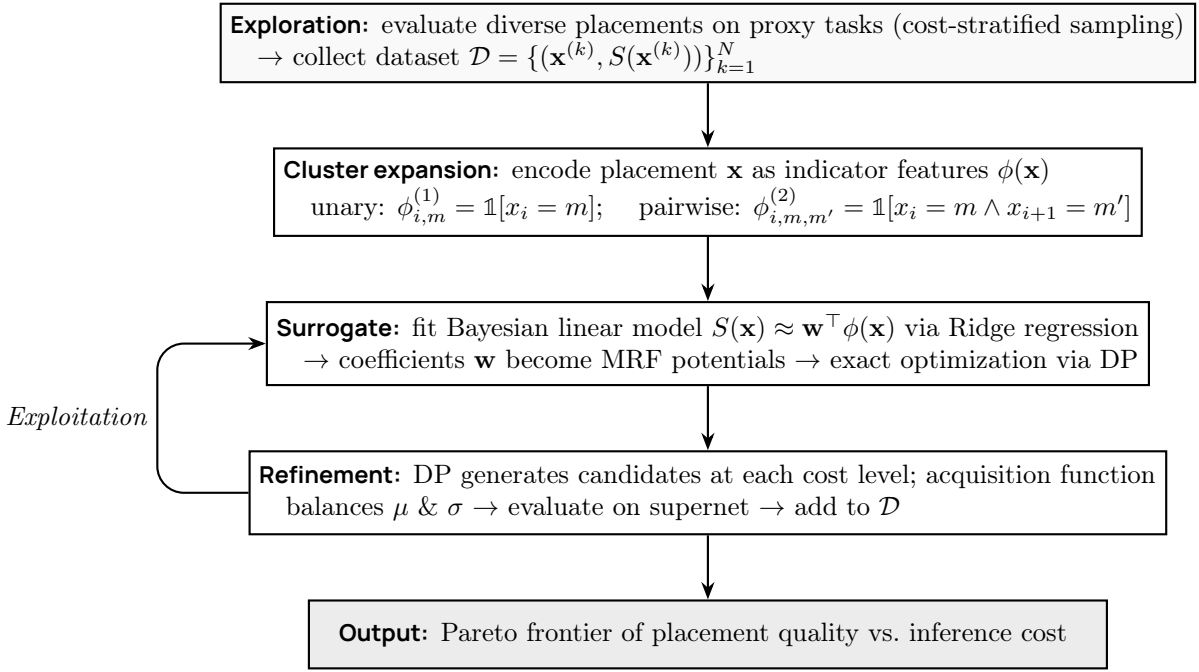
\begin{figure}[t]
\centering
\begin{tikzpicture}[
    node distance=0.85cm,
    box/.style={
        rectangle,
        draw=black,
        thick,
        minimum width=10.5cm,
        minimum height=1.1cm,
        align=left,
        font=\small
    },
    initbox/.style={
        rectangle,
        draw=black,
        thick,
        minimum width=10.5cm,
        minimum height=0.9cm,
        align=left,
        font=\small,
        fill=gray!5
    },
    outputbox/.style={
        rectangle,
        draw=black,
        thick,
        minimum width=10.5cm,
        minimum height=0.9cm,
        align=left,
        font=\small,
        fill=gray!15
    },
    arrow/.style={
        ->,
        >=Stealth,
        thick
    },
    looparrow/.style={
        ->,
        >=Stealth,
        thick,
        rounded corners=10pt
    }
]

\node[initbox] (step0) {
    \textbf{Exploration:} evaluate diverse placements on proxy tasks (cost-stratified sampling)\\
    \hspace{1em}$\rightarrow$ collect dataset $\mathcal{D} = \{(\mathbf{x}^{(k)}, S(\mathbf{x}^{(k)}))\}_{k=1}^{N}$
};

\node[box, below=of step0] (step1) {
    \textbf{Cluster expansion:} encode placement $\mathbf{x}$ as indicator features $\phi(\mathbf{x})$\\
    \hspace{1em}unary: $\phi^{(1)}_{i,m} = \mathbb{1}[x_i = m]$; \quad pairwise: $\phi^{(2)}_{i,m,m'} = \mathbb{1}[x_i = m \land x_{i+1} = m']$
};

\node[box, below=of step1] (step2) {
    \textbf{Surrogate:} fit Bayesian linear model $S(\mathbf{x}) \approx \mathbf{w}^\top \phi(\mathbf{x})$ via Ridge regression\\
    \hspace{1em}$\rightarrow$ coefficients $\mathbf{w}$ become MRF potentials $\rightarrow$ exact optimization via DP
};

\node[box, below=of step2] (step3) {
    \textbf{Refinement:} DP generates candidates at each cost level; acquisition function\\
    \hspace{1em}balances $\mu$ \& $\sigma$ $\rightarrow$ evaluate on supernet $\rightarrow$ add to $\mathcal{D}$
};

\node[outputbox, below=of step3] (output) {
    \textbf{Output:} Pareto frontier of placement quality vs.\ inference cost
};

\draw[arrow] (step0) -- (step1);
\draw[arrow] (step1) -- (step2);
\draw[arrow] (step2) -- (step3);
\draw[arrow] (step3) -- (output);

\draw[looparrow]
    (step3.west) -- ++(-1.2,0)
    |- node[pos=0.25, left, align=center, font=\small\itshape] {Exploitation}
    (step2.west);

\end{tikzpicture}
\caption{Surrogate-guided placement optimization (Section~\ref{sec:placement}). We collect a dataset of (placement, score) pairs by evaluating cost-stratified placements on proxy tasks. Each placement is encoded as indicator features: unary features for each (layer, mixer) assignment and pairwise features for adjacent (layer, layer+1, mixer, mixer) combinations. A Bayesian linear model (Ridge regression) maps features to predicted scores; the learned coefficients become MRF potentials enabling exact cost-constrained optimization via dynamic programming. An acquisition function balances predicted quality ($\mu$) with uncertainty ($\sigma$) to select candidates for real evaluation. The surrogate is iteratively refined, yielding a Pareto frontier of placement quality vs.\ inference cost.}
\label{fig:surrogate_search}
\end{figure}

\paragraph{Surrogate.}
We fit a Bayesian linear model on the cluster-expansion features.
Because the features decompose over local layer tuples (individual
layers, adjacent pairs, and optionally contiguous triplets), the model
assigns one learned coefficient to each feature.  Under the Bayesian
formulation, these are posterior-mean estimates that double as Markov
random field (MRF) potentials; the predicted score for any placement is
simply their sum.  Each potential involves only a small contiguous group
of layers, so the optimal placement can be found exactly by dynamic
programming---processing layers sequentially while tracking a bounded
window of recent assignments.  Given any additive per-layer cost model,
the optimum at every cost level can be found in a single forward pass
(seconds on CPU), covering all feasible allocations simultaneously.
A nonlinear surrogate could approximate the score landscape more
flexibly, but would forfeit exact optimization, reducing the search to a
heuristic over the surrogate's predictions.  The formulation is in
Appendix~\ref{app:search}.

\paragraph{Adaptive expansion selection.}
The expansion can in principle include arbitrarily high-order
interactions, but each order introduces exponentially more parameters
(Appendix~\ref{app:search}, Table~\ref{tab:features}).  A feature guard
enforces a minimum 2:1 sample-to-feature ratio.  Among
admitted configurations, we select the expansion order and interaction
range by maximizing Bayesian model evidence, which penalizes
unnecessary complexity.  The selection is re-run as the adaptive loop
collects more data: once enough samples accumulate to pass the feature
guard at a higher order, that order becomes eligible and is adopted if
the evidence supports it.  The selected order is also diagnostic: it
reveals whether the supernet's quality landscape is governed by
per-layer effects alone or by multi-layer interactions
(Appendix~\ref{app:search}).

\paragraph{Refinement.}
The surrogate is iteratively refined.  At each round, the
dynamic-programming pass generates candidates at each cost level from
the posterior mean; an acquisition function selects those that balance
predicted quality with uncertainty, and the selected candidates are
evaluated on the real supernet and added to the training set.  A few
rounds suffice (details in Appendix~\ref{app:search}). A schematic representation of the full process is depicted in Figure~\ref{fig:surrogate_search}.

\paragraph{Pareto frontier.}
Because the dynamic-programming table stores optimal sub-placements at each cumulative cost, a single pass can extract the optimal placement at every cost level, yielding a complete Pareto frontier of placement quality vs.\ inference cost. Recovering the full frontier requires evaluating candidates at each cost level, which is the default in the exploration phase. When only a small number of operating points is needed, the search can be restricted to the target costs, reducing the evaluation budget in the exploitation phase (the loop in Figure~\ref{fig:surrogate_search}). In practice, the frontiers depicted in Figures~\ref{fig:pareto_combined} and \ref{fig:pareto_frontier_prelim} are obtained by fitting a single surrogate on combined data from exploration run and multiple cost-targeted exploration runs (one per target preset in Table~\ref{tab:presets}) and running DP-solver in parallel for each cost band.

\section{Placement Landscape Dynamics}
\label{sec:landscape_dynamics}

The supernet's multi-mixer architecture
creates a dynamic quality landscape in which placement rankings---their ordering
by score at a given checkpoint---can shift during training, a phenomenon we call
\emph{landscape drift}.  Each placement's quality depends on the performance of its
constituent mixers, and each mixer's performance evolves as it learns from the
teacher and adapts to the shared representation.
Because mixers have different computational signatures and learn at
different rates, the relative quality of placements
can change as training progresses, leading to landscape drift.
Such drift has structure tied to mixer composition and training convergence dynamics,
which we investigate empirically in this section.

Ranking stability has direct implications for training strategy and deployment outcomes.

If rankings are stable, placement optimization and training decouple.
The ranking measured at an early checkpoint identifies the
Pareto-optimal placements at each cost level for all of training and beyond,
so the training strategy can be designed around that knowledge.
Targeted training (of a single or multiple placements) can then focus on the Pareto-optimal placements from the start,
accelerating their improvement and delivering better deployment candidates for
the same compute budget. By contrast, uniform sampling
spends compute on placements that will never ship,
so it is less efficient at improving the configurations that matter most for deployment.

If rankings are unstable, however, early measurements of placement ranking are misleading and any
commitment made on their basis is wrong.
Placement optimization must therefore wait until training ends, when the final rankings are revealed.
During training, the strategy must be agnostic to placement quality, so uniform placement sampling is the only safe choice.

The more interesting case lies between these extremes, where
rankings are partially stable, and the Pareto frontier is identifiable before training ends.
The question is then not whether to decouple placement optimization and training, but when.
Intuitively, convergence of training dynamics should stabilize the landscape, so rank stability should also improve over time.
The Pareto-optimal placements should emerge as the clear winners in their cost tiers before the end of training.
In this scenario, the training strategy can adapt to the evolving landscape,
starting with uniform sampling to allow the rankings to stabilize,
and switching to targeted sampling once the Pareto-optimal placements can be reliably identified,
concentrating training signal on the configurations most likely to be deployed and improving them faster.
But determining when that threshold has been crossed requires a direct measure of rank stability during training.

However, if placement rankings are measured too early while the landscape is still drifting,
the risk of \emph{false optima} arises:
If targeted training is applied to placements that appear optimal at an early checkpoint
but are not truly optimal had training with uniform sampling progressed further,
then we have a risk of deploying suboptimal placements.
Such false optima appear optimal only because they are being trained, not because the landscape independently favors them.
This failure mode is subtle because the same signal that creates false optima also confirms them:
once targeted training commits gradient signal to the selected configurations,
the process becomes self-reinforcing.
The selected configurations improve, and their rankings shift further in their favor.
Re-evaluation will then confirm their superiority, but that confirmation is circular.
Whether the identified placements are genuine optima or false ones
is not falsifiable from targeted training alone,
because training and evaluation are coupled:
the rankings are no longer an independent signal of placement quality,
but a reflection of the training strategy's bias.
The true Pareto frontier remains unknown.

Null results are as informative as positive findings: rank stability under
all regimes would mean placement search is robust to training strategy, an
operationally useful result that simplifies deployment.
These considerations motivate four testable hypotheses:
\begin{enumerate}
  \item[\textbf{H1}] \textbf{Landscape drift.}
    Placement rankings shift measurably during training even under uniform
    stochastic sampling.
  \item[\textbf{H2}] \textbf{Frontier fragility.}
    Pareto-frontier placements exhibit higher rank volatility than median
    placements across all training regimes, as a consequence of their
    geometric isolation in the score distribution.
  \item[\textbf{H3}] \textbf{Targeted training.}
    Targeted training widens the performance gap between trained and canary
    placements beyond the stochastic reference.
  \item[\textbf{H4}] \textbf{Hybrid stability.}
    A mixed strategy---cycling through Pareto-optimal placements with
    probability~$p$ and sampling uniformly with probability~$1{-}p$---preserves
    rank integrity while concentrating improvement toward deployment
    candidates, with $p$~controlling the tradeoff between training
    efficiency and protection against false optima.
\end{enumerate}
Stochastic and targeted steps serve complementary roles: the former
improves the shared representations by training all mixer combinations, while the latter directs optimization toward
specific shipping candidates.

\subsection{Core Ablation: Training Strategy and Rank Stability}
\label{sec:landscape_dynamics:ablation}

To study Supernet training dynamics at a tractable scale, we construct a 0.5B
development supernet (Appendix~\ref{app:dev_model}) that replicates the
\Super~Apriel mixer vocabulary on a smaller base model.
We compare those results to the landscape drift observed on the 15B stochastic distillation run to confirm whether the patterns transfer to scale.
Three training regimes are applied to the 0.5B model on identical data, with the same number of
steps and hyperparameters (see Table~\ref{tab:training}).  The only variable is
the placement sampling distribution (Table~\ref{tab:training_regime}).
The stochastic regime is the control condition: it establishes whether
the landscape drifts at all~(H1) and whether that drift concentrates at
the frontier~(H2).
The targeted and hybrid regimes test H3 and H4 against this baseline.

\begin{table}[h]
\centering
\caption{Placement sampling for each training regime.}
\small
\begin{tabular}{@{}lp{4.5cm}p{7cm}@{}}
\toprule
\textbf{Regime} & \textbf{Sampling rule} & \textbf{Purpose} \\
\midrule
Full stochastic & Uniform random placement each step (local sampling) & Control: shared parameters receive unbiased gradient signal \\
Targeted & Cycle through $k$ shipping presets &
Tests H3: Optimize preset placements for deployment, assuming optimal presets have been found and potentially degrading canary placements.  \\
Weighted hybrid & Targeted with probability $p$, stochastic with probability $1-p$ &
Tests H4: whether partial targeting preserves landscape integrity while improving shipping candidates \\
\bottomrule
\end{tabular}
\label{tab:training_regime}
\end{table}

At each intermediate checkpoint, we evaluate a fixed set of $1{,}000$ random canary placements. These placements serve as a probe to analyze how placement rankings change during distillation.
In this section, placements are scored using the average log-likelihood on math traces as defined in Section~\ref{sec:eval:suite}.
For a given checkpoint, the frontier is defined as the set of placements that are optimal for their cost category.
We examine how placement rankings evolve globally, and per cost bands.

\subsection{Results}
\label{sec:landscape_dynamics:results}

\subsubsection{Stochastic training}
\label{sec:landscape_dynamics:results:stochastic}

\begin{figure}[t]
  \centering
  \begin{subfigure}[t]{0.48\textwidth}
    \centering
    \includegraphics[width=\linewidth]{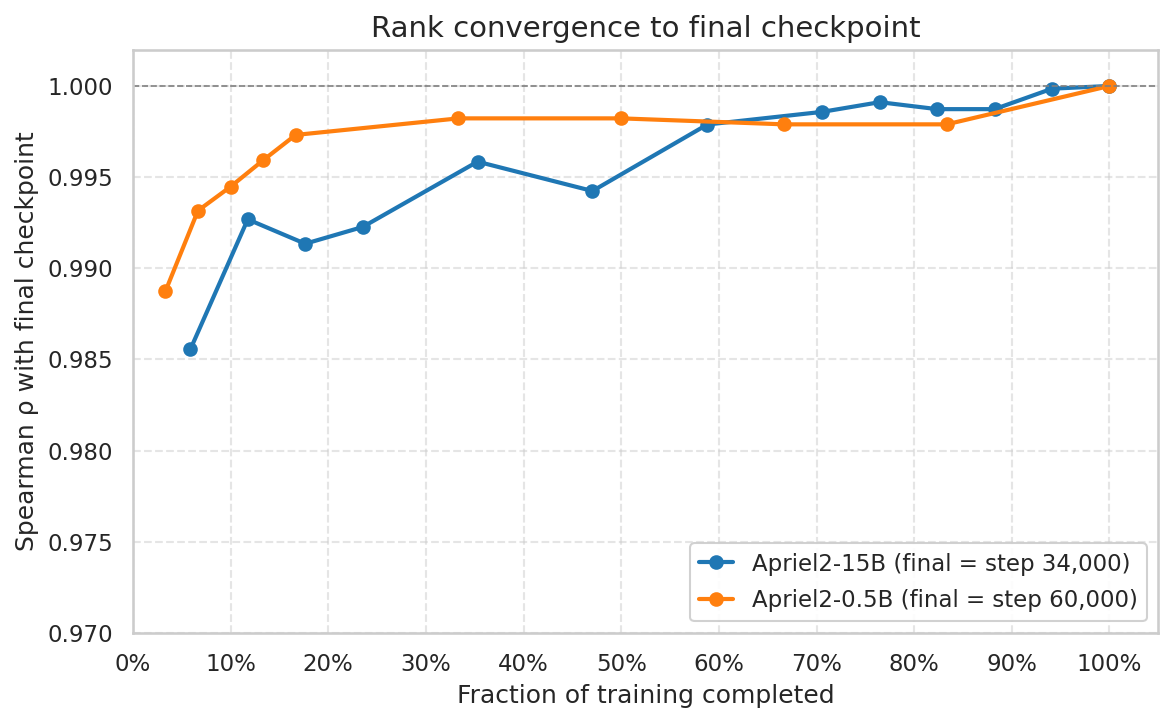}
    \caption{Convergence to final ranking.}
    \label{fig:rho_convergence:main}
  \end{subfigure}
  \hfill
  \begin{subfigure}[t]{0.48\textwidth}
    \centering
    \includegraphics[width=\linewidth]{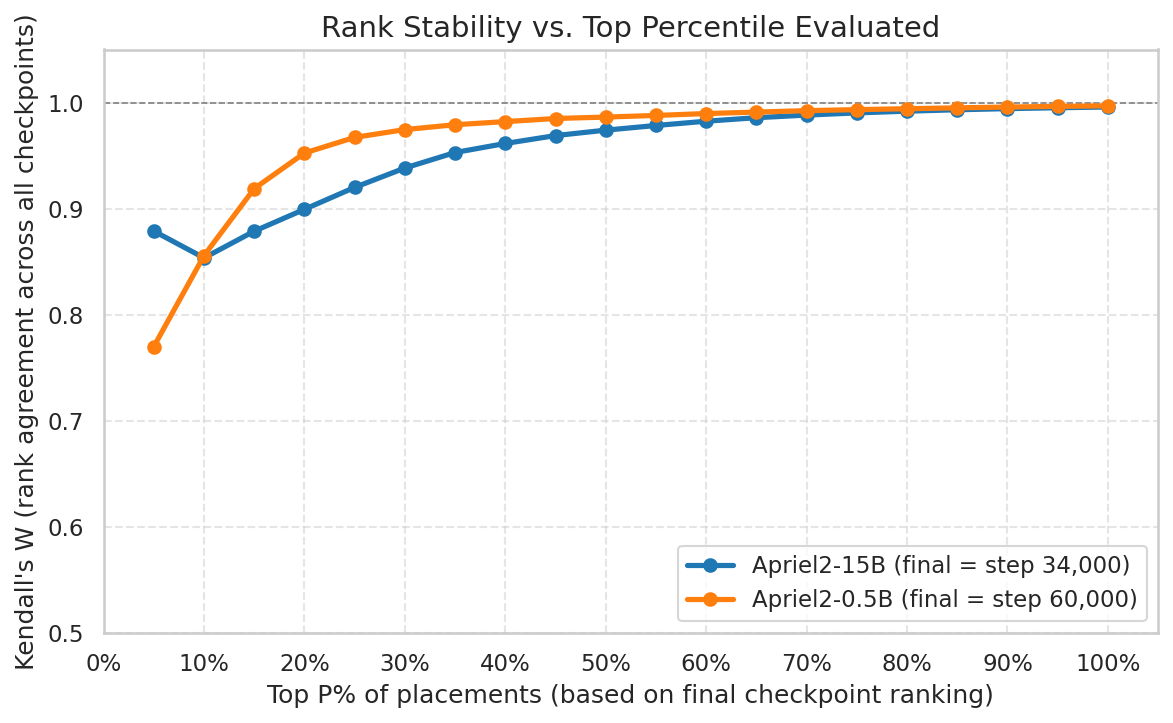}
    \caption{Multi-rater concordance as a function of the percentile tier of placements included.}
    \label{fig:rho_convergence:kendall}
  \end{subfigure}
  \caption{
    Rank stability under stochastic distillation at two scales.
    \textbf{Panel~(a):} Spearman $\rho$ between each intermediate checkpoint and
    the final checkpoint (15B: step~34K; 0.5B: step~60K).  Both scales
    show $\rho > 0.98$ from the earliest measured checkpoint,
    indicating that rankings are effectively determined within the
    first few percent of training.
    \textbf{Panel~(b):} Kendall's $W$ (multi-rater concordance across all distillation
    checkpoints) as a function of the percentile tier of placements
    included.  As the subset is restricted to higher-scoring
    placements, concordance decreases monotonically.  The top-5\%
    tier---corresponding to frontier placements most relevant for
    deployment---shows $W{=}0.77$ for the 0.5B model, indicating substantial but
    imperfect agreement.  \textbf{This confirms H2}: rank stability is lowest
    precisely where it matters most.
  }
  \label{fig:rho_convergence}
\end{figure}

\begin{figure}
    \centering

    \begin{subfigure}{\linewidth}
        \centering
        \includegraphics[width=\linewidth]{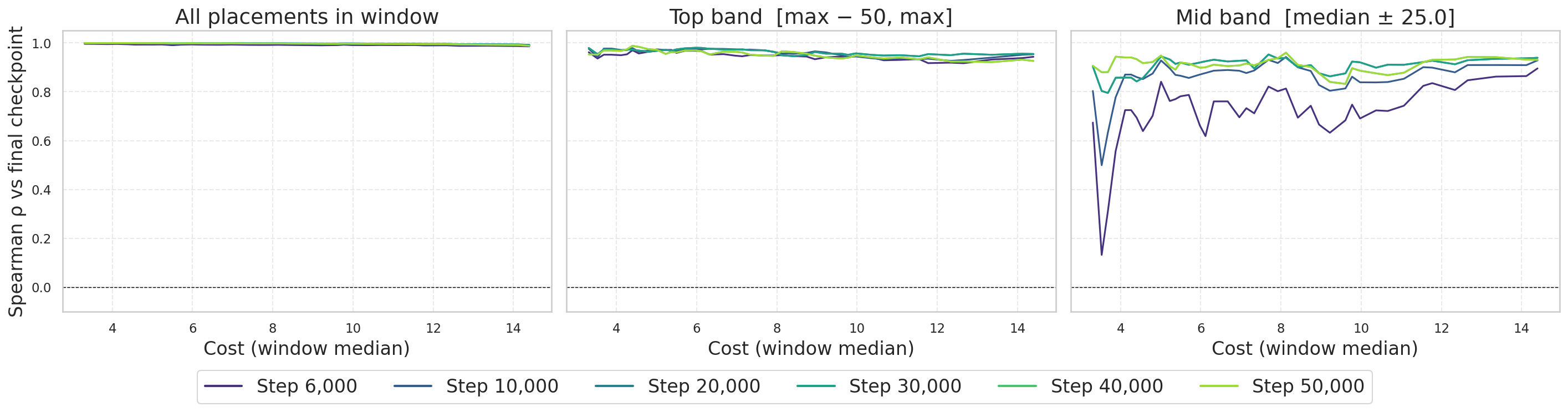}
        \caption{0.5B model}
        \label{fig:rank_stability_05b}
    \end{subfigure}

    \begin{subfigure}{\linewidth}
        \centering
        \includegraphics[width=\linewidth]{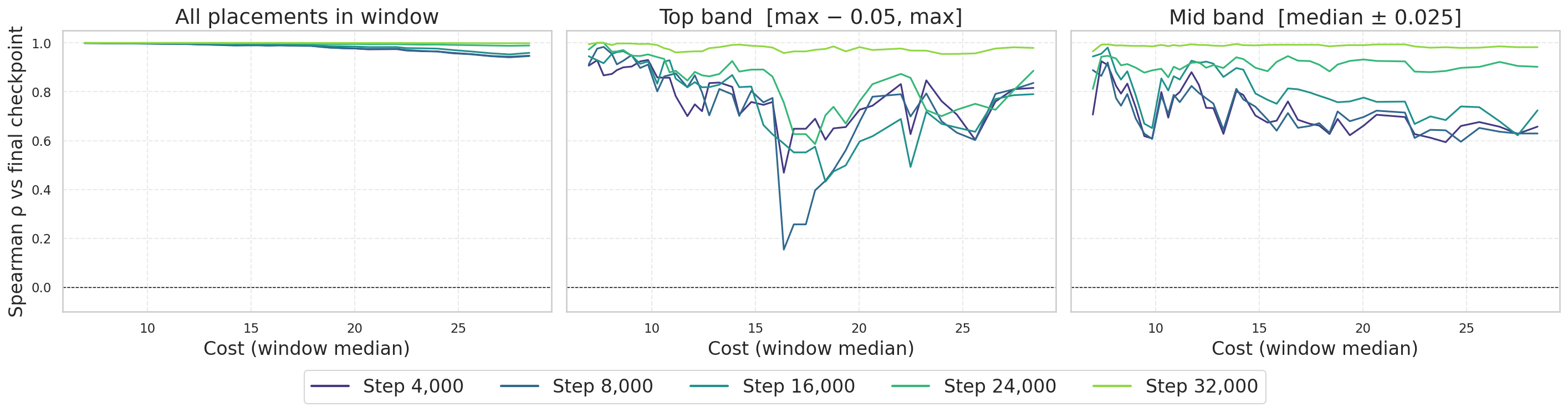}
        \caption{15B model}
        \label{fig:rank_stability_15b}
    \end{subfigure}

    \caption{
    Rank-stability measure with Spearman-rho at 0.5B (top) and 15B (bottom) for rolling cost-windows of 200 placements during stochastic distillation. \textbf{Left:}~Ranks stability of intermediate checkpoints vs final checkpoint considering all placements in window.  \textbf{Middle:}~Considering placements within 50 pts (0.5B) or 0.05 pts (15B) of the top score in window according to final checkpoint, i.e the frontier. \textbf{Right:}~Considering placements within 25 pts (0.5B) or 0.025 pts (15B) of the median score in window. Placements are very stable overall (left).
    The 0.5B model shows stability at the frontier, more than in the bulk (top middle and right).
    When zooming-in on narrow score bands of the 15B model (bottom middle and right), the rank-agreement goes down. The frontier exhibits higher levels of instability than the bulk, especially for medium-cost placements.
    }
    \label{fig:rank_stability_window}
\end{figure}

We first analyze placement rankings at a coarse scale.
For both model sizes, Spearman $\rho$ between scores in early checkpoints and the final checkpoint exceeds 0.98 from the earliest
measured checkpoint (Figure~\ref{fig:rho_convergence:main}).
Kendall's $W$ measures the multi-rater concordance between all intermediate checkpoints (Figure~\ref{fig:rho_convergence:kendall}). We consider different percentile-tiers of placements, based on their score on the final checkpoint.
Kendall's $W$ drops monotonically from ${\sim}1.0$ (all placements) to $0.77$ (top 5\%) at 0.5B and $0.88$ at 15B.
These results show that the overall landscape is stable during stochastic training, for both 0.5B and 15B models.

A more fine-grained analysis compares stability in the bulk of the distribution and at the Pareto frontier.
To do so, we consider rolling cost-windows of 200 placements. For each cost-window, we compare what happens between placements close to the top, the median, or all placements.
Figure~\ref{fig:rank_stability_window} shows Spearman $\rho$ across cost-windows for all placements (left), placements on the frontier (middle), or placements with median performance at different checkpoints of stochastic distillation.
Overall placement rankings are stable in all cost-windows, confirming the previous discussion of Figure~\ref{fig:rho_convergence}.
At the 0.5B scale, the frontier is also stable. The median-band is where most placements concentrate and even small score variations create rank instability, explaining the higher volatility.
At the 15B scale, the frontier shows more instability, especially for medium-cost placements that matter the most for deployment.
Early checkpoints up to iteration 24K show low-correlation with the final checkpoint between costs 16 and 23.
For the 15B model, contrary to the 0.5B, rankings fluctuate highly along the frontier.
These results show that learnings from smaller-scale experiments do not transfer well to larger scale. Experiments at a larger scale are necessary to make optimal decisions.

\begin{figure}
    \centering
    \begin{subfigure}{\linewidth}
        \includegraphics[width=\linewidth]{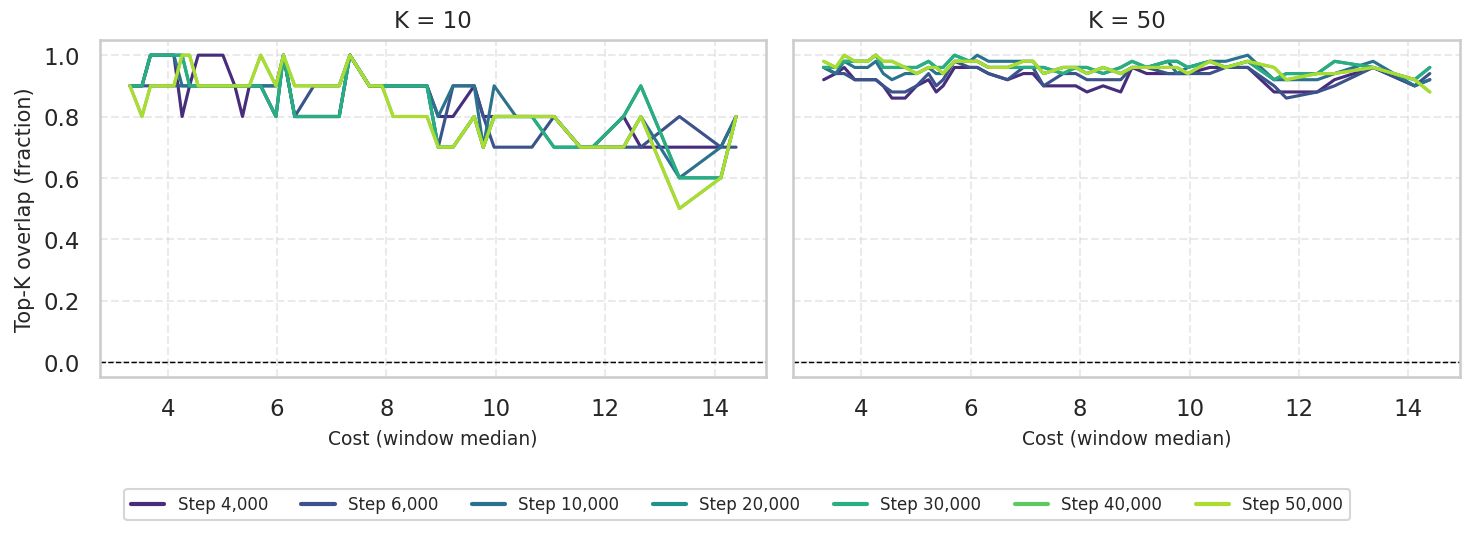}
    \end{subfigure}
    \begin{subfigure}{\linewidth}
        \includegraphics[width=\linewidth]{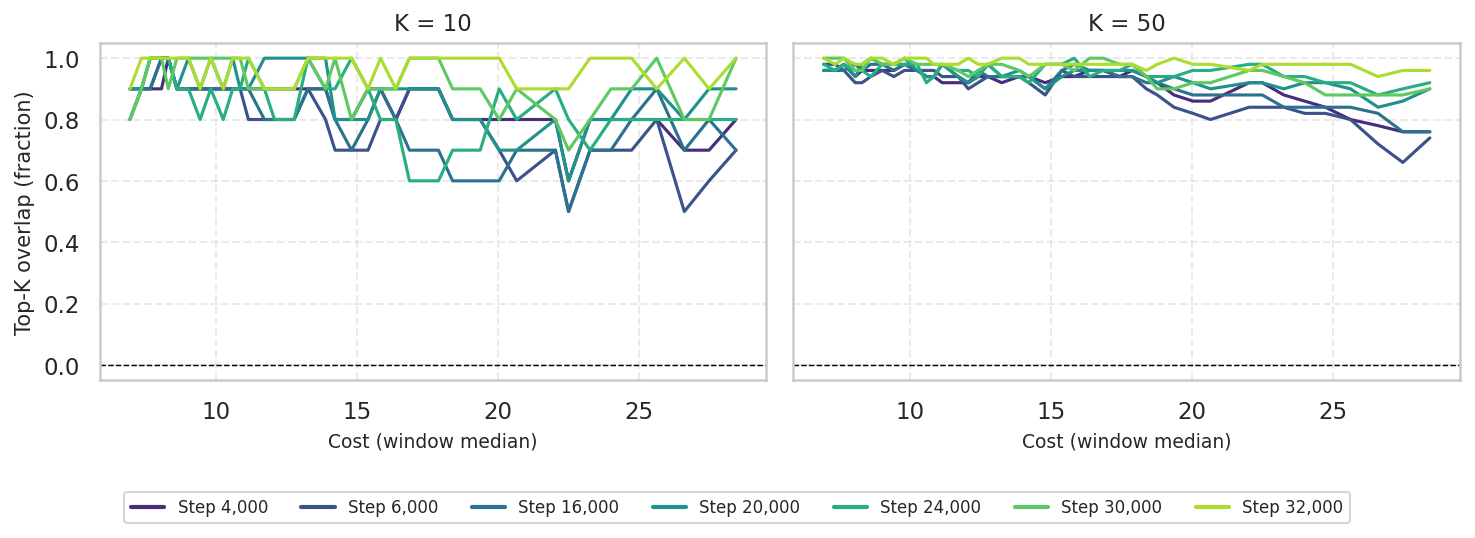}
    \end{subfigure}
    \caption{Top-k overlap fraction between early checkpoints and final checkpoints at 0.5B (top) and 15B (bottom) for rolling cost windows of 200 placements.}
    \label{fig:top_k_stability}
\end{figure}

Figure~\ref{fig:top_k_stability} shows that top-k overlap fraction between early checkpoints and the final checkpoint is close to 1 for $k=50$, and larger than $0.5$ for $k=10$ for both models. This shows that across all cost-bands, the potential drift of top-performing placements is limited, since most top-k placements for an early checkpoint stay in the top-k for the final checkpoint.

These results \textbf{nullify H1}: on a coarse scale, placements rankings are very stable throughout stochastic distillation.
\textbf{H2 is partially confirmed}: the 15B model exhibits substantial instability among frontier placements, particularly for medium-cost configurations---precisely those most relevant for deployment. Consequently, placements selected early in training may differ from those selected at convergence. However, early-selected placements do not drift far from the frontier, suggesting that targeted training remains viable even with early selection.

\subsubsection{Targeted training}
\label{sec:landscape_dynamics:results:targeted}

\begin{figure}
    \centering
    \includegraphics[width=\linewidth]{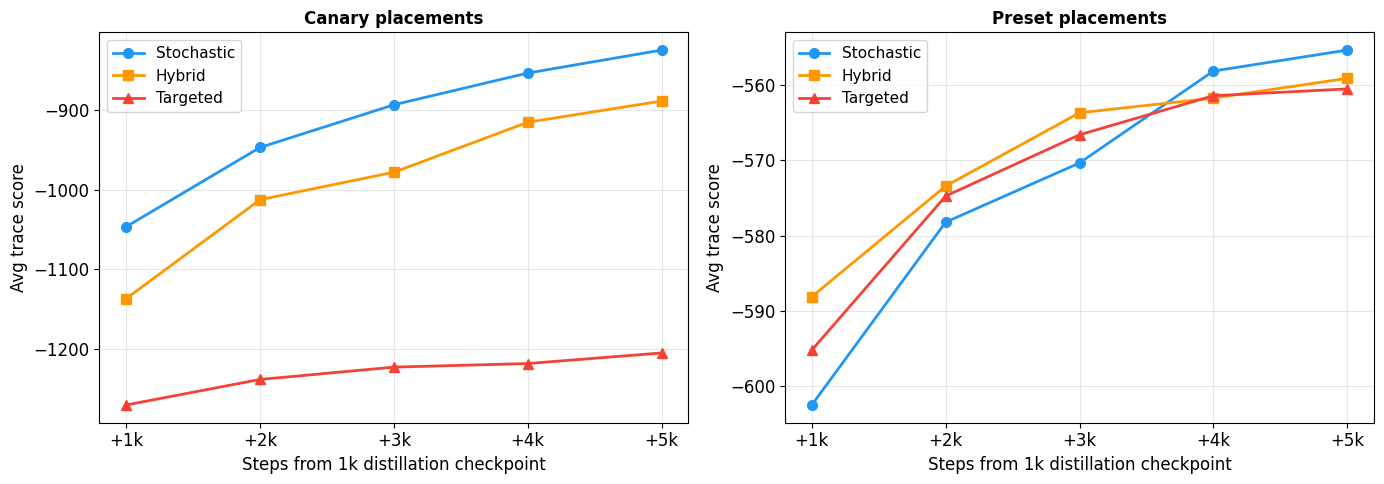}
    \caption{Trace scores during continued distillation of the 0.5B models with three training regimes: stochastic (blue), hybrid (orange), and targeted (red). \textbf{Left:}~Scores on random canary placements. \textbf{Right:}~Scores on preset placements. The scores are the average log-likelihood on Math500.}
    \label{fig:training_regime_ablation}
\end{figure}

We perform training regime ablations using the 0.5B model.
Using the procedure described in Section~\ref{sec:eval:presets} (adapted to 0.5B model here), we obtain placement presets optimized for math-reasoning tasks.
We then continue distillation with three different regimes described in Section~\ref{sec:landscape_dynamics:ablation}: stochastic, targeted (cycle through the presets), or hybrid (cycle through the presets with probability $p=0.5$, and stochastic with probability $1-p$).
We observe how the performance of random canary placements and preset placements evolves during distillation with each regime.
Figure~\ref{fig:training_regime_ablation} (left) shows that canary placements improve the most under stochastic training. Hybrid training samples canaries with less frequency than stochastic training, which translates to slower improvement of the canaries.
Targeted training never samples canary placements. The canaries still show slight improvement, possibly because they may have mixers in common with the preset placements.
The preset placements learn faster in hybrid and targeted regimes at the beginning of training (right).
Interestingly, the stochastic training catches up, ending up with a higher score on preset placements.

At the 0.5B scale, \textbf{H3 is confirmed and H4 is nullified}. Stochastic distillation brings the best of both worlds by consistently improving all placements. Focusing the training on a few placement presets---via a targeted or hybrid regime---slows-down training of the majority of placements that are not the presets, and doesn't necessarily lead to better performance of the presets on the 0.5B scale.
Because comparisons between the 0.5B and the 15B model in Section~\ref{sec:landscape_dynamics:results:stochastic} showed that learnings at a small-scale do not necessarily translate to larger-scale models, we conclude that similar experiments on the 15B model will be necessary to confirm or refute these findings.

\subsubsection{Summary}

The results characterize the landscape drift during stochastic training at two scales---the 0.5B development supernet and the 15B production supernet---and establish the baseline against which target and hybrid regimes are measured.
Across the full placements set, H1 is nullified: rankings crystallize almost immediately. The 0.5B landscape does not confirm H2, but the 15B landscape shows higher rank-instability at the frontier than in the bulk, particularly on medium-cost placements.
There is a risk that placements identified as presets on an early checkpoint would not match the actual frontier on the final checkpoint.
Early-selected frontier placements drift but stay close to the final frontier, showing that the risk of selecting early is mild.
The behavior of the 0.5B and 15B models is vastly different on this aspect, since the 0.5B frontier is much more stable. Findings on smaller models must be verified and do not automatically transfer to larger models.

Results with targeted and hybrid training on the 0.5B model confirm H3 and nullify H4. On this model, stochastic training improves all placements consistently: even preset placements' performance catches up with targeted training.
These findings on small-scale models favor stochastic training during distillation. By training all placements, we avoid the risk of false optima and obtain the best frontier at the end of distillation.
Possible future works will repeat these experiments at larger scale to confirm those results. For the 15B \Super~Apriel, full distillation phase was performed using fully stochastic local sampling. As described in Section~\ref{sec:training:sft}, we opt for the targeted regime for the SFT phase of the 15B \Super~Apriel.

\section{Inference}
\label{sec:inference}

The supernet is served via a vLLM plugin implemented in
Fast-LLM~\citep{fastllm} with the following serving modes: \texttt{single-preset} mode --- only the weights of a single selected mixer placement are loaded and executed using CUDA graphs (FULL\_AND\_PIECEWISE); \texttt{supernet} mode --- full supernet is loaded into memory and, as detailed below, placement switching can be supported at a per-request level, enabling a multi-deployment scenario.
Unless stated otherwise, the \texttt{single-preset} mode is used for throughput benchmarking and cost model tuning.

\paragraph{Heterogeneous state.}
FA and SWA layers maintain standard KV caches (growing with sequence
length for FA, bounded at $w = 4096$ for SWA).  GDN layers maintain a
fixed-size recurrent state
$\mathbf{S} \in \mathbb{R}^{h_k \times d_k \times d_v}$ plus a short
convolution buffer.  KDA layers maintain a similar recurrent state plus
three convolution buffers (one per projection stream).  A unified
page-size architecture pads all per-layer state allocations to a common
page size, so that the paged memory manager handles heterogeneous
layers transparently.

\paragraph{Prefill and decode.}
GDN and KDA support dual execution paths.  Prefill uses chunkwise
parallel algorithms from the FLA (Flash Linear Attention) library for
hardware-efficient tensor-core utilization.  Decode uses fused recurrent
kernels that update the state in-place with incremental convolution
updates.

\paragraph{Preset switching.} Enables \texttt{supernet} serving mode.
A supernet checkpoint contains trained weights for all four mixer types
at every layer. Switching presets does not require reloading from disk; only the
mixer blocks change while shared parameters (FFNs, embeddings,
normalization) remain in place.  A switch transfers the required mixer
weights between CPU and GPU and recaptures CUDA graphs, typically taking
5--15\,s depending on how many layers change mixer type. This is reasonable when placement switching must happen per-task.

\paragraph{Per-request placement switching.} A per-request placement routing solution is \emph{under development}, which extends vLLM 0.18.0 to let a single loaded supernet checkpoint serve requests with different per-layer mixer architectures, selected on a per-request basis. The design draws inspiration from vLLM's LoRA serving, which also enables per-request model variants from a shared base, but operates at the architectural level rather than the weight level -- swapping which mixer module runs at each layer.

A \texttt{SupernetConfig} declares which placements to support and their layer-by-layer mixer mappings. At request time, clients pass a \texttt{placement\_id}, which propagates through the input processor, engine request, and scheduler all the way to the GPU worker. Because the entire compute graph changes per placement, the scheduler enforces same-placement batching: all requests in a given scheduling round must share the same \texttt{placement\_id}, with mismatched requests deferred to the next iteration.
During startup, CUDA graphs are pre-captured for allowed placements, and at runtime the model runner selects the correct graph.

Inside the model itself, a stochastic mixer module holds all allowed mixers per layer with their full weights loaded; during forward, it reads the current \texttt{placement\_id} from the forward context, looks up the \texttt{SupernetConfig} to determine which mixer to run at each layer index, and dispatches accordingly -- no global mutable state, making it safe for concurrent serving. The net effect is that a single GPU-resident checkpoint can behave like many different architecture variants (e.g., all-attention, alternating attention/GDN, etc.) without duplicating weights, at the cost of pre-capturing a set of graphs per supported placement.

\paragraph{Multi-deployment.}
A single \texttt{supernet}  serving instance supports multiple operating points.  The
\texttt{all-FA} preset performs on par with the teacher.  Hybrid presets trade
quality for throughput along the Pareto frontier discovered by the
placement optimizer.  Because the hybrid shares a checkpoint with the \texttt{all-FA} baseline,
it can be deployed as an additional serving mode rather than a separate
model.

\paragraph{Throughput considerations.}
In \texttt{single-preset} mode, inactive mixer weights are entirely offloaded to CPU.
Serving multiple placements concurrently
requires keeping all registered mixer weights on GPU, reducing KV cache
capacity and maximum batch size. In particular, a set of placements with a larger number of distinct mixers per layer (i.e., lower overlap between placements) will yield lower throughput overall as more mixer weights occupy available GPU memory. Mitigations are discussed in
Section~\ref{sec:outlook:deployment}.

\section{Evaluation}
\label{sec:eval}

\subsection{Evaluation Suite}
\label{sec:eval:suite}

We evaluate placement quality on a suite spanning knowledge, mathematical reasoning, information extraction and tool calling. We mark a set of benchmarks that we use during model development as \emph{dev benchmarks} --- during the model development we use all or a subset of these benchmarks to run ablations to make decisions about model development including hyperparameters, layer freezing, etc. Some of these benchmarks are also used as targets for placement optimization as discussed below. We include a set of \emph{unseen benchmarks} in the final evaluations. These benchmarks are never used during the model development and are only evaluated on the final \texttt{S2: S1 $\rightarrow$ SFT 60B Targeted 8} checkpoint.

\paragraph{Dev benchmarks}
We use \texttt{LM eval-harness} \citep{eval-harness} formulations of the following tasks.
\textbf{MMLU}~\citep{hendrycks2021measuring}: 57-subject multiple-choice
knowledge benchmark (${\sim}14$K problems), scored by log-likelihood
ranking over answer options.
\textbf{GSM8K}~\citep{cobbe2021gsm8k}: grade-school math word problems
(1,319 test), generative with exact-match on the final numeric answer.
\textbf{MATH500}~\citep{hendrycks2021math}: 500-problem subset of
competition mathematics, generative.
\textbf{AIME~2024 / AIME~2025}: competition-level mathematical reasoning
from the American Invitational Mathematics Examination, generative. We use \texttt{answer-in-text} filter for both AIME and Math500 benchmarks here, since we are interested mostly in the model's ability to preserve reasoning capabilities after distillation, rather than their ability to adhere to a specific output format. We include a more common \texttt{evalchemy} \citep{evalchemy} version in the unseen set.
\textbf{FDA}~\citep{arora2023structured}: zero-shot key--value extraction
from FDA drug-label PDFs.
\textbf{SWDE}~\citep{arora2023structured,lockard2019openceres}: zero-shot
relation extraction from semi-structured web pages.
\textbf{NIAH}~\citep{kamradt2023needleinhaystack} and \textbf{RULER}~\citep{hsieh2024ruler}: synthetic long-context retrieval and reasoning. Specifically, we use the \texttt{niah\_single\_2} (16384 context) and \texttt{ruler\_qa\_squad} (8192 and 16384 context) variants of the RULER benchmark.

\paragraph{Unseen Benchmarks} These benchmarks are only evaluated on final models and are not used for any type of decision-making during training (akin to a held-out test set), ensuring unbiased assessment of generalization.
\textbf{MMLU-Pro}~\citep{wang2024mmlu}: Enhanced version of MMLU with more challenging questions and increased answer choices (10 options vs 4), reducing guessing probability and requiring deeper reasoning.
\textbf{GPQA}~\citep{rein2023gpqa}: Graduate-level science questions (physics, chemistry, biology) written and validated by domain experts with PhD-level knowledge, testing advanced scientific reasoning.
\textbf{HLE (Humanity's Last Exam)}~\citep{phan2025humanity}: A comprehensive evaluation benchmark designed to test advanced AI capabilities across extremely challenging problems spanning mathematics, science, and reasoning that push the boundaries of current models.
\textbf{LCB}~\citep{jain2024livecodebench}: LiveCodeBench, a contamination-free code generation benchmark with problems collected after model training cutoff dates, ensuring true zero-shot evaluation.
\textbf{$\boldsymbol{\tau^2}$-Bench}~\citep{barres2025tau2}: A tool-augmented, dual-control benchmark for conversational agents that evaluates multi-step reasoning, coordination, and user-guidance in shared, dynamic environments.
\textbf{IFEval}~\citep{zhou2023ifeval}: Instruction-following evaluation measuring adherence to explicit constraints (format, length, content requirements) in generated responses.
We also evaluate on \textbf{AIME (NV)}~\citep{nemo_skills2024} -- a version of AIME~2025 problems with Nvidia's extraction protocol using the \texttt{evalchemy} filter, which enforces strict output formatting compared to our \texttt{answer-in-text} variant used in dev benchmarks. \textit{All evaluations are conducted in a fully generative setting, where models must produce free-form outputs.}

\paragraph{Placement scoring}
The default placement-optimization objective (Section~\ref{sec:placement}) uses the uniform average of mathematical reasoning dev benchmarks: MATH500, AIME-24, and AIME-25. For these benchmarks, we use \emph{log-likelihood} on teacher-generated traces under candidate placements, which serves as a proxy for generative accuracy.
The proxy was validated by correlation with exact-match scores on a random sample of placements in preliminary experiments.
Non-uniform weighting is possible when downstream priorities differ but is not explored here.
In a preliminary experiment (Appendix~\ref{app:prelim_exp}) we also explored including the rest of the all dev benchmarks into the placement-optimization procedure when using the \texttt{Idealized} cost model.
To make trace-likelihoods comparable to other accuracy-based benchmarks, we normalize the log-likelihoods based on the maximum and minimum values observed on a set of 100 random placements, including full-attention. Placements obtained using full set of dev benchmarks are marked with \texttt{All}, while placements obtained using log-likelihoods of teacher-generated traces are marked with \texttt{Lklhd}.
The 0.5B placements (Section~\ref{sec:landscape_dynamics}) are scored on MATH500 traces only and are not normalized.
Full evaluation protocols are in Appendix~\ref{app:evaluation}.

\paragraph{Throughput measurement}
All decode throughput figures are measured in \texttt{single-preset} mode.
The details on the exact setup used for throughput benchmarking can be found in Appendix~\ref{app:throughput_eval}.

\subsection{Pareto Frontier}
\label{sec:eval:pareto}

Applying the surrogate-assisted pipeline with \texttt{Regression}-based cost model from Section~\ref{sec:placement}
to the 15B \Super~Apriel after distillation (Section~\ref{sec:training:distillation})
yields the Pareto frontier in Figure~\ref{fig:pareto}.
The expansion selection procedure selected second order at range 1 features, showing non-trivial interactions between layers.

The frontier is smooth up to ${\sim}6{\times}$ speedup.
The scores decline gradually as FA layers are progressively replaced.
Beyond ${\sim}6{\times}$ the decline steepens sharply, coinciding with the eviction of the last SWA layer.
The surrogate overestimates quality at the highest speedups relative to the validated curve, reflecting extrapolation into regions sparsely covered during exploration.
The poor fit also reveals the non-linear nature of the placement-to-score relationship. Future works could explore using more complex surrogate models.

The layer-wise heatmap (bottom panel) reveals the eviction order: early
layers (0--14) switch from FA to cheaper mixers first, mid-network layers
(14--28) retain sliding-window attention longest,
with an overall transition from FA$\to$SWA$\to$KDA$\to$GDN as the cost budget tightens.
This eviction order is consistent with the per-layer mixer preferences
identified in Appendix~\ref{app:analysis}.

\subsection{Placement Presets}
\label{sec:eval:presets}
The smoothness of the Pareto frontier allows most placements from that frontier to achieve a good tradeoff between speed and performance.
For further fine-tuning and deployment, we handpick some placements from that frontier that act as presets.
Some of the experiments with surrogate-assisted search were carried out using an \texttt{Idealized} cost-model and using different sets of benchmarks.
We thus name the placement presets \texttt{\{Reg/Idealized\} | \{Lklhd/All\} -- cost} depending on:
\begin{itemize}
    \item the cost-model used: \texttt{Idealized} or \texttt{Regression}-based, see Appendix~\ref{app:cost_models} for details;
    \item the set of benchmarks used: only trace-likelihood-based math benchmarks (\texttt{Lklhd}) or additionally also the rest of the dev tasks (\texttt{All}) (we never include the full accuracy on AIME or Math since such evals are too expensive for search);
    \item the cost of the placement according to the cost-model used to obtain it (note the maximum possible cost for both cost models explored here is 48).
\end{itemize}

\begin{table}[t]
  \centering
  \caption{Recommended presets. Each row is a placement discovered from the \texttt{S1: Distil.} supernet checkpoint and used for targeted SFT to obtain the \texttt{S1: S1 $\rightarrow$ SFT 60B Targeted 8} checkpoint. Avg Acc. is the average across all evaluation tasks (dev + unseen benchmarks) from Table~\ref{tab:s2v2_full}. Retention is calculated relative to Apriel-1.6 teacher (73.9\%). Decode speedup measured @32k sequence length relative to the \texttt{all-FA} preset.
  }
  \label{tab:presets}
  \begin{tabular}{lccccccc}
    \toprule
    \textbf{Preset} & \textbf{FA} & \textbf{SWA} & \textbf{KDA}
      & \textbf{GDN} & \textbf{Avg Acc.} & \textbf{Retention} & \textbf{Speedup @32k} \\
    \midrule
    \texttt{all-FA}
      & 48 & 0 & 0 & 0
      & 74.2 & 100\% & 1.0$\times$ \\
    Reg$|$Lklhd--26
      & 12 & 26 & 6 & 4
      & 71.1 & 96\% & 2.9$\times$ \\
    Reg$|$Lklhd--18
      & 3 & 25 & 4 & 16
      & 69.7 & 94\% & 4.8$\times$ \\
    Reg$|$Lklhd--13
      & 0 & 16 & 13 & 19
      & 60.2 & 81\% & 6.9$\times$ \\
    Reg$|$Lklhd--10
      & 0 & 10 & 5 & 33
      & 57.2 & 77\% & 10.7$\times$ \\
    \midrule
    Idealized$|$All--18
      & 13 & 32 & 1 & 2
      & 71.8 & 97\% & 2.0$\times$ \\
    Idealized$|$Lklhd--6
      & 0 & 30 & 5 & 13
      & 66.8 & 90\% & 6.2$\times$ \\
    Idealized$|$All--6
      & 0 & 30 & 5 & 13
      & 65.3 & 88\% & 6.1$\times$ \\
    \bottomrule
  \end{tabular}
\end{table}

\subsection{Benchmark Comparison}
\label{sec:eval:comparison}

\begin{figure}[htbp]
    \centering
    \includegraphics[width=1\linewidth]{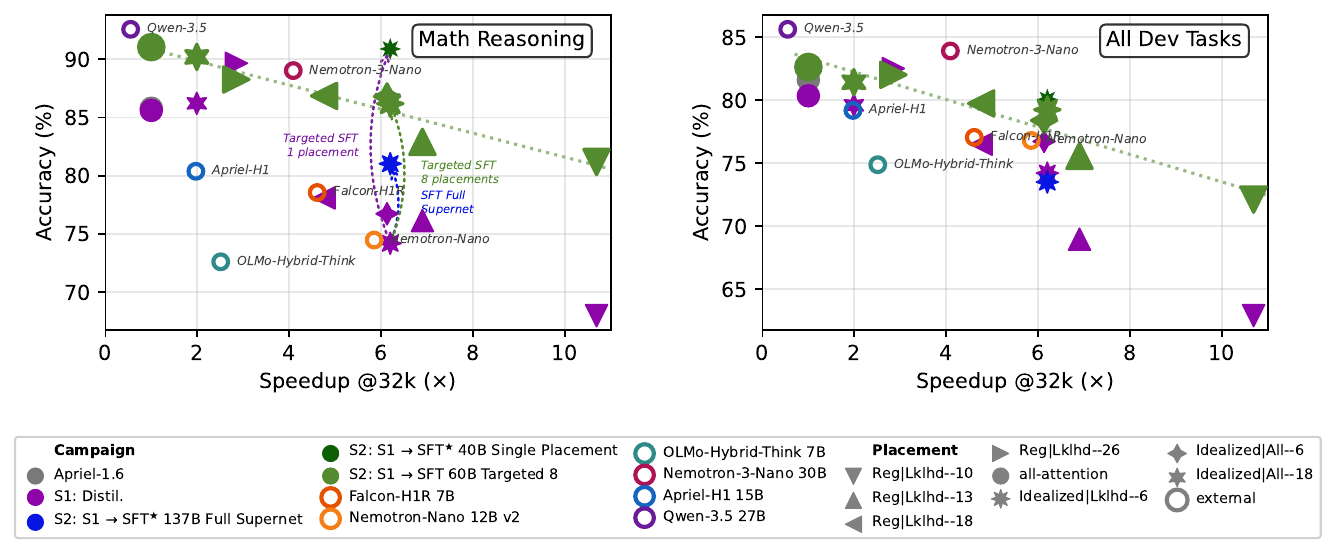}
    \caption{Performance of \Super~Apriel 15B versus decode throughput speedup @32k sequence length (over 500 samples) relative to the \texttt{all-FA} preset (Apriel-1.6). \textbf{Left:} average over math reasoning dev tasks (AIME'24, AIME'25, MATH-500, GSM8K). Here we demonstrate the effect of 3 different SFTs for \texttt{Idealized|Lklhd-6} layout: all starting from \texttt{S1: Distil.} using full supernet SFT (\dasharrow[sftfull]) using \texttt{SFT$^*$} data (no tool calling data), training in targeted way on 8 placements on full \texttt{SFT} data (\dasharrow[sfttargeted]) and training a single layout in isolation (\dasharrow[sftsingle]) on \texttt{SFT$^*$} data. \textbf{Right:} average over all dev tasks excluding NIAH.  Each color represents a model checkpoint. marker shapes denote the layer placement configuration. Filled markers are \Super~Apriel variants at different cost–accuracy operating points; open circles are external hybrid baselines. The dotted line shows a linear fit to the \texttt{S2: S1 $\rightarrow$ SFT 60B Targeted 8} checkpoint placements. Higher and further right is better. Full data is presented in Table~\ref{tab:eval_results_on_dev}, evaluation details including filters used to extract the final answers are detailed in Appendix~\ref{app:evaluation}. }
    \label{fig:per_task15B_dev}
\end{figure}

We evaluate \Super~Apriel placements against the teacher baseline (Apriel~1.6) and six external hybrid models across our benchmark suite.
Similar to Olmo-Hybrid \citep{merrill2025olmo}, we disable CUDA graphs by using \texttt{eager-mode} for linear mixers (both our and external models).
This is motivated by observed numerical instabilities in the current kernel implementations when using the graph mode. CUDA graph optimizations for hybrid architectures remain an active area of kernel development and we hope that these limitations will be mitigated in the future kernel releases. Additionally, for external hybrids we set  \texttt{-{}-disable-cascade-attn} and \texttt{-{}-mamba-ssm-cache-dtype float32}.

Figure~\ref{fig:per_task15B_dev} visualizes the performance-throughput tradeoff for three SFT campaigns (\texttt{S2: S1 $\rightarrow$ SFT$^*$ Full Supernet 137B}, \texttt{S2: S1 $\rightarrow$ SFT$^*$ Single placement 40B}, and \texttt{S2: S1 $\rightarrow$ SFT 60B Targeted 8}) alongside external baselines on the dev benchmarks. Figure~\ref{fig:per_task15B_all} focuses on all tasks averages. Full detailed numbers are in Table~\ref{tab:eval_results_on_dev}.  We focus our detailed analysis on the final \texttt{S2: S1 $\rightarrow$ SFT 60B Targeted 8 placements} campaign, which achieves the best quality-efficiency frontier.

\paragraph{Linear Pareto Frontier}
Figures~\ref{fig:per_task15B_all} and \ref{fig:per_task15B_dev} visualize the nearly linear Pareto frontier across all downstream benchmarks formed by \Super~Apriel placements, confirming the validity of our linear cost model assumptions underlying the cluster-expansion surrogate optimization (Section~\ref{sec:placement}). This linearity validates the surrogate-based search: placements discovered via dynamic programming on the surrogate model achieve predictable speedups in practice \footnote{Discrepancy between the cost model predictions and the measured throughput speedups reported in Table~\ref{tab:s2v2_full} arise from several sources: (1) the cost model was trained using throughput measurements at 16k sequence length with 500 prompts, while the final speedup measurements use 32k and 16k generation lengths with 1000 and 500 prompts (Table~\ref{tab:s2v2_full}). As shown in Section~\ref{sec:eval:efficiency}, efficient placements exhibit increasing relative speedup at longer contexts, so cost model predictions trained at 16k may underestimate actual speedups measured at 32k; (2) the \texttt{Idealized} cost model used a smaller SWA window size introducing additional discrepancy for placements optimized under that cost model. Despite these mismatches, the linear relationship holds across the measured range even on the downstream (dev and all) benchmarks}.

External hybrid models with different architectural choices occupy different regions of the tradeoff space. Some models prioritize quality while being slower (Qwen-3.5 27B at 0.5$\times$), while others target aggressive efficiency (Nemotron-Nano 12B v2 at 5.9$\times$). Notably, targeted supervised fine-tuning of efficient \Super~Apriel placements in isolation can push performance far beyond the initial distillation frontier. For instance, the \texttt{Idealized|Lklhd–6} placement achieves substantial gains of +4.7 on math reasoning over the full supernet SFT and +14.2 over \texttt{S1: Distil.} (see Figure~\ref{fig:per_task15B_dev} targeted SFT$^*$ training \texttt{S2: S1 $\rightarrow$ SFT$^*$ Single placement 40B}  \textcolor[HTML]{8D04A8}{\ding{88}}~\dasharrow[sftsingle]~\textcolor[HTML]{0E5E04}{\ding{88}}), demonstrating that domain \& cost-specific fine-tuning of discovered placements can further improve the quality-efficiency tradeoff.

\paragraph{Comparison to Teacher Baseline}
The \texttt{all-FA} placement serves as our internal teacher baseline, mostly matching or exceeding the original Apriel~1.6 teacher on the benchmarks: 74.2 vs.\ 73.9 on average on all tasks (see Table~\ref{tab:s2v2_full} in Appendix~\ref{app:downstream_evals} for complete benchmark results).

Efficient placements achieve significant speedups while retaining strong performance across all evaluation tasks:
\begin{itemize}
\item \textbf{Reg|Lklhd–26} (2.9$\times$ speedup @32k): Achieves 71.1 all-tasks average (96.2\% retention) and 88.3 math average (102.9\% of teacher). This configuration offers near-teacher quality while delivering substantial throughput gains, making it suitable for production deployments where quality cannot be compromised.

\item \textbf{Reg|Lklhd–18} (4.8$\times$ speedup @32k): Achieves 69.7 all-tasks average (94.3\% retention) and 86.8 math average (101.2\% of teacher). The minimal degradation across diverse task types demonstrates the robustness of efficient mixers for general workloads.

\item \textbf{Reg|Lklhd–13} (6.9$\times$ speedup @32k): Achieves 60.2 all-tasks average (81.5\% retention) and 82.9 math average (96.6\% retention) at substantial speedup, with graceful degradation on long-context and retrieval tasks. This operating point balances aggressive efficiency with production-viable quality.

\item \textbf{Reg|Lklhd–10} (10.7$\times$ speedup @32k): Achieves 57.2 all-tasks average (77.4\% retention) and 81.2 math average (94.6\% retention), delivering near-11$\times$ throughput improvement. The balanced degradation pattern indicates this placement remains viable for high-throughput production workloads where aggressive efficiency is required.
\end{itemize}

\paragraph{Comparison to Apriel-H1 and External Hybrid Models}
We compare \Super~Apriel presets against both internal and external baselines. Apriel-H1 (15B parameters) \citep{ostapenko2025aprielh1} achieves 2.0$\times$ speedup with 80.4 math reasoning average and 58.4 all-tasks average. \Super~Apriel provides multiple superior options near this speedup point: \texttt{Reg|Lklhd–26} at 2.9$\times$ achieves 88.3 math average (+7.9 points) and 71.1 all-tasks average (+12.7 points). These advantages can be attributed to both, stronger teacher model (Apriel~1.6 vs.\ 1.5) as well as the supernet's placement flexibility.

Among external models:
\begin{itemize}
\item \textbf{Qwen-3.5 27B:} a hybrid attention-GDN sparse mixture-of-experts (MoE) model~\citep{qwen3.5}. Despite being nearly 2$\times$ larger (27B vs 15B), operates at 0.5$\times$ speedup in our throughput tests (slower than full attention due to different architectural focus), achieving 92.6 math average and 80.7 all-tasks average. This model achieves the highest performance on our benchmarks, especially dominating the tool calling tasks like Tau2.

\item \textbf{Nemotron-3-Nano 30B:}~\citep{nvidia2025nemotron3nano} a hybrid sparse MoE model trained from scratch as an attention-Mamba hybrid. At 30B parameters (2$\times$ \Super~Apriel size), achieves a remarkable 4.1$\times$ speedup in our tests with 89.0 math average and 72.6 all-tasks average. \Super~Apriel's \texttt{Idealized|Lklhd–6} achieves comparable math performance (90.9) at better efficiency (6.2$\times$ speedup) after placement and domain-specific fine-tuning on SFT$^*$ dataset (excludes tool calling) despite being half the model size.

\item \textbf{Falcon-H1R 7B:} \citep{falcon-h1r} This smaller transformer-Mamba hybrid (7B vs 15B) achieves 4.6$\times$ speedup with 78.6 math reasoning average and 64.9 all-tasks average, demonstrating competitive efficiency for its size class.

\item \textbf{Nemotron-Nano 12B v2:} \citep{nvidia2025nano2} Achieves 5.9$\times$ speedup at 12B parameters, with 74.5 math reasoning average and 62.4 all-tasks average, prioritizing throughput over quality.

\item \textbf{OLMo-Hybrid-Think 7B:} \citep{merrill2025olmo} At 7B parameters and 2.5$\times$ speedup, achieves 72.6 math reasoning average and 56.1 all-tasks average, illustrating the challenges smaller models face on reasoning and tool calling tasks.
\end{itemize}

\Super~Apriel's supernet training produces a family of placements spanning the quality-efficiency space effectively. While larger models (Qwen-3.5 27B, Nemotron-3-Nano 30B) can achieve higher absolute quality, \Super~Apriel provides superior flexibility at the 15B scale, offering multiple operating points from the same checkpoint.

\paragraph{Supervised Fine-Tuning Improvements Over Distillation}

Comparing distillation-only results (\texttt{S1: Distil}) with targeted SFT (\texttt{S2: S1 $\rightarrow$ SFT 60B Targeted 8}) reveals substantial quality improvements across all placements. The \texttt{all-FA} placement improves from 90.0/86.7/85.0/80.7 (S1) to 93.3/86.7/91.8/92.3 (S2) on AIME'24/AIME'25/MATH-500/GSM8K, representing gains of +3.3/0.0/+6.8/+11.6 points respectively. The most significant improvements occur on mathematical reasoning tasks (MATH-500: +6.8, GSM8K: +11.6), where SFT recovers reasoning capabilities that distillation alone cannot capture.

This pattern extends across efficient placements, with even more dramatic improvements. For example, \texttt{Idealized|All–6} (6.1$\times$ speedup) shows remarkable gains from \texttt{S1} (63.3/63.3/84.2/96.1, 60.8 all-tasks) to \texttt{S2} (83.3/80.0/92.2/91.7, 65.3 all-tasks) on AIME'24/AIME'25/MATH-500/GSM8K, representing improvements of +20.0/+16.7/+8.0 points on AIME benchmarks and +4.5 points on all-tasks average. The particularly large gains on math (AIME'24: +20.0, AIME'25: +16.7) demonstrate that targeted SFT substantially improves reasoning capabilities for efficient placements.

Beyond the development benchmarks, \texttt{S2} introduces evaluation on additional unseen tasks (Tau2, MMLU-Pro, AIME(NV), GPQA, HLE, LCB, IFEval) that were not measured during \texttt{S1}, results are shown in Table~\ref{tab:s2v2_full}. For \texttt{Idealized|All–6}, these unseen tasks show strong performance in \texttt{S2}, with particularly notable results on Tau2 (+21.9 points from S1's 12.3 to S2's 34.2), AIME(NV) (+13.0 points, from 43.7 to 56.7), and LCB (+11.5 points, from 44.4 to 55.9).

\subsection{Efficiency}
\label{sec:eval:efficiency}

\begin{figure}[htbp]
    \centering
    \includegraphics[width=1\linewidth]{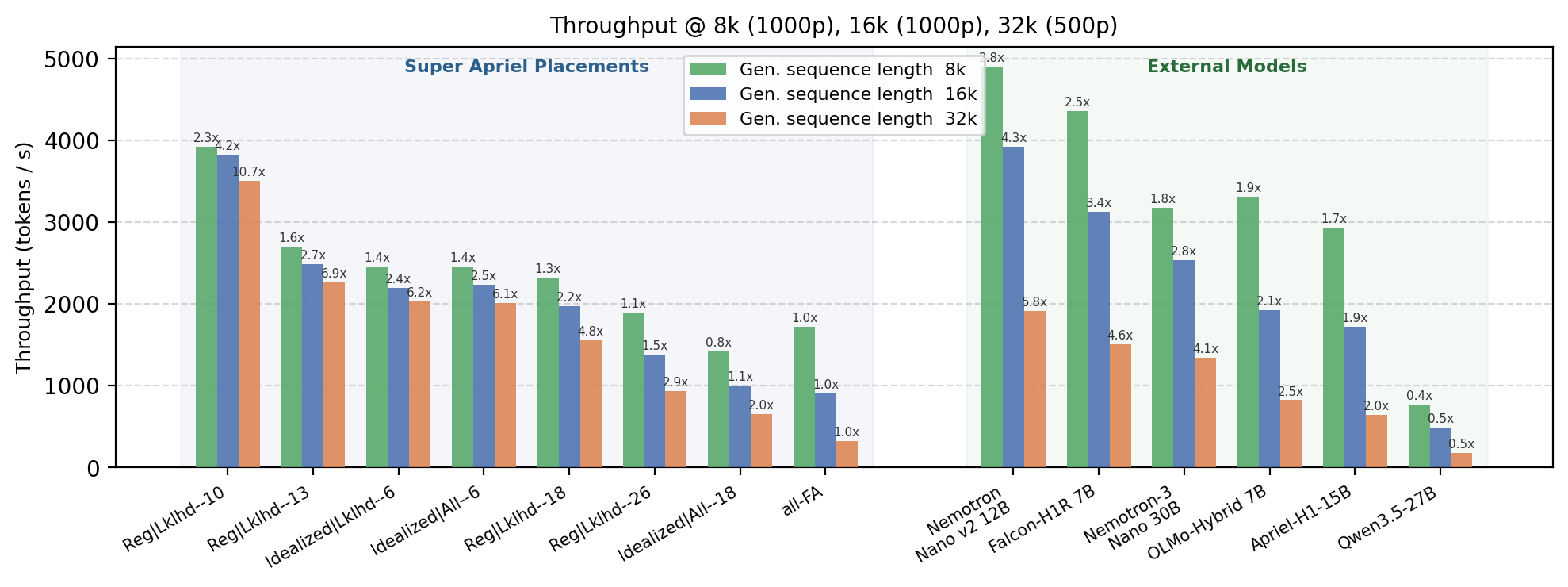}
    \caption{Throughput of \Super~Apriel placements in a \texttt{single-preset} mode and external hybrid models across context lengths. Decode throughput (tokens/s) on a single GPU at 8k, 16k (1000 prompts) and 32k context lengths. Bars are annotated with speedup relative to the \texttt{all-FA} baseline at the same context length (1722.27, 910.63, and 327.95 tok/s at 8k, 16k and 32k respectively). Left: \Super~Apriel placements from surrogate-guided search alongside the full-attention baseline. Right: external hybrid models shown for reference.}
    \label{fig:speedups}
\end{figure}

We evaluate decode throughput speedups for \Super~Apriel placements and external hybrid models at three sequence lengths: 8K, 16K, and 32K tokens. Table~\ref{tab:s2v2_full} reports speedups relative to the \texttt{all-FA} baseline measured on H100 GPUs using vLLM (details in Appendix~\ref{app:throughput_eval}), and Figure~\ref{fig:speedups} visualizes the throughput behavior across placements and context lengths. We detail the vLLM setup for these measurements in Appendix~\ref{app:throughput_eval}. We also provide preliminary results with a multi-placement serving implementation (currently under development) in Table \ref{tab:eval-multiplacement-throughput}.

We observe that \Super~Apriel placements achieve substantially higher relative speedups at longer contexts. While external hybrid models also benefit from longer sequences, their gains are significantly smaller (5--46\% increase from 16K to 32K) compared to \Super~Apriel placements (80--155\% increase over the same range).

For \Super~Apriel placements, the speedup improvements from 16K to 32K are consistent and are especially substantial for placements with fewer \texttt{FA} layers:
\begin{itemize}
\item \textbf{Reg|Lklhd--10:} 4.2$\times$ (16K) $\rightarrow$ 10.7$\times$ (32K), a 2.5$\times$ increase in relative speedup
\item \textbf{Reg|Lklhd--13:} 2.7$\times$ (16K) $\rightarrow$ 6.9$\times$ (32K), a 2.6$\times$ increase
\item \textbf{Idealized|Lklhd--6:} 2.4$\times$ (16K) $\rightarrow$ 6.2$\times$ (32K), a 2.6$\times$ increase
\item \textbf{Reg|Lklhd--18:} 2.2$\times$ (16K) $\rightarrow$ 4.8$\times$ (32K), a 2.2$\times$ increase
\item \textbf{Reg|Lklhd--26:} 1.5$\times$ (16K) $\rightarrow$ 2.9$\times$ (32K), a 1.9$\times$ increase (12 \texttt{FA} layers)
\item \textbf{Idealized|All--18:} 1.1$\times$ (16K) $\rightarrow$ 2.0$\times$ (32K), a 1.8$\times$ increase (13 \texttt{FA} layers)
\end{itemize}

This scaling advantage becomes more pronounced for more aggressive placements (those with fewer \texttt{FA} layers), indicating that efficient mixers provide increasing benefits as context length grows. For comparison, external hybrid models show the following gains:
\begin{itemize}
\item \textbf{OLMo-Hybrid-Think 7B:} 2.1$\times$ (16K) $\rightarrow$ 2.5$\times$ (32K), a 19\% increase
\item \textbf{Nemotron-Nano 12B v2:} 4.3$\times$ (16K) $\rightarrow$ 5.9$\times$ (32K), a 37\% increase
\item \textbf{Falcon-H1R 7B:} 3.4$\times$ (16K) $\rightarrow$ 4.6$\times$ (32K), a 35\% increase
\item \textbf{Nemotron-3-Nano 30B:} 2.8$\times$ (16K) $\rightarrow$ 4.1$\times$ (32K), a 46\% increase
\item \textbf{Apriel-H1 15B:} 1.9$\times$ (16K) $\rightarrow$ 2.0$\times$ (32K), a 5\% increase
\item \textbf{Qwen-3.5 27B:} 0.5$\times$ (16K) $\rightarrow$ 0.55$\times$ (32K), a 10\% increase
\end{itemize}

This difference in scaling magnitude reflects the architectural advantages. While all hybrid models benefit from longer sequences, \Super~Apriel's efficient mixers (SWA, KDA, GDN) have fixed memory footprints independent of sequence length, yielding 80--155\% relative speedup gains from 16K to 32K. In contrast, external hybrids with more attention layers show only 5--46\% gains over the same range, as their remaining full-attention components still incur quadratic KV cache overhead and memory-bandwidth bottlenecks.

This scaling behavior may change the deployment calculus based on workload characteristics. For short-context workloads where speedup gains are more modest, quality-focused placements (\texttt{Reg|Lklhd--26}, \texttt{Idealized|All--18}) may be preferred. For long-context workloads where efficiency advantages compound, aggressive placements (\texttt{Reg|Lklhd--10}, \texttt{Reg|Lklhd--13}) become significantly more attractive---not because they are different placements, but because the same placements deliver substantially higher throughput improvements. Supernet's deployment-time configurability enables selecting the appropriate operating point based on workload requirements, making the throughput--quality tradeoff a runtime control rather than a pre-release commitment.

\subsection{Speculative Decoding}
\label{sec:speculative}
\begin{figure}[t]
    \centering
    \includegraphics[width=0.8\linewidth]{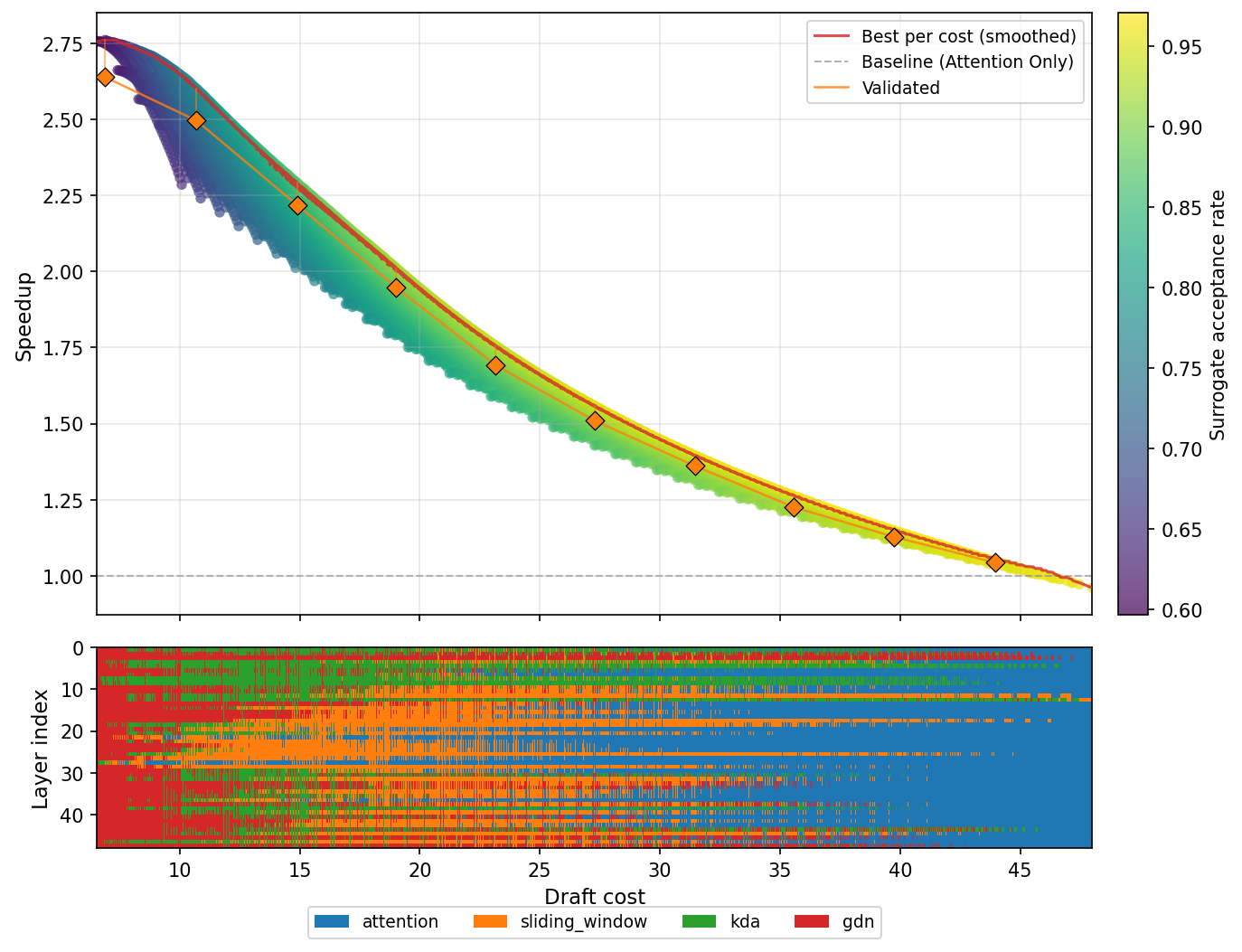}
    \caption{Speculative decoding speedup for the 15B supernet, averaged over
four math benchmarks (GSM8K, MATH500, AIME-24, AIME-25).  Each point
represents a candidate draft placement; the $x$-axis shows draft placement
cost and the $y$-axis shows the net decoding speedup after accounting for
the acceptance rate.  The cheapest draft (all-GDN) achieves the highest
speedup, indicating that acceptance rates remain high across the full
cost range.}
    \label{fig:averaged_speculative}
\end{figure}
Given access to a cheap draft model and an expensive target model, speculative decoding \citep{leviathan2023fast} allows one to achieve high efficiency without quality degradation by generating tokens from the draft and verifying them with fewer forward passes using the target.
Speculative decoding works best if the target and draft are similar in distribution, yielding a high acceptance rate.
Due to the joint training of different placements, our supernet is a natural candidate where the target model is given by the \texttt{all-FA} placement, and we can search over draft placements to find one with the optimal inference cost, determined jointly by the acceptance rate and cost of the draft placement.

Figure~\ref{fig:averaged_speculative} illustrates the landscape of speedups from different placements, averaged over four math benchmarks, where the acceptance rate is measured by fitting and searching a surrogate model as in Appendix~\ref{app:search}, and validating the results on a smaller subset of placements.
In general, a lower cost draft placement will be cheaper for decoding speculative tokens, but it might have a larger distance from the target placement's distribution and therefore a lower acceptance rate, which can increase the overall speculative decoding latency.
However, as Figure~\ref{fig:averaged_speculative} demonstrates, for the 15B model the supernet training seems to yield sufficiently high acceptance rates across all allocations, and the optimal speedup is achieved by the cheapest draft, in this case the all-GDN placement.

We provide details of efficient estimation of the acceptance rate and search for optimal draft placement in Appendix~\ref{app:speculative}.

\section{Limitations}
\label{sec:limitations}

\begin{itemize}
  \item \textbf{Long-range task regression.}
    Faster presets replace FA layers with recurrent mixers that compress
    context into a fixed-size state.
    Long-range retrieval tasks such as RULER and NIAH drastically regress on these presets (Table~\ref{tab:eval_results_on_dev}).

  \item \textbf{Short-range surrogate assumption.}
    The cluster expansion truncates at short-range interactions.  If
    long-context tasks induce long-range inter-layer dependencies, the
    surrogate may under-fit, and the placement optimizer may miss
    important interactions.

  \item \textbf{Log-likelihood proxy.}
    Placement optimization uses teacher-trace log-likelihood as a proxy
    for generative accuracy
    (Section~\ref{sec:eval:presets}).  The proxy was validated by
    correlation with exact-match scores on a random sample of placements,
    but a gap may persist for placements or tasks where likelihood and
    generation quality diverge.

  \item \textbf{Single teacher.}
    All results are specific to Apriel~1.6.  The interaction structure,
    per-layer preferences, and optimal presets may differ for other
    teacher models.

  \item \textbf{Ablations at scale.}
    Ablations on the training regime in
    Section~\ref{sec:landscape_dynamics} are performed on the 0.5B model.
    However, results have shown that the findings from these experiments
    do not necessarily transfer to the 15B model.  Experiments at a larger
    scale are necessary to better understand the landscape dynamics for
    large supernet training.
    Furthermore, the 15B SFT campaigns
    (Section~\ref{sec:training:sft}) differ in data composition and token
    budget, preventing controlled comparison of training strategies
    (stochastic vs.\ targeted placement sampling) at production scale.

  \item \textbf{Outlier placements.}
    Near-singleton placements were excluded from the optimization due to
    a poor fit of the linear cost model (see
    Appendix~\ref{app:cost_models}).
    As hybrid support matures in inference engines, the search could
    explore more feasible placements.

  \item \textbf{Inference engine variability.}
  Related to the previous point, inference engines such as vLLM evolve rapidly, and measured speedups---even accuracy numbers---may depend on the specific implementation and execution mode.
    Similar to OLMo-Hybrid~\citep{merrill2025olmo}, we use \texttt{eager-mode} for benchmark evaluation of all linear mixers to avoid potential numerical instabilities in current kernel implementations. Throughput measurements, by contrast, were conducted with CUDA graphs enabled. CUDA graph optimizations for hybrid architectures remain an active area of kernel development, and we expect performance discrepancies between modes to diminish as the ecosystem matures.
    Unless stated otherwise, we used \texttt{vLLM~v0.14.0rc1} for throughput benchmarking and cost model tuning. In additional experiments, we observed that relative speedups may vary across versions, highlighting the concern of serving engine's version becoming a moving target.

\end{itemize}

\section{Outlook}
\label{sec:outlook}

The released checkpoints include the distillation base model as well as
supervised fine-tuned variants (Section~\ref{sec:training:sft}).  Below
we describe reinforcement learning and deployment directions.

\subsection{Post-Training}
\label{sec:outlook:posttraining}

Following SFT, we plan RL fine-tuning using group relative policy
optimization~\citep{shao2024deepseekmath} on reasoning and agentic tasks.
A viable RL approach for supernets should balance performance, placement coverage, and training stability.
FA may be used as a stabilizing reference through an auxiliary KL term, particularly during early training when shared parameters may drift toward easier-to-optimize but less robust placements. Related strategies include initially restricting mixer changes or briefly biasing training toward FA before expanding to broader hybrid training.

Rollout design is another key choice. Using a single placement for all rollouts of a prompt is simple but may underutilize inference-time flexibility. Multi-placement rollout groups offer broader coverage and may support both placement-specific and aggregated advantage estimates; including FA in such groups can additionally enable cross-placement regularization. In more compute-intensive settings, FA generations could also supervise placements that fail to obtain reward.

\subsection{Scaling and Generalization}
\label{sec:outlook:scaling}

Several open directions remain.
First, extending the supernet approach to teacher models beyond
Apriel~1.6 would test the generality of the training recipe and
interaction structure observed here.
Second, the mixer vocabulary could be expanded with alternatives such as
Mamba-2~\citep{dao2024mamba2} or Lightning Attention~\citep{chen2025minimaxm1},
potentially improving the cost--quality frontier.
Third, a controlled comparison of placement sampling strategies
(stochastic vs.\ targeted) at production scale is a priority;
the 0.5B ablations
(Section~\ref{sec:landscape_dynamics:results:targeted}) favored
stochastic training, but confirming this at 15B requires
campaigns matched in data composition and token budget.
More broadly, the cost model depends on the inference engine and hardware
configuration, both of which evolve independently of the supernet
checkpoint.  If the Pareto frontier shifts at deployment time,
placements trained stochastically remain viable while targeted
placements may no longer be optimal.  The current release hedges by
targeting presets from two cost models, but a principled resolution
requires understanding the interaction between training strategy,
cost-model uncertainty, and deployment conditions.
Finally, validating whether the landscape-dynamics findings---early rank
stabilization and frontier volatility---hold at scales beyond 15B would
inform training-monitoring practices for future supernets.

\subsection{Deployment Roadmap}
\label{sec:outlook:deployment}

The current release serves one placement at a time. The \texttt{supernet} serving model with per-sample preset switching is under development.
Because all registered presets' mixer weights
must reside on GPU simultaneously, memory overhead in the \textbf{per-sample} \texttt{supernet} serving mode grows with the number
of distinct mixer types across presets.  Two mitigations apply:
\emph{model thinning}, which prunes each layer to the mixer types used
by at least one shipping preset; and \emph{memory-aware preset
selection}, which favors preset sets whose per-layer mixer vocabularies
overlap, minimizing the total distinct (layer, mixer-type) pairs and
thereby reducing GPU memory and improving batch capacity.  An additional
direction is speculative decoding with a fast hybrid draft and the
\texttt{all-FA} verifier, exploiting the shared checkpoint to avoid loading a
separate draft model.

\section{Related Work}
\label{sec:related}

\subsection{Hybrid-Attention Architectures}
\label{sec:related:hybrid}

Several groups released hybrid models mixing full attention with
efficient alternatives in 2025.
Nemotron-H~\citep{blakeman2025nemotronh} interleaves Mamba-2 with FA at a
9:1 ratio (47B parameters, ${\sim}3\times$ throughput).
Falcon-H1~\citep{zuo2025falcon} uses parallel Mamba-2 and FA streams
(34B, ${\sim}4\times$ prefill, ${\sim}8\times$ decode at long context).
MiniMax~M1~\citep{chen2025minimaxm1} uses Lightning Attention with FA at
7:1 (456B total, ${\sim}3$--$4\times$ decode at 100K).
Qwen3-Next~\citep{yang2025qwen3} uses GDN with FA at 3:1 (80B,
${>}10\times$ throughput at ${>}$32K).
Kimi~K2~\citep{kimi2025linear} uses KDA with MLA at 3:1 (48B total,
3B activated, 75\% KV cache reduction, $6\times$ decode at 1M).
Nemotron Nano~2~\citep{nvidia2025nano2} uses Mamba-2 with FA at 8:1
(30B, 3B activated, up to $3.3\times$ decode).

All of these are trained from scratch (or from a shared base with
extensive continued pretraining).  In all cases, the mixer pattern is
fixed, repetitive, and position-independent.  By contrast, we operate in
the conversion regime, where the placement pattern is inferred from the
teacher's learned circuits rather than fixed a priori.

\subsection{Attention-to-Efficient Conversion}
\label{sec:related:conversion}

The Mamba-in-the-Llama (MIL) framework~\citep{wang2024mamba_llama}
demonstrated that pretrained attention weights can initialize linear RNN
mixers, enabling conversion with academic-scale compute.
Apriel-H1~\citep{ostapenko2025aprielh1} extended this with progressive
conversion and reverse-KL distillation on reasoning data.
MOHAWK~\citep{bick2024mohawk} proposed a multi-stage conversion procedure
for Mamba-2.  Llamba~\citep{bick2025llamba} distilled Llama-3 into pure
Mamba-2 models at 1B--8B scale using the MOHAWK procedure.
Jet-Nemotron~\citep{gu2025jetnemotron} introduced PostNAS, a post-training
architecture search pipeline that starts from a pretrained FA model with
frozen MLPs and searches over FA, SWA, and linear attention placements
using hierarchical beam search.

\textbf{One-shot NAS supernets}~\citep{bender2018oneshot,pham2018enas,
guo2020single,cai2020once}.  Standard NAS supernets train architectural
choices (e.g., kernel sizes, channel widths) from scratch with path
sampling.  Our supernet differs in initialization (all mixers are
warm-started from the teacher via MIL/DIL/KIL surgery), objective
(distillation from a frozen teacher rather than self-supervised
training), and the nature of the choices (fundamentally different
architectures --- attention vs.\ linear recurrence --- rather than
parametric variants of the same operation).
A known limitation of weight-sharing supernets is that sub-network
rankings under the shared weights may not predict rankings after
standalone training~\citep{bender2018oneshot,yu2020evaluating}.
Section~\ref{sec:landscape_dynamics} addresses the analogous question
for our setting: whether placement rankings at one training stage
predict rankings at the next.

\textbf{Jet-Nemotron PostNAS}~\citep{gu2025jetnemotron}.  Concurrent work that
constructs a similar supernet with stochastic layer sampling, but
considers a different mixer vocabulary (FA, SWA, JetBlock) and uses
hierarchical beam search for placement.  Our work extends this paradigm to GDN and KDA and replaces beam search
with exact optimization of a fitted cluster-expansion surrogate, which
covers all cost-feasible placements rather than the limited subset
explored by beam search.

\textbf{Apriel-H1 progressive conversion}~\citep{ostapenko2025aprielh1}.
H1 evaluates one placement perturbation per probing step, building up a
placement sequentially.  The supernet enables batch evaluation of many
placements, from which a surrogate of the score landscape is fitted and
optimized.  Training is also parallelized: all mixers train
simultaneously, and placement search is decoupled from training.

\subsection{Surrogate-Assisted Optimization}
\label{sec:related:bo}

The explore--exploit loop follows the standard Bayesian optimization
paradigm~\citep{jones1998ego,shahriari2016bo}.  Our use of exact
optimization (rather than heuristic search as in
SMAC~\citep{hutter2011smac} or random-restart local search) is possible
because of the linear decomposable structure of the cluster-expansion
surrogate.  Predictor-based NAS
methods~\citep{wen2020predictor,white2021bananas,kandasamy2018nasbo}
train models to predict architecture quality from structural features.
The cluster-expansion surrogate replaces heuristic search over
predictions with exact optimization.

\subsection{Cluster Expansion in Materials Science}
\label{sec:related:cluster}

The cluster expansion~\citep{sanchez1984cluster,defontaine1994cluster,
vandewalle2002automating,zunger1994alloy} is the standard method for
parameterizing configuration-dependent properties of crystalline alloys.
Lattice sites correspond to our transformer layers, atomic species to
mixer types, and formation energy to benchmark score.  The convergence
theorem --- the expansion is exact and converges rapidly for short-range
interactions --- applies to our setting because transformer layer
interactions are mediated by the residual stream and decay geometrically
with layer distance.  We are not aware of prior work applying the cluster expansion formalism
to neural architecture search.

\subsection{Efficient Mixer Architectures}
\label{sec:related:mixers}

GDN~\citep{yang2025gateddeltanet} builds on DeltaNet~\citep{schlag2021linear},
which applied the delta rule~\citep{widrow1960adaptive} to linear
attention.  GDN adds Mamba-style scalar gating for adaptive forgetting.
KDA~\citep{kimi2025linear} extends GDN with channel-wise gating, the
key modification being the replacement of a scalar gate with a
per-dimension gate vector.  Both use chunkwise parallel training
algorithms that enable hardware-efficient computation via tensor cores
while preserving the linear recurrence semantics.  The broader landscape
includes Mamba/Mamba-2~\citep{gu2023mamba,dao2024mamba2},
RWKV~\citep{peng2023rwkv}, RetNet~\citep{sun2023retnet}, and
Griffin~\citep{de2024griffin}.  Our mixer vocabulary was chosen to span
the cost--expressiveness frontier while remaining compatible with a
common initialization from attention weights.

\section{Conclusion}
\label{sec:conclusion}

\Super~Apriel demonstrates that a single supernet checkpoint can serve
multiple speed presets, from teacher-equivalent quality to
near-$10\times$ throughput hybrids, via placement switching at serving
time.
The shared checkpoint doubles as a speculative-decoding pair: the
\texttt{all-FA} placement serves as verifier and any efficient placement
as draft, avoiding a separate draft model
(Section~\ref{sec:speculative}).
In practice, deployment flexibility manifests as:
(a)~operational simplicity---one artifact to version, track, and deploy,
    with the ability to add new operating points without retraining;
(b)~load-adaptive serving---the released checkpoint ships eight
    Pareto-optimal presets spanning $2.9\times$ to $10.7\times$ speedup,
    switchable at runtime based on request load;
(c)~task-sensitive quality tradeoffs---capabilities degrade
    non-uniformly as FA layers are replaced, so the operating point can
    be matched to the application's sensitivity to long-range recall.

During distillation, all four mixer types are trained simultaneously
via stochastic placement sampling.
The cluster-expansion surrogate admits exact cost-constrained
optimization over the combinatorial placement space; Bayesian model
evidence selects the expansion order, which for the 15B model on
reasoning tasks is second-order at range~1, revealing non-trivial
pairwise interactions between adjacent layers
(Section~\ref{sec:eval:pareto}).
The full-attention preset performs on par with the teacher on all benchmarks. Recommended presets range from $2.9\times$ speedup with 96\% quality retention to $10.7\times$ speedup with 77\% quality retention, with a smooth Pareto frontier in between.
Throughput advantages scale with context-length: efficient placements gain 80--155\% relative speedup from 16K to 32K sequence length, substantially outpacing external hybrid baselines (5--46\% gains), making the supernet increasingly attractive for long-context workloads.
Controlled ablations on a 0.5B model show that placement rankings
stabilize early under stochastic training ($\rho > 0.98$ from
3--6\% of tokens), supporting stochastic distillation as a default
strategy.  At 15B scale, frontier placements exhibit higher
volatility, and small-scale findings do not automatically transfer
(Section~\ref{sec:landscape_dynamics}).
SFT substantially improves all evaluated placements, with the largest
gains on mathematical reasoning tasks
(Section~\ref{sec:eval:comparison}).
The supernet weights, training code, vLLM serving extension, and
placement optimization toolkit are released as open source in
Fast-LLM~\citep{fastllm}.
See Section~\ref{sec:limitations} for limitations and
Section~\ref{sec:outlook} for planned RL and deployment directions.

\section*{Author Contributions}
\label{sec:contributions}

\begin{itemize}
\item \textbf{Torsten Scholak} led the project, contributing to
  experiment design, the supernet model and surgery code, the training
  implementation, the vLLM serving implementation, the surrogate-based
  placement optimization pipeline, the placement evaluation framework,
  and writing.

\item \textbf{Oleksiy Ostapenko} contributed to model development and experiment
  planning, optimizing the vLLM supernet serving infrastructure, running
  supernet training (distillation and SFT), contributing to mixer
  initialization and to the placement evaluation code, benchmarking
  inference throughput, and writing.

\item \textbf{Raymond Li} contributed to experiment design, the placement optimization
  and evaluation code, the placement landscape experiments and Pareto
  frontier extraction, activation distillation implementation for supernet
  training, and writing.

\item \textbf{Joel Lamy Poirier} developed and maintained Fast-LLM, the
  training framework used throughout this work, contributed to loss functions
  used for supernet distillation, implemented streaming dataset
  and weight broadcasting support for the PipelineRL integration, and
  contributed to checkpoint conversion between training and inference
  formats.

\item \textbf{Denis Kocetkov} led the integration of PipelineRL with
  Fast-LLM, enabling asynchronous online RL training for the supernet,
  including initial streaming dataset and weight broadcasting
  implementations and PipelineRL-side configuration.

\item \textbf{Alireza Mousavi-Hosseini} designed and implemented speculative
  decoding evaluation for test-time placement selection.

\item \textbf{Nanda H Krishna} prototyped vLLM supernet modeling
  and solved per-request placement switching.

\item \textbf{Rafael Pardinas} implemented the PipelineRL actor-side inference
  pipeline, including Fast-LLM--vLLM weight transfer integration and
  inference-side quantization support.

\item \textbf{Shruthan Radhakrishna} contributed to experiment design, implementation of tensor parallelism, fixes for layer placement bugs, placement landscape dynamics experiments on the 0.5B model, and benchmarking and evaluation.

\item \textbf{Kelechi Ogueji} contributed to the initial plan for supernet RL training.

\item \textbf{Aman Tiwari} contributed to the experimental design, executed placement landscape experiments, conducted Pareto frontier extraction and analysis, and performed ablation studies focused on the 0.5B development supernet, along with benchmarking and evaluation efforts.

\item \textbf{Sathwik Tejaswi Madhusudhan} provided high-level guidance
  in the early stages of the project.

\item \textbf{Srinivas Sunkara} provided organizational support and helped
  align the supernet work with production needs.

\item \textbf{Val\'{e}rie B\'{e}caert} provided organizational support.

\end{itemize}

\bibliographystyle{apriel}
\bibliography{refs}

\newpage
\beginappendix

\section{Serving scenarios}
\label{app:motivation_serving}

In principle, \Super~Apriel enables flexible deployment strategies based on two key factors: \textbf{domain knowledge:} whether annotated benchmarks are available for the target domain, \textbf{optimization priority:} quality-constrained or throughput-constrained. We provide a summary of potential deployment scenarios and strategies in Table~\ref{tab:deployment_scenarios}.
\begin{table}[h]
\centering
\small
\begin{tabular}{|p{2.8cm}|p{5.8cm}|p{6cm}|}
\hline
\textbf{Domain Knowledge} & \textbf{Quality-Constrained} \newline (Maintain quality, maximize throughput) & \textbf{Throughput-Constrained} \newline (Meet throughput target, maximize quality) \\
\hline
\textbf{Known Domain} \newline (annotated benchmarks available) &
\textbf{Goal:} Maintain domain quality (match FA or within threshold), maximize throughput.

\textbf{Strategy:} Use \textbf{speculative decoding} (SD) for zero quality loss. Run placement optimization to find optimal draft  (function of draft cost and SD acceptance rate). \textbf{Search:} Maximize overall SD throughput using cheap proxy (SD acceptance rate or likelihood on domain prompts). \textbf{Validate:} Evaluate top draft candidates on domain benchmark. Deploy FA teacher + best draft.

\textbf{Result:} Zero quality loss (matches FA teacher), maximum throughput from domain-validated draft.

\textbf{Example:} Medical QA requiring high accuracy. &
\textbf{Goal:} Meet hard throughput requirement, maximize domain performance.

\textbf{Strategy:} Run placement optimization with throughput constraint $B$ (or sweep multiple $B$ values to generate Pareto frontier). \textbf{Search:} Use cheap proxy (SD acceptance rate or likelihood on domain prompts) if domain benchmark is too expensive. \textbf{Validate:} Evaluate top candidates on domain benchmark, select best.

\textbf{Result:} Best-performing placement on domain benchmarks within throughput constraint.

\textbf{Example:} Code completion with latency SLA---search using likelihood, validate on HumanEval. \\
\hline
\textbf{Unknown Domain} \newline (no annotated benchmarks) &
\textbf{Goal:} Maintain quality (match FA), maximize throughput without annotated benchmarks.

\textbf{Strategy:} Use \textbf{speculative decoding} for zero quality loss. Run placement optimization to find draft that maximizes overall SD throughput on unlabeled requests (function of draft cost and SD acceptance rate). Deploy FA teacher + optimized draft.

\textbf{Result:} Zero quality loss (matches FA teacher), maximum SD throughput from optimal draft.

\textbf{Example:} Production chatbot where quality is paramount. &
\textbf{Goal:} Meet throughput requirement, maximize quality without annotated benchmarks.

\textbf{Strategy:} Use \textbf{SD acceptance rate as quality proxy} to guide placement search. Run placement optimization to find placement that maximizes SD acceptance rate (quality proxy) subject to throughput constraint $B$. Deploy placement directly.

\textbf{Alternative:} Use pre-computed Pareto frontier from general benchmarks, select placement at throughput target.

\textbf{Result:} Best-quality placement (by SD acceptance rate proxy) within throughput constraint.

\textbf{Example:} Generic API with resource limits---find placement with high SD acceptance rate meeting throughput SLA. \\
\hline
\textbf{Multi-Domain} \newline (prompt-conditioned routing) \newline \textit{[Future Work]} &
\textbf{Goal:} Maintain per-request quality by routing to optimal placement for each prompt.

\textbf{Strategy:} Train routing mechanism to predict highest-quality placement for each prompt using prompt features, task detection, or learned embedding.

\textbf{Open Questions:} Ground truth for ``optimal placement'' per prompt? How to train router? Routing overhead vs benefits?

\textbf{Example:} Production assistant---route complex reasoning to FA, simple completion to efficient placements. &
\textbf{Goal:} Meet throughput targets while maximizing quality across heterogeneous workload.

\textbf{Strategy:} Route requests to placements balancing per-domain quality and overall throughput constraint. Balance routing granularity with batching efficiency.

\textbf{Open Questions:} How to predict placement from prompt? Heterogeneous batching challenges? Managing mixed KV cache and recurrent states? Optimal routing granularity?

\textbf{Example:} Multi-domain API serving retrieval, generation, reasoning---route retrieval to FA-heavy, generation to efficient placements while meeting overall throughput SLA. \\
\hline
\end{tabular}
\caption{Deployment scenarios based on domain knowledge and optimization priority.}
\label{tab:deployment_scenarios}
\end{table}

\section{Mixer Architecture Details}
\label{app:mixer_details}

\subsection{Full Attention (FA)}

Standard grouped-query attention~\citep{ainslie2023gqa} with rotary
position embeddings~\citep{su2024roformer}.  The query, key, and value
projections follow the teacher's configuration.  FA computes exact
pairwise token interactions with $O(n^2)$ time and $O(n)$ KV cache per
layer.

At decode step~$t$, the KV-read operational intensity per layer is
approximately $h_q / (2\, h_{kv}\, b)$, where $h_q$ and $h_{kv}$ are the
numbers of query and KV heads and $b$ is the bytes per element.  For
typical GQA configurations on current hardware, this is orders of
magnitude below the compute--memory ridge point, placing FA decode deep
in bandwidth-bound territory.  The decode latency contribution of each FA
layer grows linearly with context length:
$T_{\text{decode}}(t) \sim 2t \, h_{kv} \, d_h \, b \, /
\, B_{\text{HBM}}$.

\subsection{Sliding Window Attention (SWA)}

Attention restricted to a fixed local window of $w = 4096$ tokens.  Each
token attends only to the $w$ most recent positions, yielding $O(nw)$
time and $O(w)$ KV cache per layer.  The query, key, value, and output
projection weights are identical to FA; only the attention mask changes.

\subsection{Gated DeltaNet (GDN)}

Gated DeltaNet~\citep{yang2025gateddeltanet} combines a scalar gating factor
$\alpha_t \in (0,1)$ for adaptive global forgetting with the delta
rule~\citep{widrow1960adaptive} for targeted memory updates.  The
recurrence is:
\[
  \mathbf{S}_t = \alpha_t
  \left(\mathbf{I} - \beta_t \mathbf{k}_t \mathbf{k}_t^\top\right)
  \mathbf{S}_{t-1}
  + \beta_t \mathbf{k}_t \mathbf{v}_t^\top
\]
where $\mathbf{k}_t, \mathbf{v}_t$ are the key and value vectors,
$\beta_t$ is a scalar learning rate, and the projection
$(\mathbf{I} - \beta_t \mathbf{k}_t \mathbf{k}_t^\top)$ erases the
component of $\mathbf{S}_{t-1}$ along $\mathbf{k}_t$ before writing the
new association.  The scalar gate $\alpha_t$ enables rapid global
forgetting when context shifts.

GDN uses a fused input projection producing Q, K, V, and a modulation
vector~$\mathbf{z}$ in a single matrix multiply, plus a separate
projection for the gating parameters ($\beta$,~$\alpha$).  A short
causal 1D convolution with SiLU activation is applied jointly to Q, K, V
before the recurrence.  The gating computation produces
$\beta = \sigma(\beta_{\text{raw}})$ and
$g = -\exp(A_{\log}) \cdot \mathrm{softplus}(\alpha + \text{dt\_bias})$,
combining learned and data-dependent decay.  After the chunkwise
recurrence, outputs pass through gated RMS normalization (with
$\mathbf{z}$ as the gate) and then the output projection.

For \Super~Apriel, GDN is configured with 32~value heads, 8~key heads, and
$d_k = d_v = 128$, giving a state matrix
$\mathbf{S} \in \mathbb{R}^{128 \times 128}$ per key head.  Queries and keys
are $L_2$-normalized.  Training uses a chunkwise parallel algorithm based
on the WY representation~\citep{yang2024b_parallel}.

\subsection{Kimi Delta Attention (KDA)}

KDA~\citep{kimi2025linear} extends GDN with channel-wise (vector-valued)
gating.  Where GDN applies a single scalar gate uniformly across all
dimensions, KDA uses a per-dimension gate vector
$\boldsymbol{\alpha}_t \in (0,1)^{d_k}$:
\[
  \mathbf{S}_t =
  \left(\mathbf{I} - \beta_t \mathbf{k}_t \mathbf{k}_t^\top\right)
  \mathrm{diag}(\boldsymbol{\alpha}_t) \,
  \mathbf{S}_{t-1}
  + \beta_t \mathbf{k}_t \mathbf{v}_t^\top
\]
Different state dimensions can decay at different rates: some maintain
long-lived factual associations while others rapidly update working
memory.

KDA uses separate Q, K, V projections, each followed by its own causal
1D convolution with SiLU activation.  The per-dimension gate
$\boldsymbol{\alpha}_t$ is produced by a low-rank factorization:
$\mathbb{R}^d \to \mathbb{R}^{d_h} \to \mathbb{R}^{n_h d_h}$,
followed by the fused gate computation.  A separate output gate with the
same low-rank structure modulates the recurrence output before the final
projection.  For \Super~Apriel, KDA uses 32~heads with $d_h = 128$.  The efficient chunkwise training algorithm uses
DPLR transition matrices~\citep{kimi2025linear}.

\section{Initialization and Training Configuration}
\label{app:init}

\subsection{Delta-in-the-Llama (DIL) \textemdash GDN Initialization}

The teacher's Q, K, V projections are mapped into GDN's fused input
structure via cyclic head tiling: for GDN key head~$g$, the query rows
come from teacher Q head~$g \bmod h_q^{\text{src}}$ and the key rows
from teacher KV head~$g \bmod h_{kv}^{\text{src}}$.  Value heads are
tiled at finer granularity since GDN uses more value heads ($h_v = 32$)
than key heads ($h_k = 8$).  Per-parameter initialization:
\begin{itemize}
  \item Modulation vector~$\mathbf{z}$: zeros (no output modulation at
    initialization)
  \item Gate projections ($\beta$,~$\alpha$): zeros
    ($\sigma(0) = 0.5$ for $\beta$; $\alpha = 0$)
  \item Causal convolution: scaled identity ($0.5$ on the last kernel
    tap, zeros elsewhere)
  \item Decay parameter~$A_{\log}$: $\log(0.1)$ (slow decay, retains
    state ${\sim}10$ steps)
  \item Timestep bias~$\text{dt\_bias}$: zeros
  \item Output projection: copied from the teacher
\end{itemize}

\subsection{Kimi-in-the-Llama (KIL) \textemdash  KDA Initialization}

Q, K, V projections are transferred individually (not fused) from the
teacher, with cyclic tiling when the target head count exceeds the
source.  The output projection is copied without modification.
Per-parameter initialization:
\begin{itemize}
  \item Per-dimension decay gate (low-rank): Kaiming normal
  \item Output gate (low-rank): Kaiming normal
  \item Per-head $\beta$ projection: randomly initialized
  \item Three per-stream causal convolutions: scaled identities
  \item $A_{\log}$ and dt\_bias: same as DIL
\end{itemize}

\subsection{Training Hyperparameters}

\begin{table}[h]
  \centering
  \caption{Distillation hyperparameters for the 15B \Super~Apriel supernet and the
  0.5B development supernet.  All runs use AdamW
  ($\beta_1 = 0.9$, $\beta_2 = 0.95$, weight decay $0.1$), bfloat16
  mixed precision, and only mixer weights are trainable.
  SFT hyperparameters are in Section~\ref{sec:training:sft}.}
  \label{tab:training}
  \resizebox{\textwidth}{!}{
  \begin{tabular}{llllll}
    \toprule
    & \multicolumn{2}{c}{\textbf{Stage~1 (1.5 teacher)}}
    & \multicolumn{2}{c}{\textbf{Stage~2 (1.6 teacher)}}
    & \textbf{0.5B Dev} \\
    \cmidrule(lr){2-3} \cmidrule(lr){4-5} \cmidrule(lr){6-6}
    \textbf{Parameter}
      & \textbf{Distill} & \textbf{Anneal}
      & \textbf{Distill} & \textbf{Anneal}
      & \\
    \midrule
    Learning rate     & $10^{-5}$ (const.)  & $10^{-5} \to 0$ (linear)
                      & $10^{-5}$ (const.)  & $10^{-5} \to 0$ (linear)
                      & $10^{-4}$ (const.) \\
    Warmup steps      & 1\,000 & 0 & 1\,000 & 0 & 100 \\
    Iterations        & 95\,332 & 5\,000 & 25\,090 & 10\,000 & 60\,000 \\
    Tokens            & 187B & 10B & 49B & 20B & ${\sim}$63B \\
    GPUs (H100)       & 192 & 160 & 120 & 120 & 64 \\
    Wall-clock (hours) & 270 & 17 & 100 & 33 & ${\sim}$11 \\
    ZeRO stage        & 3 & 3 & 2 & 2 & 3 \\
    Seq.\ length      & 16\,384 & 16\,384 & 16\,384 & 16\,384 & 8\,192 \\
    Batch (sequences) & 120 & 120 & 120 & 120 & 128 \\
    $\lambda_{\text{act}}$ & 0.5 & 0.5 & 0.5 & 0.5 & 1.0 \\
    $\lambda_{\text{RKL}}$ & 1.0 & 1.0 & 0.9 & 0.9 & 0.9 \\
    $\lambda_{\text{FKL}}$ & ---   & ---   & 0.1 & 0.1 & 0.1 \\
    \bottomrule
  \end{tabular}}
\end{table}

\paragraph{Apriel~1.5 provenance.}
Supernet development began while Apriel~1.6 was still in training, so
Stage~1 distilled from the then-current Apriel~1.5
teacher~\citep{radhakrishna2025apriel15}.  When 1.6 became available,
the trained GDN and KDA mixer weights were transplanted into a new
supernet whose shared parameters (FFNs, embeddings, norms, vision
encoder) come from the 1.6 checkpoint.  Because 1.6 is a post-trained
continuation of 1.5 with the same decoder architecture, the transplant
required no adaptation and Stage~2 training resumed immediately.

\subsection{0.5B Development Supernet}
\label{app:dev_model}

\paragraph{Architecture and initialization.}
The development supernet is derived from
Qwen2-0.5B~\citep{yang2024qwen2}: 24~layers, $d{=}896$, 14~query /
2~KV heads ($d_h{=}64$), vocabulary of 151\,936~tokens, text-only (no
vision encoder).  Qwen2 has no architectural relationship to the Apriel
lineage; it was chosen as a widely available small GQA transformer for
fast ablation turnaround.  The supernet equips every layer with the
same four mixer types as \Super~Apriel (FA, SWA, GDN, KDA), using the
general initialization described in
Section~\ref{sec:training:surgery}: FA and SWA weights come from
Qwen2's own attention, and GDN/KDA projections are initialized via
DIL/KIL (Appendix~\ref{app:init}) from those same weights.  Unlike the
15B supernet, there is no prior distillation stage---all four mixers
start from the same source checkpoint and are trained simultaneously.
This model serves as a controlled proxy: small enough for fast
iteration across multiple training regimes, large enough for the
landscape dynamics to be meaningful.

\paragraph{Training.}
The supernet is trained in two stages.  First, Qwen2-0.5B is
fine-tuned for 18\,000~steps (learning rate $5{\times}10^{-4}$,
sequence length 8\,192, batch size 128) with standard next-token
prediction on a reasoning dataset~\citep{guha2025openthoughtsdatarecipesreasoning} with 1M high-quality examples covering math, science, code, and puzzles.  Both the teacher (full-attention
variant) and the student (supernet with all four mixer types) are then
initialized from this checkpoint and the student is distilled for
60\,000~steps (learning rate $10^{-4}$, constant schedule, 100 warmup
steps) using per-block activation distillation
($\lambda_{\text{act}}{=}1.0$), reverse KL
($\lambda_{\text{RKL}}{=}0.9$), and forward KL
($\lambda_{\text{FKL}}{=}0.1$); the standard LM loss is disabled.
At each step, one mixer per layer is sampled uniformly at random.
The full hyperparameter set is in Table~\ref{tab:training}.

\section{Cost model}
\label{app:cost_models}
\begin{table}[t]
  \centering
  \caption{Relative per-layer cost of each mixer (attention = 1.00),
  measured on H100 with 15B-parameter Apriel 1.6 hybrids,
  sequence length 16k.
\textbf{Idealized$^*$} column --- cost for mixer $X$ is simply: $c_X = \frac{\text{throughput}_{\text{att}}}{\text{throughput}_X}$ where both throughputs are measured on pure placements — all 48 layers set to the same mixer type. This assumes each layer contributes independently and equally to total latency, with no cross-mixer overhead ($w$ -- window size for SWA); \textbf{Idealized} -- uses a slightly different vLLM setup and $w{=}2048$, details below. \textbf{Regression (clean)} = linear fit of $1/\text{throughput}$ on mixer counts from 84 mixed placements (each mixer $\geq 3$ layers);
  \label{tab:cost_model_detailed}
  \textbf{Regression (all)} --- same fit on all 123 placements including singleton placements with mixer counts $\leq 3$. See below for a discussion of the outlier effect induced by singleton mixer in VLM (unless stated otherwise, we use vLLM v0.14.0rc1).}
  \label{tab:mixer_costs}
  \begin{tabular}{l cccc}
    \toprule
    \textbf{Mixer} & \textbf{Idealized} & \textbf{Idealized$^*$} & \textbf{Regression} & \textbf{Regression} \\
    & ($w{=}2048$) & ($w{=}4096$) & (clean, $w{=}4096$) & (all, $w{=}4096$) \\
    \midrule
    FA    & 1.00 & 1.00 & 1.00 & 1.00 \\
    SWA   & 0.16 & 0.44 & 0.48 & 0.50 \\
    KDA   & 0.11 & 0.16 & 0.21 & 0.36 \\
    GDN   & 0.09 & 0.10 & 0.14 & 0.28 \\
    \midrule
    $R^2$          & ---  & ---  & 0.977 & 0.576 \\
    $N$            &  4   & 4    & 84    & 123 \\
    \bottomrule
  \end{tabular}
\end{table}

To enable cost-constrained placement search, we use a linear cost model that predicts inference throughput from the mixer composition of a placement.
We model the inverse throughput (latency per token) as a sum of per-layer contributions:
\begin{equation}
  \frac{1}{\text{throughput}} = \sum_{m \in \mathcal{M}} c_m \cdot n_m,
  \label{eq:cost-model}
\end{equation}
where $n_m$ is the number of layers assigned to mixer type $m$ and $c_m$ is the per-layer cost coefficient.
Normalizing by the attention coefficient $c_{\text{FA}}$ yields the relative costs reported in Table~\ref{tab:mixer_costs} and \ref{tab:mixers}.

\paragraph{Idealized cost.}
The simplest cost estimate uses pure placements only: $c_X = \frac{\text{throughput}_{\text{FA}}}{ \text{throughput}_X}$.
This yields relative costs of 0.44 for SWA ($w{=}4096$), 0.16 for KDA, and 0.10 for GDN (Table~\ref{tab:mixer_costs}, \textbf{Idealized$^*$}).
A measurement with a slightly different vLLM modeling file and SWA window size $w{=}2048$ gave lower costs (SWA: 0.16, KDA: 0.11, GDN: 0.09); the SWA difference is largely explained by the halved window size (Table~\ref{tab:mixer_costs}, \textbf{Idealized}).  \emph{We note that the later Idealized model is the one we used to obtain results in our preliminary experiments reported in the main body of this report}.

\paragraph{Regression fit.}
We benchmark 123 placements of a 48-layer Super Apriel~1.6 (~15B) on a single H100 GPU with 80GB memory. Unless stated otherwise, we use \texttt{vLLM~v0.14.0rc1} with FlashInfer attention backend, \texttt{VLLM\_USE\_FLASHINFER\_SAMPLER}=1, \texttt{gpu\_mem\_utilization}=0.8\footnote{Higher \texttt{gpu\_mem\_utilization} resulted in OOM errors for some placement presumably due to memory fragmentation.}, and \texttt{FULL\_AND\_PIECEWISE} CUDA graph mode.
Each placement is evaluated on 500 requests with output length 16k tokens and input length of 1 token using \texttt{vllm bench throughput}\footnote{\url{https://docs.vllm.ai/en/v0.14.0/cli/bench/throughput/}} script. The placements span the full range from pure (all 48 layers of a single mixer) to uniformly mixed compositions across all four mixer types (FA, SWA, KDA, GDN). We use \texttt{surgery} mode for this benchmarking, meaning that for each run we only load the placement in question to avoid memory waste (as opposed to loading all $48^4$ mixers).

We fit the linear model in Eq.~\ref{eq:cost-model} via ordinary least squares on $1/\text{throughput}$.
When fit on all 123 placements, the model achieves $R^2 = 0.576$ with a mean absolute latency prediction error of 35.9\%. This high error is driven by the singleton placements, i.e. placements with a small number of mixers of a single type. Excluding placements where any mixer type has fewer than 3 layers (retaining 84 of 123 placements) yields a substantially better fit: $R^2 = 0.977$, mean error 7.6\%, with relative costs of 0.48 (SWA), 0.21 (KDA), and 0.14 (GDN). These regression-based costs are consistently higher than the idealized estimates, reflecting the overhead of mixing heterogeneous mixer types in vLLM's unified KV cache allocator.

\begin{figure}[H]
    \centering
    \includegraphics[width=0.9\linewidth]{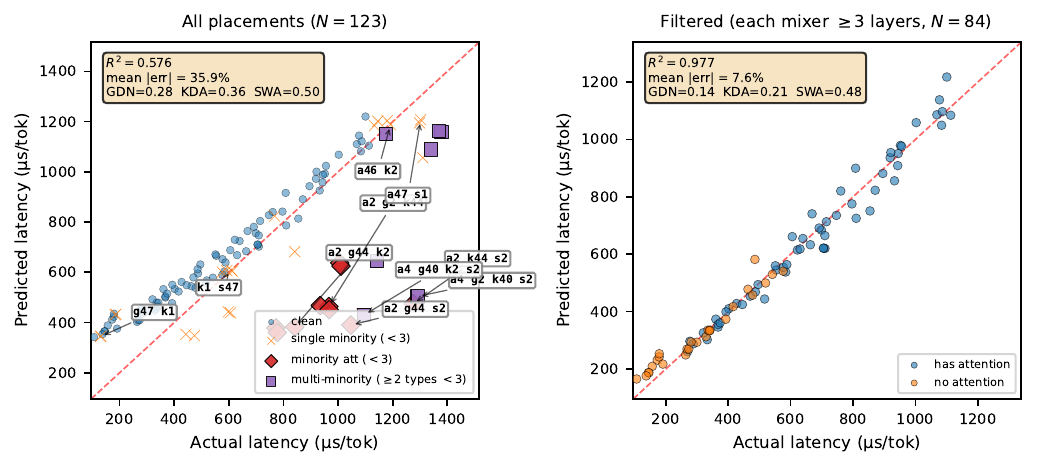}
  \caption{Predicted vs.\ actual per-token latency on 15B Super Apriel under the linear cost model $1/\text{throughput} = \sum_m c_m \cdot n_m$. \textbf{Left:} fit on all 123 placements ($R^2 = 0.576$); we observe large prediction error mostly on layouts with strong minority mixers, see discussion in the main text. \textbf{Right:} excluding placements where any mixer has fewer than 3 layers yields a tight fit ($R^2 = 0.977$, mean error 7.6\%) on the 84 remaining mixed placements. We use these filtered regression coefficients as the cost model for placement search.}
  \label{fig:cost-model-fit}
\end{figure}

\paragraph{Outlier placements.}
\label{sec:outlier_placements}
As discussed above, including singleton placements with strong minority token mixers leads to significantly lower $R^2$ of the linear fit. In Figure~\ref{fig:cost-model-fit} we plot linear fits induced by all tested placements (right) as well as with strong minority placements (aka singletons) excluded. To better understand the nature of the outlier placements, we grouped them into three categories and annotated some of them in the left plot in Figure~\ref{fig:cost-model-fit}. \textbf{This annotation reveals that all outliers are singletons but not all singletons are outliers.} More specifically, we observe that: adding a minority of a slower mixer into a fast majority seem to cause disproportionate overhead (attention or SW into GDN/KDA). Adding a minority of a faster mixer into a slow majority does not seem to have the same effect.
Based on this finding, here we simply restrict candidates from the refinement phase (Section~\ref{sec:placement}) to placements that include 0 or at least 3 of each mixer type. This prevents the search from selecting minority-placement candidates, which would not fit the cost model.

We attribute this to the current vLLM design, where the slowest mixer type determines memory allocation (KV cache pages, block tables) and scheduling constraints for the whole model. The per-layer latency of token mixers should in principle combine linearly. Currently, vLLM's hybrid model support (unified  page size, recurrent state management, CUDA graph handling) introduces overhead that breaks this assumption for some configurations. \emph{As vLLM's hybrid model support matures — which is likely given the trend toward hybrid  architectures --- the linear cost model should become more accurate, making placement optimization via simple cost budgets even more accurate.} Unless stated otherwise, in this work we rely on the coefficients obtained either from the idealized model or the regression fit with all singletons excluded (Figure~\ref{fig:cost-model-fit} right).

\begin{figure}[H]
    \centering
    \includegraphics[width=1\linewidth]{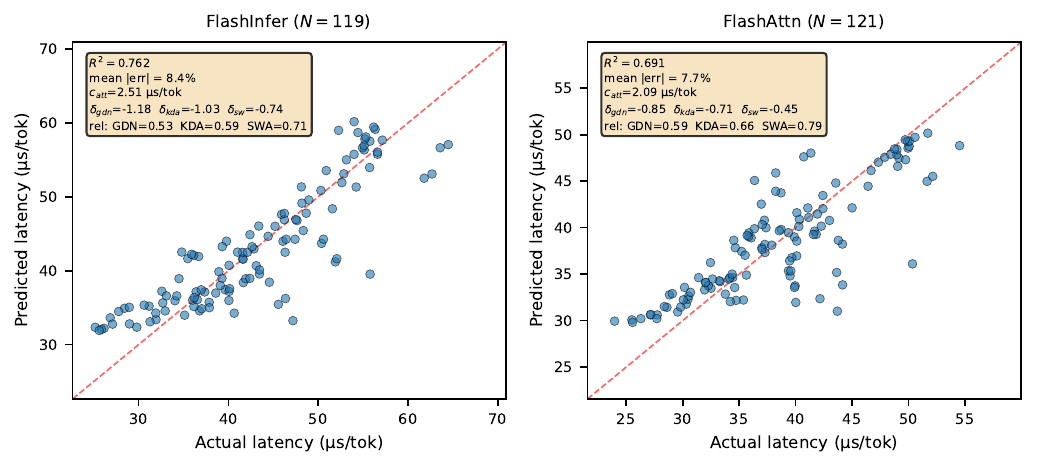}
  \caption{Linear cost model fit for 0.5B supernet. Each point represents a distinct layer placement evaluated at 16k sequence length. The model decomposes per-token latency as $1/\text{throughput} = \sum_i c_i \cdot n_i$, where $c_i$ is the
  per-layer cost and $n_i$ the layer count for mixer type $i \in {\text{att, gdn, kda, sw}}$. Coefficients are reported as deltas relative to the attention baseline ($\delta_i = c_i - c_\text{att}$) and as relative costs ($c_i /
  c_\text{att}$). \textbf{Left}: FlashInfer attention backend. \textbf{Right}: FlashAttn attention backend. For 0.5B model we do not exclude outliers and directly use the coefficients induced by the FlashAttn backend.}
  \label{fig:cost-model-fit-05b}
\end{figure}

\section{Throughput evaluation}
\label{app:throughput_eval}

\begin{table}[H]
\centering
\caption{Throughput evaluation setup for all models. We run all benchmarking on a single H100 GPU. All experiments use \texttt{TP=1}, \texttt{bfloat16} precision and \texttt{load format=dummy} with prefix caching disabled. CUDA graph mode is \texttt{FULL\_AND\_PIECEWISE} and vLLM \texttt{v0.14.0 rc.1} unless stated otherwise. Some external models were evaluated with vLLM \texttt{v0.18}, which may auto-adjust some settings like cuda-graph mode. Final evaluations use 1000 prompts at 16k sequence length and 500 prompts at 32k sequence length. We use \texttt{vllm bench throughput}\protect\footnotemark{} script for these evaluations. We aim to compare all models in identical setup. We lower the \texttt{gpu\_mem\_utilization} to prevent OOM errors for some models (potentially caused by memory fragmentation in heterogeneous hybrids).}
\label{tab:eval-setup}
\resizebox{\textwidth}{!}{
\begin{tabular}{@{}lllllllll@{}}
\toprule
\textbf{Model} & \textbf{Size} & \textbf{vLLM} & \textbf{Attention} & \textbf{GPU} & \textbf{Max} & \textbf{Input} & \textbf{Output} & \textbf{Num} \\
               &               & \textbf{Version} & \textbf{Backend} & \textbf{Util.} & \textbf{Len} & \textbf{Len} & \textbf{Len} & \textbf{Prompts} \\
\midrule
\multicolumn{9}{l}{\textit{Supernet Models}} \\
\midrule
Super Apriel 0.5B & 0.5B & v0.14.0 rc1 & FlashInfer & 0.9 & 65536 & 1 & 16000 & 100 \\
(all placements) &      &        &            &     &       &   &       &     \\
\cmidrule(lr){1-9}
Super Apriel 15B & 15B & v0.14.0 rc1 & FlashInfer & 0.8 & 65536 & 1 & 16000 & 500$^\dagger$ \\
(multiple placements) &  &   &            &     &       &   &       &     \\
               &     &        &            &     &       & 1 & 16000 & 1000 \\
               &     &        &            &     &       & 1 & 32000 & 500 \\
\midrule
\multicolumn{9}{l}{\textit{Baseline Models (Fixed Architecture)}} \\
\midrule
Apriel-H-1-15B & 15B & v0.14.0 rc1 & FlashInfer & 0.8 & 65536 & 1 & 16000 & 1000 \\
Thinker-SFT &     &        &            &     &       &   &       &     \\
               &     &        &            &     &       & 1 & 32000 & 500 \\
\cmidrule(lr){1-9}
OLMo-Hybrid & 7B & v0.18 & auto-select$^*$ & 0.8 & 32768 & 1 & 16000 & 1000 \\
Instruct-SFT &    &        &            &     &       &   &       &     \\
               &    &        &            &     &       & 1 & 32000 & 500 \\
\cmidrule(lr){1-9}
Qwen3.5-27B & 27B & v0.18 & auto-select$^*$ & 0.8 & 65536 & 1 & 16000 & 1000 \\
            &     &        &            &     &       & 1 & 32000 & 500 \\
\cmidrule(lr){1-9}
Falcon-H1R-7B & 7B & v0.18 & auto-select$^*$ & 0.8 & 65536 & 1 & 16000 & 1000 \\
            &     &        &            &     &       & 1 & 32000 & 500 \\
\cmidrule(lr){1-9}
NVIDIA Nemotron & 12B & v0.18 & auto-select$^*$ & 0.8 & 65536 & 1 & 16000 & 1000 \\
Nano-12B-v2 &     &        &            &     &       & 1 & 32000 & 500 \\
\cmidrule(lr){1-9}
NVIDIA Nemotron & 30B & v0.18 & auto-select$^*$ & 0.8 & 65536 & 1 & 16000 & 1000 \\
3-Nano-30B &     &        &            &     &       & 1 & 32000 & 500 \\
\bottomrule
\end{tabular}
}
\end{table}
\footnotetext{\url{https://docs.vllm.ai/en/v0.14.0/cli/bench/throughput/}}

\begin{table}[H]
\centering
\caption{Throughput results with a custom implementation based on vLLM \texttt{v0.18.0}, capable of simultaneously serving multiple placements. We run all benchmarking on a single H100 GPU. All experiments use \texttt{TP=1}, \texttt{bfloat16} precision, \texttt{load format=dummy} with prefix caching disabled, attention backed \texttt{FLASH\_ATTN}, and CUDA graph mode \texttt{FULL\_AND\_PIECEWISE}. All evaluations use 1000 prompts with input length 1 and sequence length 16000. We lower the \texttt{gpu\_mem\_utilization} to 0.8 to prevent OOM errors for some models (potentially caused by memory fragmentation in heterogeneous hybrids). In the multiple placement serving case, we report throughput numbers for the non-{all-FA} placement.}
\label{tab:eval-multiplacement-throughput}
\begin{tabular}{lll}
\toprule
\textbf{Placements} & \textbf{Throughput (tok/s)} & \textbf{Speedup over all-FA} \\
\midrule
\multicolumn{3}{l}{\textit{Single placement}} \\
\midrule
{all-FA} & 926 & -- \\
{Idealized|All--18} & 1791 & 1.93$\times$ \\
{Idealized|All--7} & 1804 & 1.95$\times$ \\
{Idealized|All--6} & 1984 & 2.14$\times$ \\
{Idealized|Lklhd--6} & 1978 & 2.14$\times$ \\
\midrule
\multicolumn{3}{l}{\textit{Multiple placements}} \\
\midrule
{all-FA + Idealized|All--18} & 1660 & 1.79$\times$ \\
{all-FA + Idealized|All--7} & 1667 & 1.8$\times$ \\
{all-FA + Idealized|All--6} & 1255 & 1.36$\times$ \\
{all-FA + Idealized|Lklhd--6} & 1257 & 1.36$\times$ \\
\bottomrule
\end{tabular}
\end{table}

\section{Global placement sampling}
\label{app:global_placement}
\begin{figure}[htbp]
  \centering
  \begin{subfigure}{0.45\textwidth}
    \centering
    \includegraphics[width=\linewidth]{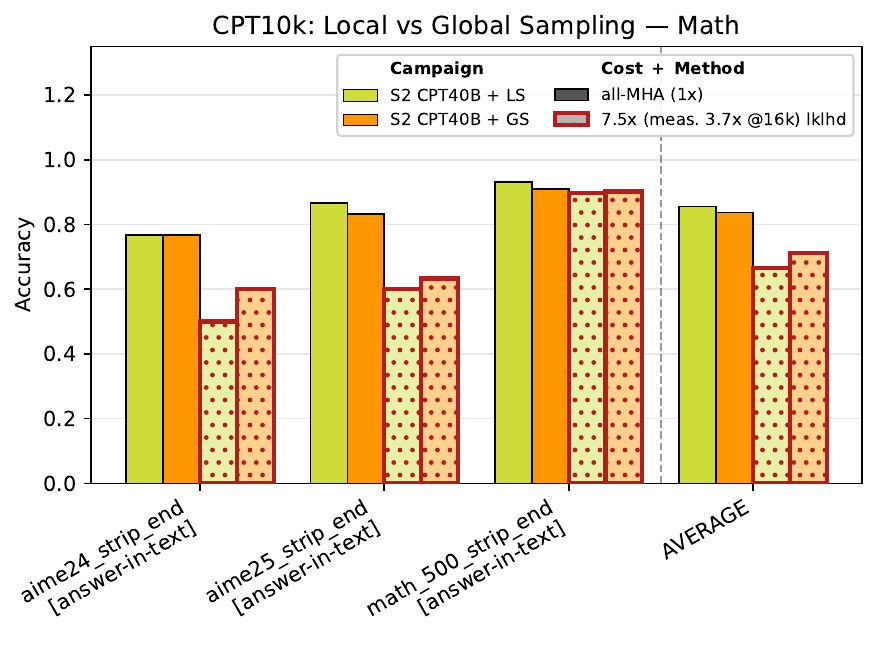}
    \caption{GS vs. LS dev. performance}
    \label{fig:left}
  \end{subfigure}
  \begin{subfigure}{0.45\textwidth}
    \centering
    \includegraphics[width=\linewidth]{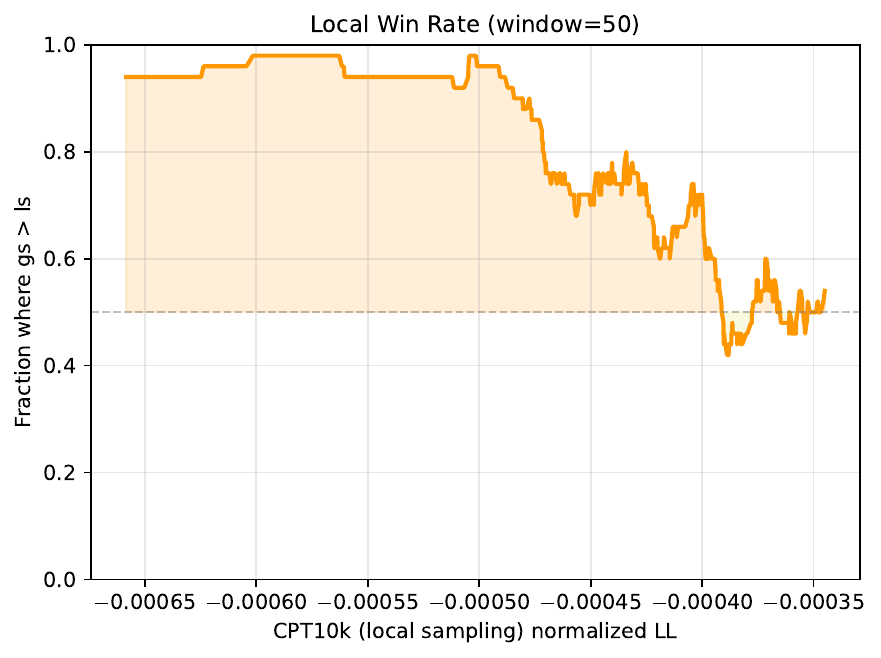}
    \caption{Log-likelihood comparison}
    \label{fig:right}
  \end{subfigure}
  \caption{Comparison local sampling (LS) vs global
  sampling (GS) on the 15B supernet checkpoint CPTed for 40B tokens. (a) Accuracy on math reasoning tasks (AIME-24, AIME-25, MATH-500) for 15B supernet checkpoints with local sampling (LS) vs global
  sampling (GS). Each group shows the full-MHA baseline and the optimized 7.5x-speedup placement found by surrogate
  search (cost 6.4, hatched bars, AIME+MATH likelihood as target). Global sampling exceeds local sampling across all tasks. (b) Local win rate of the global sampling (GS) checkpoint over local sampling (LS) across 600 identical layer placements, evaluated by log-likelihood on AIME+MATH traces generated by all-MHA placement from the corresponding checkpoint. For each placement, both checkpoints are scored by log-likelihood on pre-computed AIME+MATH completion traces (generated by the corresponding all-attention placement) and normalized by total completion tokens (to avoid generation length bias). Placements are sorted by the LS model's normalized LL (x-axis). The y-axis shows the fraction of placements where GS
   assigns higher likelihood within a sliding window of 50 placements. The dashed line at 0.5 indicates parity. The curve remains above 0.5 across most of the placements range, demonstrating that GS training yields better-calibrated model.}
  \label{fig:gs_vs_ls}
\end{figure}

Global sampling decouples the two choices: first sample an allocation
uniformly from the
$\binom{L+\abs{\mathcal{M}}-1}{\abs{\mathcal{M}}-1}$ compositions, then
sample a placement uniformly within that
allocation.\footnote{A counterintuitive consequence: conditional on
observing type~$m$ at one layer, other layers are \emph{more} likely to
also be type~$m$ under global than local sampling
(${\approx}\,0.40$ vs.\ $1/\abs{\mathcal{M}} = 0.25$ for $L{=}48$,
$\abs{\mathcal{M}}{=}4$).  Seeing type~$m$ at one layer makes a
type-$m$-heavy allocation more likely, which in turn raises the
probability of type~$m$ elsewhere.}
Whether this affects the trained mixers depends on the landscape's
interaction structure.  Each mixer's gradient depends on the full placement
through the residual stream and backpropagated loss.  When mixer outputs
are close to the original attention, the residual stream is approximately
placement-independent and both schemes train each mixer to the same
parameters; the cluster expansion (Section~\ref{sec:placement}) diagnoses
this as a unary landscape.  When perturbations are large enough to couple
layers, the schemes diverge, but which produces better mixers is an
empirical question (Figure~\ref{fig:gs_vs_ls}).
We use global sampling in the SFT phase.

\section{Surrogate Search Details}
\label{app:search}

\subsection{Feature Indicator Equations}

\paragraph{Order 1 (self-energy).}
For each layer~$i$ and type $m \in \mathcal{M}$:
\[
  \phi^{(1)}_{i,m} = \mathbb{1}[x_i = m]
\]
These $L \times \abs{\mathcal{M}}$ indicators capture per-layer preferences.  The $\abs{\mathcal{M}}$
allocation counts $n_m = \sum_i \phi^{(1)}_{i,m}$ are included as
explicit features.

\paragraph{Order 2 (pairwise interactions).}
For each pair of layers $(i, j)$ with $|i - j| \leq r_{\max}$ and each
pair of types $(m, m')$:
\[
  \phi^{(2)}_{ij,mm'} = \mathbb{1}[x_i = m \wedge x_j = m']
\]

\paragraph{Order 3 (triplet interactions).}
For contiguous triplets $(i, i{+}1, i{+}2)$ and type triples
$(m, m', m'')$:
\[
  \phi^{(3)}_{i,mm'm''} = \mathbb{1}[x_i = m \wedge x_{i+1} = m'
  \wedge x_{i+2} = m'']
\]

\paragraph{Feature counts} are in Table~\ref{tab:features}.
\begin{table}[H]
  \centering
  \caption{Feature counts at each expansion order ($L = 48$, $\abs{\mathcal{M}} = 4$).}
  \label{tab:features}
  \begin{tabular}{llr}
    \toprule
    \textbf{Order} & \textbf{Features} & \textbf{Cumulative} \\
    \midrule
    1 (unary)      & $L\abs{\mathcal{M}} + \abs{\mathcal{M}}$          & 196    \\
    2 (range 1)    & $(L{-}1)\abs{\mathcal{M}}^2$      & 948    \\
    2 (range 2)    & $(L{-}2)\abs{\mathcal{M}}^2$      & 1,684  \\
    2 (range 3)    & $(L{-}3)\abs{\mathcal{M}}^2$      & 2,404  \\
    3 (triplets)   & $(L{-}2)\abs{\mathcal{M}}^3$      & 5,348  \\
    \bottomrule
  \end{tabular}
\end{table}

\subsection{Bayesian Linear Surrogate}

The surrogate is a Bayesian linear model with a Normal-Inverse-Gamma
(NIG) conjugate prior on the weight vector $\mathbf{w}$ and noise
variance $\sigma^2$:
\[
  \mathbf{w} \mid \sigma^2 \sim \mathcal{N}(\mathbf{0},\, \sigma^2
  \alpha^{-1} \mathbf{I}), \qquad
  \sigma^2 \sim \mathrm{InvGamma}(a_0, b_0)
\]
Given data $(\mathbf{X}, \mathbf{y})$ where $\mathbf{X}$ is the
cluster-expansion feature matrix, the posterior is available in closed
form:
\[
  \boldsymbol{\mu}_n = \mathbf{V}_n \mathbf{X}^\top \mathbf{y},
  \qquad
  \mathbf{V}_n = (\mathbf{X}^\top \mathbf{X} + \alpha \mathbf{I})^{-1}
\]
The predictive distribution for a new placement $\mathbf{x}^*$ is
Student-$t$ with scale:
\[
\hat{\sigma}(\mathbf{x}^*) = \sqrt{c\,(1 +
\boldsymbol{\phi}(\mathbf{x}^*)^\top \mathbf{V}_n\,
\boldsymbol{\phi}(\mathbf{x}^*))}
\]
where $c = b_n / a_n$.  The variance grows with distance from the
training data, providing calibrated uncertainty without resampling.

The log marginal likelihood,
\[
  \log p(\mathbf{y} \mid \alpha, a_0, b_0)
  = -\tfrac{n}{2}\log 2\pi
    + \tfrac{d}{2}\log\alpha
    - \tfrac{1}{2}\log|\mathbf{A}|
    + a_0 \log b_0 - a_n \log b_n
    + \log\Gamma(a_n) - \log\Gamma(a_0)
\]
where $\mathbf{A} = \mathbf{X}^\top\mathbf{X} + \alpha\mathbf{I}$,
is used for expansion-order selection
(Section~\ref{sec:app:adaptive}).

The posterior-mean coefficients $\boldsymbol{\mu}_n$ decompose as MRF
potentials---unary $V_i(m)$, pairwise $V_{ij}(m, m')$, and (when the
expansion includes order~3) triplet $V_{ijk}(m, m', m'')$---which
admit the exact Viterbi optimization described next.

\subsection{Viterbi DP}

The cost-constrained Viterbi augments the DP state with partial
allocation counts.  The state at layer~$i$ is
$(x_{i-m+1}, \ldots, x_i, n_1, \ldots, n_{\abs{\mathcal{M}}})$ where $m$ is the memory
(interaction range).  For $\abs{\mathcal{M}} = 4$, $L = 48$ with pairwise range~1,
this yields approximately 80{,}000 states at the final layer.  The
forward pass completes in seconds on CPU\@.  Each feasible allocation
corresponds to a distinct set of count values at the final layer; the
optimal placement per allocation is recovered by backtracking.
All cost-feasible allocations---up to $\binom{L+\abs{\mathcal{M}}-1}{\abs{\mathcal{M}}-1}$ in the
unconstrained case ($20{,}825$ for $L{=}48$, $\abs{\mathcal{M}}{=}4$)---are solved
simultaneously in a single pass.

\subsection{Risk-Bucket Candidate Selection}

Candidate placements for supernet evaluation are generated from the
posterior mean and selected using a risk-bucket scheme that controls the
explore--exploit balance.

\paragraph{Candidate generation.}
Extract MRF potentials from $\boldsymbol{\mu}_n$, run the
all-allocations Viterbi pass, and keep the top~$M$ placements per
allocation to obtain an allocation-diverse candidate pool.

\paragraph{Scoring and selection.}
Each candidate is scored under two buckets that combine the posterior
predictive mean $\hat{\mu}$ and standard deviation $\hat{\sigma}$:
\begin{itemize}
  \item \textbf{Safe} (sign${}=-1$): score
    $= \hat{\mu} - \beta\,\hat{\sigma}$.  Penalizes uncertainty,
    selecting placements the model is confident are good (exploitation).
  \item \textbf{Upside} (sign${}=+1$): score
    $= \hat{\mu} + \beta\,\hat{\sigma}$.  Rewards uncertainty, selecting
    placements whose true quality could exceed the prediction
    (exploration).
\end{itemize}
A configurable quota (e.g., 70\% safe / 30\% upside) apportions the
per-round evaluation budget across buckets.  Each bucket independently
ranks its eligible candidates and selects the top~$k$ by its scoring
rule; ties are broken by $\hat{\mu}$.  A mean-floor filter optionally
restricts each bucket to candidates above a specified quantile of
$\hat{\mu}$ in the pool.

\paragraph{Refinement loop.}
The selected placements are evaluated against the real supernet, added to
the training set, and the surrogate is refit.  Defaults: 4 refinement
rounds, 500 evaluations per round.  The candidate pool targets 1000
placements, distributed as $\lceil 1000 \,/\, |\text{feasible
allocations}| \rceil$ per allocation to maintain diversity across cost
levels.

\subsection{Adaptive Exploration Protocol}
\label{sec:app:adaptive}

\begin{enumerate}
  \item Sample $N_0 = 1000$ diverse feasible placements using
    cost-weighted allocation sampling.
  \item Evaluate all placements against the supernet.
  \item Select the expansion order by comparing log marginal likelihoods
    across all feasible (order, range) combinations (subject to the
    feature guard below).  The configuration with the highest evidence
    is selected.
  \item Refit the Bayesian surrogate at the selected expansion, run
    risk-bucket refinement, add evaluations, and repeat expansion
    selection.
\end{enumerate}
The expansion selection is deterministic (no CV splits, no RNG) and is
re-run each iteration as the training set grows; if a higher order or
longer range gains the most evidence, the surrogate is upgraded.

\paragraph{Feature guard.}
The feature guard enforces a 2:1 sample-to-feature ratio.  At 1000
samples: unary features (196) are within budget; pairwise range-1
features bring the total to 948 (within budget); triplet features (5348
cumulative) are blocked until ${\sim}2500$ samples are collected.

\section{Evaluation Protocols}
\label{app:evaluation}
\paragraph{Evaluator types.}
Three evaluator types are used for placement scoring:
\emph{generative} (GSM8K --- exact-match accuracy on generated
solutions), \emph{likelihood} (MMLU --- log-probability scoring over
multiple-choice options), and \emph{trace log-likelihood} (log-probability
of pre-computed teacher traces for faster evaluation).

\paragraph{Dynamic placement switching.}
Each placement is evaluated by switching the active mixer configuration
via the serving engine's dynamic placement API --- no model reloading is
required.  This enables evaluating hundreds of placements per hour from
a single running inference instance.

\section{Detailed Downstream results}
\label{app:downstream_evals}

Table~\ref{tab:benchmark_details} summarizes the evaluation protocol for each benchmark. Prompts and task formatting follow the \texttt{lm-eval-harness}~\citep{eval-harness} defaults unless otherwise noted.

\begin{table}[h]
\centering
\caption{Per-benchmark evaluation details.}
\label{tab:benchmark_details}
\small

\resizebox{\textwidth}{!}{
\begin{tabular}{llcll}
\toprule
\textbf{Benchmark} & \textbf{Scoring} & \textbf{Shots} & \textbf{Filter / Extraction} & \textbf{Notes} \\
\midrule
\multicolumn{5}{l}{\textit{Dev Benchmarks}} \\
MMLU         & log-liklh & 5 & ---                         & 10 problems/subject \\
GSM8K        & gen     & 5 & \texttt{flexible-extract}   & greedy decoding \\
MATH-500     & gen     & 0 & \texttt{answer-in-text}     & greedy decoding \\
AIME '24     & gen     & 0 & \texttt{answer-in-text}     & greedy decoding \\
AIME '25     & gen     & 0 & \texttt{answer-in-text}     & greedy decoding \\
FDA          & gen     & 5 & exact-match                 & greedy decoding \\
SWDE         & gen     & 5 & exact-match                 & greedy decoding \\
NIAH         & gen     & 0 & exact-match                 & 16k context, greedy decoding \\
RULER-QA     & gen     & 0 & exact-match                 & 8k \& 16k context, greedy decoding \\
\midrule
\multicolumn{5}{l}{\textit{Unseen Benchmarks}} \\
MMLU-Pro & gen & 5 & exact-match & — \\
GPQA Diamond & gen & 0 & exact-match & — \\
HLE & gen & 0 & exact-match & — \\
LCB & gen & 0 & pass@1 & 3 seeds average \\
$\tau^2$-Bench & gen     & --- & GPT-4 user simulator & 3 seeds average \\
IFEval & gen & 0 & constraint adherence & 3 seeds average \\
AIME (NV) & gen & 0 & NeMo-Skills~\citep{nemo_skills2024} & AIME25 with \texttt{evalchemy} prompt, 10 seeds average \\
\bottomrule
\end{tabular}
}
\end{table}

\begin{table}[htbp]
\centering
\caption{Evaluation results across campaigns and dev. tasks.}
\label{tab:eval_results_on_dev}
\resizebox{\textwidth}{!}{
\begin{tabular}{llcccccccccccc}
\toprule
\textbf{Campaign} & \textbf{Config} & \textbf{@32k} & AIME'24 & AIME'25 & MATH-500 & GSM8K & MMLU & FDA & SWDE & RULER & NIAH & \textbf{Math} & \textbf{All} \\
\midrule
\multirow{9}{*}{S1: Distil. 69B} & all-attention & 1.0$\times$ & 90.0 & 86.7 & 85.0 & 80.7 & 64.7 & 81.3 & 91.0 & 63.4 & 99.8 & 85.6 & 80.4 \\
 & Idealized$|$Lklhd--6 & 6.2$\times$ & 76.7 & 60.0 & 83.4 & 76.7 & 63.0 & 79.4 & 90.2 & 64.4 & 27.2 & 74.2 & 74.2 \\
 & Idealized$|$All--6 & 6.1$\times$ & 63.3 & 63.3 & 84.2 & 96.1 & 59.6 & 80.9 & 87.8 & 78.7 & 26.6 & 76.7 & 76.7 \\
 & Idealized$|$All--7 & 4.3$\times$ & 76.7 & 56.7 & 85.4 & 95.0 & 63.5 & 81.6 & 90.4 & 67.1 & 29.8 & 78.4 & 77.0 \\
 & Idealized$|$All--18 & 2.0$\times$ & 86.7 & 76.7 & 85.6 & 95.9 & 64.2 & 81.4 & 89.5 & 56.9 & 82.0 & 86.2 & 79.6 \\
 & Reg$|$Lklhd--10 & 10.7$\times$ & 56.7 & 40.0 & 88.4 & 86.9 & 56.8 & 59.7 & 78.0 & 36.7 & 25.9 & 68.0 & 62.9 \\
 & Reg$|$Lklhd--13 & 6.9$\times$ & 66.7 & 56.7 & 89.0 & 92.3 & 63.9 & 51.0 & 77.0 & 55.4 & 28.4 & 76.2 & 69.0 \\
 & Reg$|$Lklhd--18 & 4.8$\times$ & 63.3 & 63.3 & 90.4 & 95.3 & 67.5 & 78.5 & 89.1 & 64.6 & 36.6 & 78.1 & 76.5 \\
 & Reg$|$Lklhd--26 & 2.9$\times$ & 80.0 & 93.3 & 90.8 & 94.5 & 69.1 & 77.8 & 88.6 & 66.1 & 67.6 & 89.6 & 82.5 \\
\midrule
\multirow{1}{*}{S2: S1 $\rightarrow$ SFT(1) 40B Single Placement} & Idealized$|$Lklhd--6 & 6.2$\times$ & 93.3 & 86.7 & 93.0 & 90.5 & 55.3 & 71.1 & 85.8 & 64.9 & 23.0 & 90.9 & 80.1 \\
\midrule
\multirow{8}{*}{S2: S1 $\rightarrow$ SFT(2) 60B Targeted 8 placements} & all-attention & 1.0$\times$ & 93.3 & 86.7 & 91.8 & 92.3 & 49.8 & 78.3 & 89.5 & 79.4 & 100.0 & 91.0 & 82.6 \\
 & Idealized$|$All--6 & 6.1$\times$ & 83.3 & 80.0 & 92.2 & 91.7 & 55.1 & 75.6 & 87.4 & 61.9 & 30.8 & 86.8 & 78.4 \\
 & Idealized$|$All--18 & 2.0$\times$ & 90.0 & 86.7 & 92.0 & 92.1 & 58.9 & 78.0 & 86.6 & 67.1 & 78.8 & 90.2 & 81.4 \\
 & Idealized$|$Lklhd--6 & 6.2$\times$ & 83.3 & 76.7 & 92.4 & 92.3 & 57.4 & 76.9 & 88.9 & 66.1 & 25.8 & 86.2 & 79.3 \\
 & Reg$|$Lklhd--10 & 10.7$\times$ & 76.7 & 66.7 & 90.6 & 90.8 & 55.6 & 65.2 & 82.9 & 48.6 & 27.4 & 81.2 & 72.1 \\
 & Reg$|$Lklhd--13 & 6.9$\times$ & 76.7 & 73.3 & 90.4 & 91.2 & 62.5 & 68.6 & 85.1 & 57.0 & 28.0 & 82.9 & 75.6 \\
 & Reg$|$Lklhd--18 & 4.8$\times$ & 86.7 & 76.7 & 92.4 & 91.7 & 58.6 & 81.7 & 89.8 & 60.5 & 57.0 & 86.8 & 79.7 \\
 & Reg$|$Lklhd--26 & 2.9$\times$ & 86.7 & 83.3 & 92.0 & 91.1 & 60.7 & 79.9 & 88.2 & 74.4 & 98.8 & 88.3 & 82.0 \\
\midrule
Falcon-H1R 7B & Hybrid & 4.6$\times$ & 66.7 & 70.0 & 90.0 & 87.6 & 68.1 & 63.6 & 89.6 & 72.0 & 92.4 & 78.6 & 75.9 \\
Nemotron-Nano 12B v2 & Hybrid & 5.9$\times$ & 60.0 & 60.0 & 83.6 & 94.3 & 78.2 & 82.5 & 92.1 & 65.2 & 58.0 & 74.5 & 77.0 \\
OLMo-Hybrid-Think 7B & Hybrid & 2.5$\times$ & 66.7 & 46.7 & 87.8 & 89.3 & 69.4 & 78.5 & 89.5 & 65.8 & 75.4 & 72.6 & 74.2 \\
Nemotron-3-Nano 30B & Hybrid MoE & 4.1$\times$ & 93.3 & 76.7 & 89.8 & 96.3 & 76.8 & 74.7 & 91.7 & 64.9 & 97.8 & 89.0 & 83.0 \\
Apriel-H1 15B & Hybrid & 2.0$\times$ & 80.0 & 53.3 & 92.8 & 95.4 & 64.4 & 73.2 & 88.6 & 71.0 & 85.6 & 80.4 & 77.3 \\
Qwen-3.5 27B & Hybrid MoE & 0.5$\times$ & 90.0 & 93.3 & 89.8 & 97.1 & 87.1 & 84.4 & 90.4 & 54.4 & 95.4 & 92.6 & 85.8 \\
\bottomrule
\end{tabular}}
\end{table}

\begin{table}[htbp]
\centering
\caption{Full benchmark results for the \texttt{S2: S1 $\rightarrow$ SFT(2) 60B Targeted 8} across all placements and tasks. For one placement we also include numbers from the \texttt{S1: Distil.} stage. We also include speedups @32k and @16k generation sequence lengths, as described in Appendix~\ref{app:throughput_eval}.}
\label{tab:s2v2_full}
\resizebox{\textwidth}{!}{
\begin{tabular}{lccccccccccccccccc}
\toprule
\textbf{Config} & \textbf{@32k} & \textbf{@16k} & AIME'24 & AIME'25 & MATH-500 & GSM8K & FDA & SWDE & RULER & Tau2 & MMLU-Pro & AIME(NV) & GPQA & HLE & LCB & IFBench & \textbf{All} \\
\midrule
Reg$|$Lklhd--10 & 10.69$\times$ & 4.2$\times$ & 76.7 & 66.7 & 90.6 & 90.8 & 65.2 & 82.9 & 48.6 & 23.4 & 68.2 & 24.4 & 52.5 & 4.5 & 50.2 & 56.2 & 57.2 \\
Reg$|$Lklhd--13 & 6.9$\times$ & 2.7$\times$ & 76.7 & 73.3 & 90.4 & 91.2 & 68.6 & 85.1 & 57.0 & 28.6 & 69.3 & 26.7 & 61.2 & 5.5 & 52.6 & 57.1 & 60.2 \\
Idealized$|$Lklhd--6 & 6.2$\times$ & 2.4$\times$ & 83.3 & 76.7 & 92.4 & 92.3 & 76.9 & 88.9 & 66.1 & 40.4 & 73.6 & 62.2 & 65.0 & 6.1 & 57.0 & 54.9 & 66.8 \\
Idealized$|$All--6 & 6.13$\times$ & 2.5$\times$ & 83.3 & 80.0 & 92.2 & 91.7 & 75.6 & 87.4 & 61.9 & 34.2 & 73.3 & 56.7 & 61.0 & 5.9 & 55.9 & 55.3 & 65.3 \\
Reg$|$Lklhd--18 & 4.76$\times$ & 2.2$\times$ & 86.7 & 76.7 & 92.4 & 91.7 & 81.7 & 89.8 & 60.5 & 46.2 & 76.3 & 74.4 & 68.7 & 6.6 & 64.8 & 59.3 & 69.7 \\
Reg$|$Lklhd--26 & 2.85$\times$ & 1.5$\times$ & 86.7 & 83.3 & 92.0 & 91.1 & 79.9 & 88.2 & 74.4 & 30.7 & 76.2 & 80.0 & 70.7 & 10.0 & 69.2 & 63.1 & 71.1 \\
Idealized$|$All--18 & 1.99$\times$ & 1.1$\times$ & 90.0 & 86.7 & 92.0 & 92.1 & 78.0 & 86.6 & 67.1 & 52.6 & 76.3 & 82.2 & 68.2 & 6.9 & 67.3 & 58.8 & 71.8 \\
all-attention & 1.0$\times$ & 1.0$\times$ & 93.3 & 86.7 & 91.8 & 92.3 & 78.3 & 89.5 & 79.4 & 56.7 & 76.8 & 82.7 & 72.0 & 8.2 & 68.6 & 63.1 & 74.2 \\
\midrule
S1: Distil. Idealized$|$All--6 & 6.13$\times$ & 2.5$\times$ & 63.3 & 63.3 & 84.2 & 96.1 & 80.9 & 87.8 & 78.7 & 12.3 & 74.5 & 43.7 & 57.9 & 5.0 & 44.4 & 59.9 & 60.8 \\
\midrule
\midrule
Apriel-1.6 & 1.0$\times$ & 1.0$\times$ & 90.0 & 83.3 & 86.2 & 83.6 & 84.5 & 90.2 & 69.1 & 62.6 & 78.8 & 81.7 & 70.7 & 8.8 & 78.8 & 66.5 & 73.9 \\
OLMo-Hybrid-Think 7B & 2.51$\times$ & 2.1$\times$ & 66.7 & 46.7 & 87.8 & 89.3 & 78.5 & 89.5 & 65.8 & 9.7 & 65.7 & 55.2 & 47.0 & 5.1 & 48.6 & 30.0 & 56.1 \\
Nemotron-Nano 12B v2 & 5.85$\times$ & 4.3$\times$ & 60.0 & 60.0 & 83.6 & 94.3 & 82.5 & 92.1 & 65.2 & 9.1 & 77.2 & 76.2 & 64.5 & 4.2 & 70.9 & 34.2 & 62.4 \\
Falcon-H1R 7B & 4.61$\times$ & 3.4$\times$ & 66.7 & 70.0 & 90.0 & 87.6 & 63.6 & 89.6 & 72.0 & 11.1 & 72.5 & 78.3 & 61.3 & 11.1 & 74.3 & 60.5 & 64.9 \\
Nemotron-3-Nano 30B & 4.09$\times$ & 2.8$\times$ & 93.3 & 76.7 & 89.8 & 96.3 & 74.7 & 91.7 & 64.9 & 31.3 & 78.3 & 89.1 & 73.0 & 10.6 & 75.3 & 71.5 & 72.6 \\
Qwen-3.5 27B & 0.55$\times$ & 0.5$\times$ & 90.0 & 93.3 & 89.8 & 97.1 & 84.4 & 90.4 & 54.4 & 93.6 & 86.1 & 83.7 & 85.5 & 18.0 & 85.5 & 77.8 & 80.7 \\
Apriel-H1 15B & 1.97$\times$ & 1.9$\times$ & 80.0 & 53.3 & 92.8 & 95.4 & 73.2 & 88.6 & 71.0 & 4.7 & 67.5 & 48.7 & 52.3 & 5.7 & 54.0 & 29.8 & 58.4 \\
\bottomrule
\end{tabular}}
\end{table}

\section{Analysis of the Placement Landscape}
\label{app:analysis}

\subsection{Preliminary experiment}
\label{app:prelim_exp}

A preliminary experiment explored the placement landscape using an idealized cost-model and a wider set of benchmarks as optimization metric.
The idealized cost-model, described as \texttt{Idealized} in Section~\ref{app:cost_models}, relied on the throughput of pure placements instead of using a regression to obtain per-mixer latency coefficient, and used a smaller window for SWA.
This idealized cost-model largely over-estimates the speedup of SWA compared to FA, and also over-estimates the speedup of KDA and GDN.
The optimization objective was an aggregate of all dev benchmarks, excluding GSM8K.
Running all dev benchmarks as part of the optimization was costly, and could result in benchmark overfitting. In the main experiments, only MATH500, AIME24 and AIME25 are used as optimization metrics.

\begin{figure}
    \centering
    \includegraphics[width=0.7\linewidth]{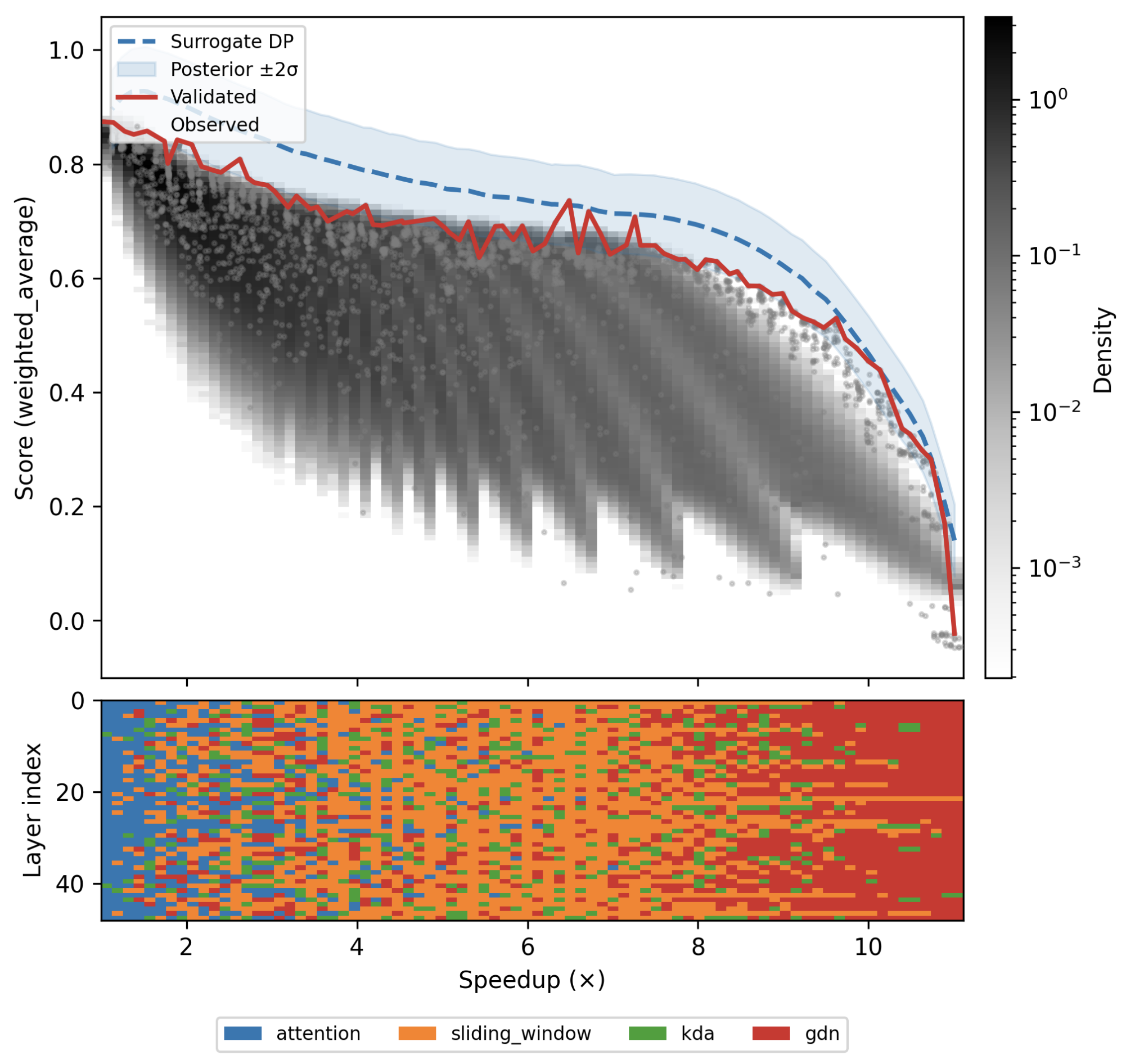}
    \caption{Preliminary Pareto frontier for the 15B supernet with \texttt{Idealized} cost-model and aggregate score over \texttt{All} dev benchmarks except GSM8K.
    \textbf{Top panels:}~Score for feasible placements (grayscale density, log scale), observed supernet evaluations (grey dots), surrogate-predicted optima (blue curve), and validated optima (red solid line).
    \textbf{Botton panels:}~Layer-wise mixer assignment of the optimal placement at each speedup-level.
    The frontier shows a performance plateau from $2.5{\times}$ to $7{\times}$ speedup.
    }
    \label{fig:pareto_frontier_prelim}
\end{figure}

The preliminary pareto frontier for the 15B model (Figure~\ref{fig:pareto_frontier_prelim}) shows a performance plateau between $2.5{\times}$ and $7{\times}$ speedup.
Placements with large speedups up to $7{\times}$ achieve performance close to the full-attention baseline.
The different cost model and the different set of benchmarks could both explain this plateau.
Benchmarks like NIAH could contribute to a plateau by showing high performance on attention-heavy placements and dropping rapidly on KDA or GDN-heavy placements.
The idealized cost-model largely over-estimates the speedup brought by SWA. One effect is that the SWA-dominated region is dilated in that frontier, ranging from $3{\times}$ to $7{\times}$ and approximately matching the plateau.

\subsection{Speculative Decoding: Estimating Acceptance Rate and Landscape Search}\label{app:speculative}
\paragraph{Estimation Procedure.}
We first recall the speculative decoding algorithm \citep{leviathan2023fast}. Let $q$ be a draft model and $p$ be a target model. Given a prefill $\bx$ and a hyperparameter $\gamma$, we first generate $\gamma$ tokens from $q(\cdot \,|\, \bx)$ autoregressively, denoted by $x_1,\hdots,x_n$.
Then, we perform a parallel forward pass on the target $p$ to obtain $p(x_i \,|\, \bx, \bx_{1:i-1})$ for $i \in [n]$. Going from $i=1$ to $i=\gamma$, each token is accepted with probability
$\min(1, p(x_i \,|\, \bx,\bx_{1:i-1})/q(x_i\,|\,\bx,\bx_{1:i-1}))$.
If a token is rejected for some position $i$, we discard tokens $i$ to $\gamma$, and generate the final token from the adjusted distribution $\operatorname{norm}\left(\max(0, p(\cdot \,|\, \bx,\bx_{1:i-1}) - q(\cdot \,|\, \bx, \bx_{1:i-1}))\right)$, where $\operatorname{norm}$ normalizes the distribution to sum to $1$.
This procedure comprises a single ``step'' of speculative decoding. We repeat these steps until we generate an EOS token or reach maximum length.

Let us define the expected step acceptance rate, denoted by $\bar{n}$, as the expected number of tokens generated in a single step of speculative decoding, and let $n$ be the stochastic counterpart. Further, let $A_i$ denote the event where the $i$th token is accepted. Then
\[
\bar{n} = \sum_{i=1}^\gamma \mathbb{P}[n \geq i] = \sum_{i=1}^\gamma \mathbb{P}\left[\cap_{j=1}^i A_j\right].
\]
Additionally,
\begin{align*}
    \mathbb{P}\left[\cap_{j=1}^i A_j\right] &= \mathbb{E}_{\bx \sim p, \bx_{1:i} \sim q(\cdot \,|\, \bx)}\left[\mathbb{P}\left[\cap_{j=1}^iA_j \,|\, \bx, \bx_{1:j-1}\right]\right] \\
    &= \mathbb{E}_{\bx \sim p, \bx_{1:i} \sim q(\cdot \,|\, \bx)}\left[\prod_{j=1}^i \min\left(1, \frac{p(x_j \,|\, \bx,\bx_{1:j-1})}{q(x_j \,|\, \bx, \bx_{1:j-1})}\right) \right]\\
    &= \mathbb{E}_{\bx \sim p, \bx_{1:i} \sim p(\cdot \,|\, \bx)}\left[\frac{q(\bx_{1:i}\,|\,\bx)}{p(\bx_{1:i} \,|\, \bx)}\prod_{j=1}^i\min\left(1, \frac{p(x_j \,|\, \bx,\bx_{1:j-1})}{q(x_j \,|\, \bx, \bx_{1:j-1})}\right)\right]\\
    &= \mathbb{E}_{\bx_{1:i},\bx \sim p}\left[\prod_{j=1}^i\min\left(\frac{q(x_j \,|\, \bx,\bx_{1:j-1})}{p(x_j \,|\, \bx,\bx_{1:j-1})}, 1\right)\right].
\end{align*}
Therefore, if for every step we collect
\[
\hat{n}(\bx,\bx_{1:i}) = \sum_{i=1}^\gamma \prod_{j=1}^i \min\left(\frac{q(x_j \,|\, \bx,\bx_{1:j-1})}{p(x_j \,|\, \bx,\bx_{1:j-1})}, 1\right),
\]
then we have an unbiased estimator of $\bar{n}$ since $\bar{n} = \mathbb{E}[\hat{n}(\bx,\bx_{1:i})]$.
Importantly, estimating $\bar{n}$ only requires access to completion traces generated under the target $p$, thus we can precompute these traces once and store them for estimating the acceptance rate of many different draft models.
With these precomputed traces, we only require access to log-probabilities at tokens along the trace, which we can obtain by efficient forward passes on $q$ without the need to generate new tokens.
We can define the expected step acceptance rate as $a \coloneqq \bar{n} / \gamma \in [0,1]$.

\begin{figure}
    \centering
    \begin{subfigure}[t]{0.49\textwidth}
        \includegraphics[width=\textwidth]{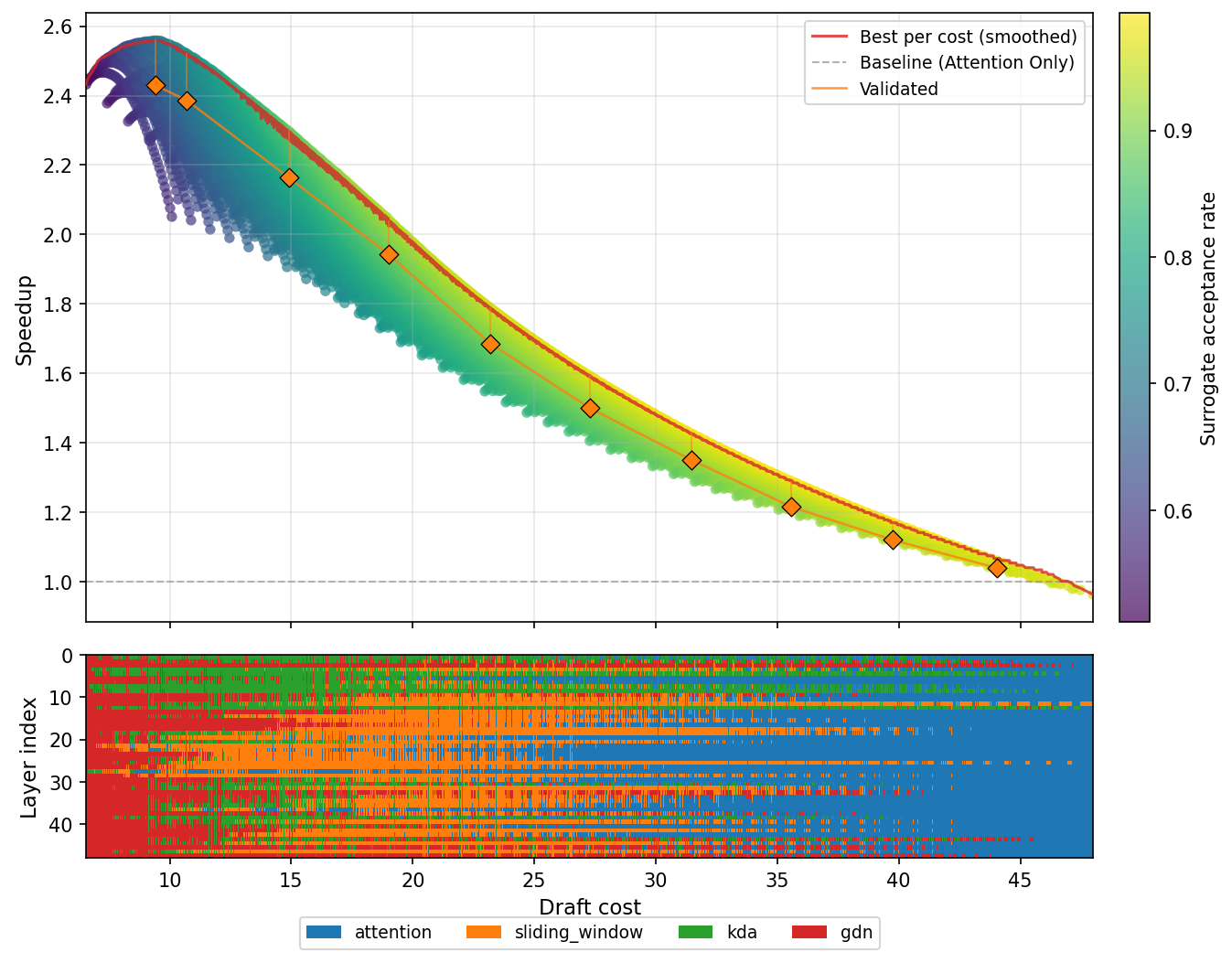}
        \caption{GSM8K}
    \end{subfigure}
    \hfill
    \begin{subfigure}[t]{0.49\textwidth}
        \includegraphics[width=\textwidth]{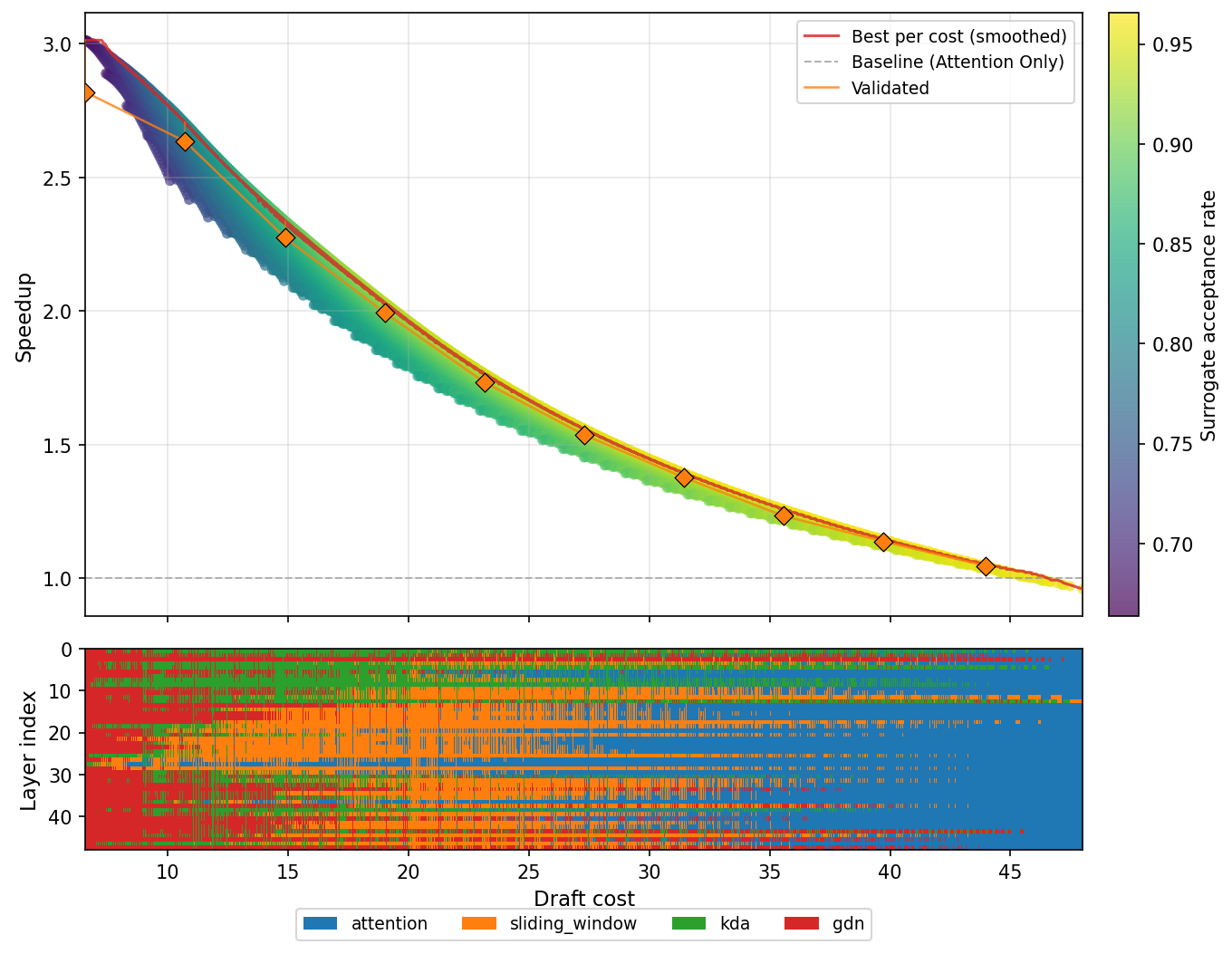}
        \caption{MATH500}
    \end{subfigure}

    \vspace{0.5cm}

    \begin{subfigure}[t]{0.49\textwidth}
        \includegraphics[width=\textwidth]{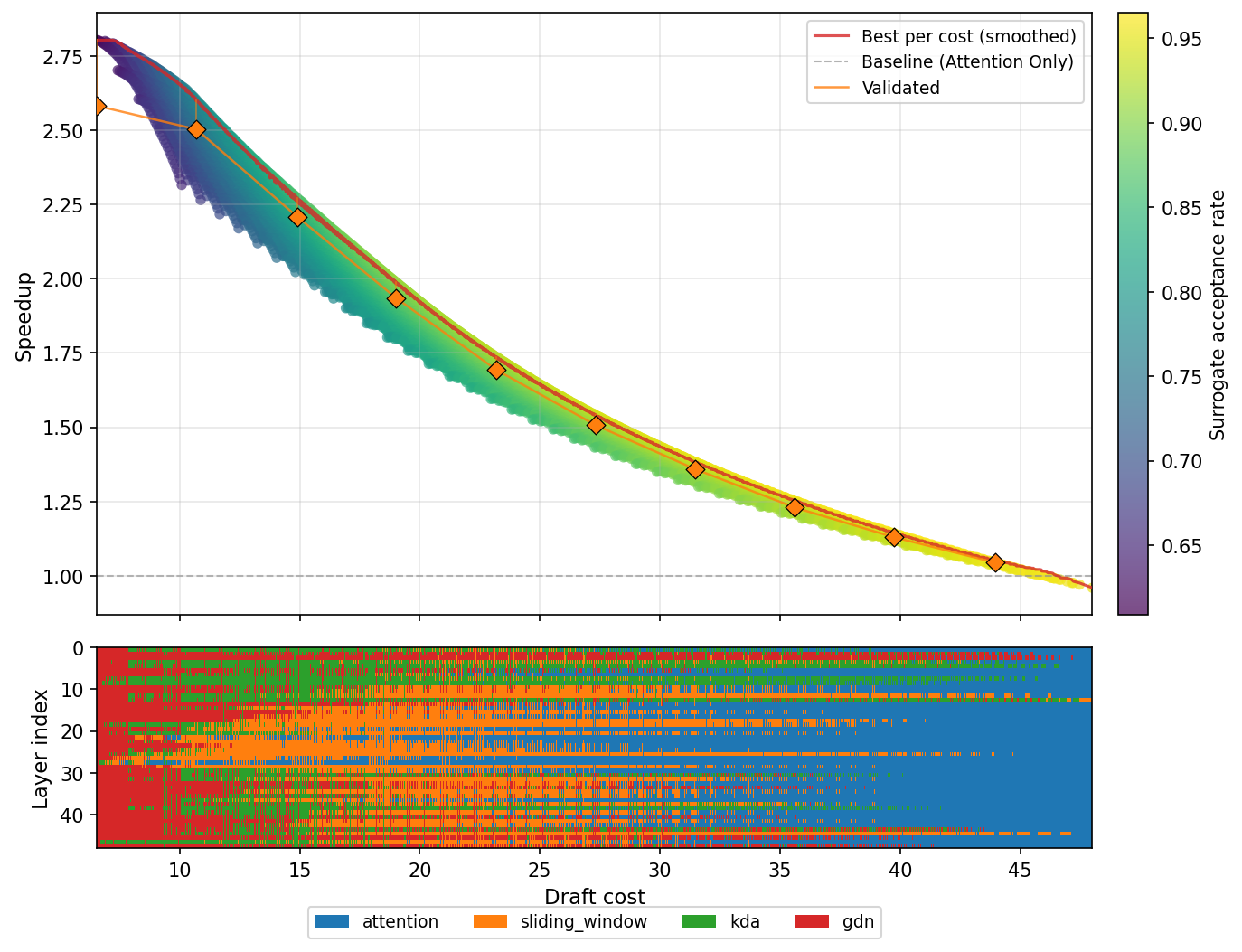}
        \caption{AIME24}
    \end{subfigure}
    \hfill
    \begin{subfigure}[t]{0.49\textwidth}
        \includegraphics[width=\textwidth]{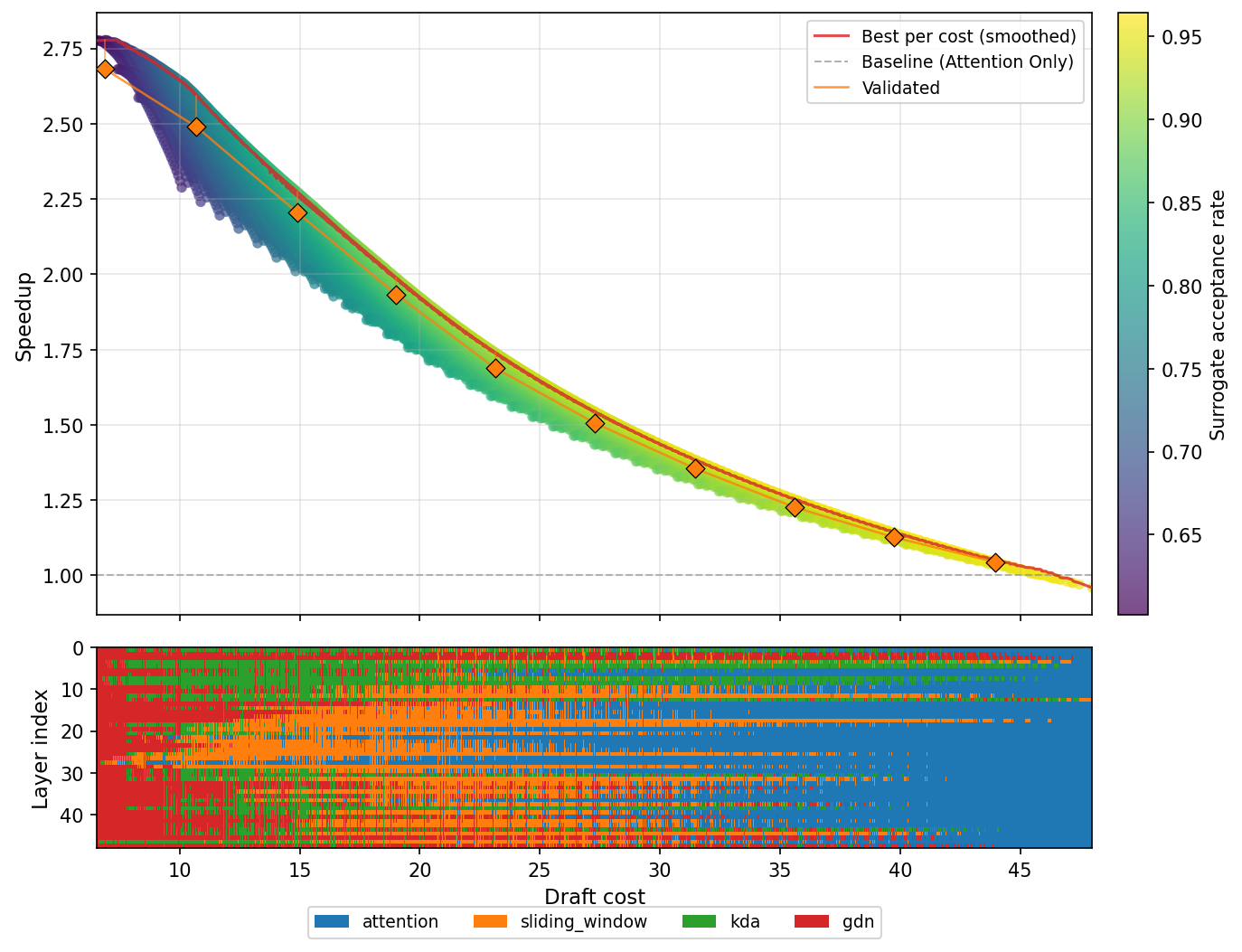}
        \caption{AIME25}
    \end{subfigure}
    \caption{Surrogate search over the landscape of speculative decoding acceptance rates, where the target model is fixed as the full-attention placement.}
    \label{fig:speculative}
\end{figure}

\paragraph{Surrogate Search and Acceptance Rate Landscape.}
To make search over the placement landscape feasible, we define a Bayesian linear surrogate model for acceptance rate, and fit this model as done in Appendix~\ref{app:search}. The resulting landscapes for the 15B model and different tasks are illustrated in Figure~\ref{fig:speculative}, and the average across the four tasks is illustrated in Figure~\ref{fig:averaged_speculative}. For these experiments, we use $\gamma = 8$.

\end{document}